\documentclass{article}

\usepackage[final, nonatbib]{neurips_data_2021}

\usepackage[utf8]{inputenc} % allow utf-8 input
\usepackage[T1]{fontenc}    % use 8-bit T1 fonts
\usepackage{hyperref}       % hyperlinks
\usepackage{url}            % simple URL typesetting
\usepackage{booktabs}       % professional-quality tables
\usepackage{amsfonts}       % blackboard math symbols
\usepackage{nicefrac}       % compact symbols for 1/2, etc.
\usepackage{microtype}      % microtypography
\usepackage{xcolor}         % colors
\usepackage{graphicx}
\usepackage{caption}
\usepackage{subcaption}
\usepackage{algorithm}      % algorithm
\usepackage[noend]{algpseudocode} % algorithm
\usepackage{wrapfig}        % wrap figures
\usepackage{booktabs}
\usepackage{multirow}
\usepackage{colortbl}
\usepackage{diagbox}
\usepackage{listings}
\usepackage{color}
\usepackage{xspace}
\usepackage{enumitem}
\usepackage{amssymb}
\usepackage{bbding}

\definecolor{dkgreen}{rgb}{0,0.6,0}
\definecolor{gray}{rgb}{0.5,0.5,0.5}
\definecolor{mauve}{rgb}{0.58,0,0.82}
\definecolor{bleudefrance}{rgb}{0.19, 0.55, 0.91}

\newcommand{\name}{GRB\xspace}

\title{Graph Robustness Benchmark: Benchmarking the Adversarial Robustness of Graph Machine Learning}

\author{%
  Qinkai Zheng$^{\dagger}$, Xu Zou$^{\dagger}$, Yuxiao Dong$^{\ddagger*}$, Yukuo Cen$^{\dagger}$, \\
  \textbf{Da Yin$^{\dagger}$, Jiarong Xu$^{\circ}$, Yang Yang$^{\diamond}$, Jie Tang$^{\dagger**}$} \\
  $^{\dagger}$ Department of Computer Science and Technology, Tsinghua University \\
  $^{\ddagger}$ Microsoft Research, Redmond $^{\circ}$ Fudan University $^{\diamond}$ Zhejiang University \\
  \texttt{\{qinkai, jietang\}@tsinghua.edu.cn} \\
  \texttt{\{zoux18, cyk20, yd18\}@mails.tsinghua.edu.cn} \\
  \texttt{ericdongyx@gmail.com},
  \texttt{jiarongxu@fudan.edu.cn},
  \texttt{yangya@zju.edu.cn}
}

\begin{document}

\maketitle

\renewcommand{\thefootnote}{\fnsymbol{footnote}}
\footnotetext[1]{Now at Meta AI, Seattle and work done when working at Microsoft Research, Redmond.}
\footnotetext[7]{Jie Tang is the corresponding author.}

\begin{abstract}
Adversarial attacks on graphs have posed a major threat to the robustness of graph machine learning (GML) models. Naturally, there is an ever-escalating arms race between attackers and defenders. However, the strategies behind both sides are often not fairly compared under the same and realistic conditions.
To bridge this gap, we present the Graph Robustness Benchmark (\name) with the goal of providing a \textit{scalable}, \textit{unified}, \textit{modular}, and \textit{reproducible} evaluation for the adversarial robustness of GML models. 
\name standardizes the process of attacks and defenses by 1) developing scalable and diverse datasets, 2) modularizing the attack and defense implementations, and 3) unifying the evaluation protocol in refined scenarios. 
By leveraging the \name pipeline, the end-users can focus on the development of robust GML models with automated data processing and experimental evaluations. 
To support open and reproducible research on graph adversarial learning, \name also hosts public leaderboards across different scenarios. 
As a starting point, we conduct extensive experiments to benchmark baseline techniques. 
\name is open-source and welcomes contributions from the community. Datasets, codes, leaderboards are available at \textcolor{blue}{\url{https://cogdl.ai/grb/home}}.
\end{abstract}

\section{Introduction}\label{sec:intro}
% \outline{Background of GNNs and adversarial attacks.}

Graph machine learning (GML) models, from network embedding~\cite{perozzi2014deepwalk, grover2016node2vec, qiu2018network} to graph neural networks (GNNs)~\cite{kipf2016semi, hamilton2017inductive, velivckovic2018graph, xu2018powerful, klicpera2018predict, wu2019simplifying}, have shown promising performance in various domains, such as social network analysis~\cite{perozzi2014deepwalk}, molecular graphs~\cite{hamilton2017inductive}, and recommender systems~\cite{ying2018graph}. 
However, GML models are known to be vulnerable to adversarial attacks~\cite{chen2018fast, zugner2018adversarial, zugner2019adversarial, ma2019attacking, sun2020adversarial, wang2020scalable, zheng2020kdd, zou2021effective}. 
Attackers can modify the original graph by adding or removing edges~\cite{chen2018fast, waniek2018hiding, du2017topology}, perturbing node attributes~\cite{zugner2018adversarial, zugner2019adversarial, ma2019attacking, sun2020adversarial}, or injecting malicious nodes~\cite{wang2020scalable, zheng2020kdd, zou2021effective} to conduct adversarial attacks. 
Despite the relatively minor changes to the graph, the performance of GML models can be impacted dramatically.

Threatened by adversarial attacks, a line of attempts have been made to have robust GML models. 
For example, recent GNN architectures such as RobustGCN~\cite{zhu2019robust}, GRAND~\cite{feng2020graph}, and ProGNN~\cite{jin2020graph} are designed to improve the adversarial robustness of GNNs. 
In addition, pre-processing based methods, such as GNN-SVD~\cite{entezari2020all} and GNNGuard~\cite{zhang2020gnnguard}, alleviate the impact of attacks by leveraging the intrinsic graph properties and thus improve the model robustness.
Despite various efforts in this direction, there are several common limitations from both the attacker and the defender sides:
\setlist[enumerate]{leftmargin=5mm}
\begin{enumerate}
    \item \textbf{Unrealistic Attack/Defense Scenarios.} The existing attack and defense setups are often ambiguously defined with unrealistic assumptions, such as ignoring the real-world capabilities of attackers and defenders, resulting in less practical applications. 
    \item \textbf{Lack of A Unified Evaluation Protocol.} Previous works often use different settings (\emph{e.g.}, datasets, data splittings, attack constraints) in their experiments, resulting in biases in the evaluation and thus making it difficult to fairly compare different methods.
	\item \textbf{Lack of Scalability.} Most existing attacks and defenses are performed on very small-scale graphs (\emph{e.g.}, $<$10,000 nodes) without considering different levels of attack/defense difficulties, which are far from the scale and complexity of real-world applications. 
\end{enumerate}
To date, there exist several well-established GML benchmarks. 
For example, the Open Graph Benchmark (OGB)~\cite{hu2020open} offers abundant datasets and a unified evaluation pipeline for GML. 
Benchmarking GNNs~\cite{dwivedi2020benchmarking} is a standardized benchmark with consistent experimental settings. 
However, they mainly focus on evaluating the performance of GML models, regardless of their robustness. 
DeepRobust~\cite{li2020deeprobust} is a toolkit with implementations of attacks and defenses on both image and graph data, which by design is not a GML benchmark. 
Therefore, to address the aforementioned limitations, there is an urgent need for public benchmarks on evaluating the \textit{adversarial robustness} of GML models.

\begin{figure*}[t]
\centering
    \includegraphics[width=\textwidth]{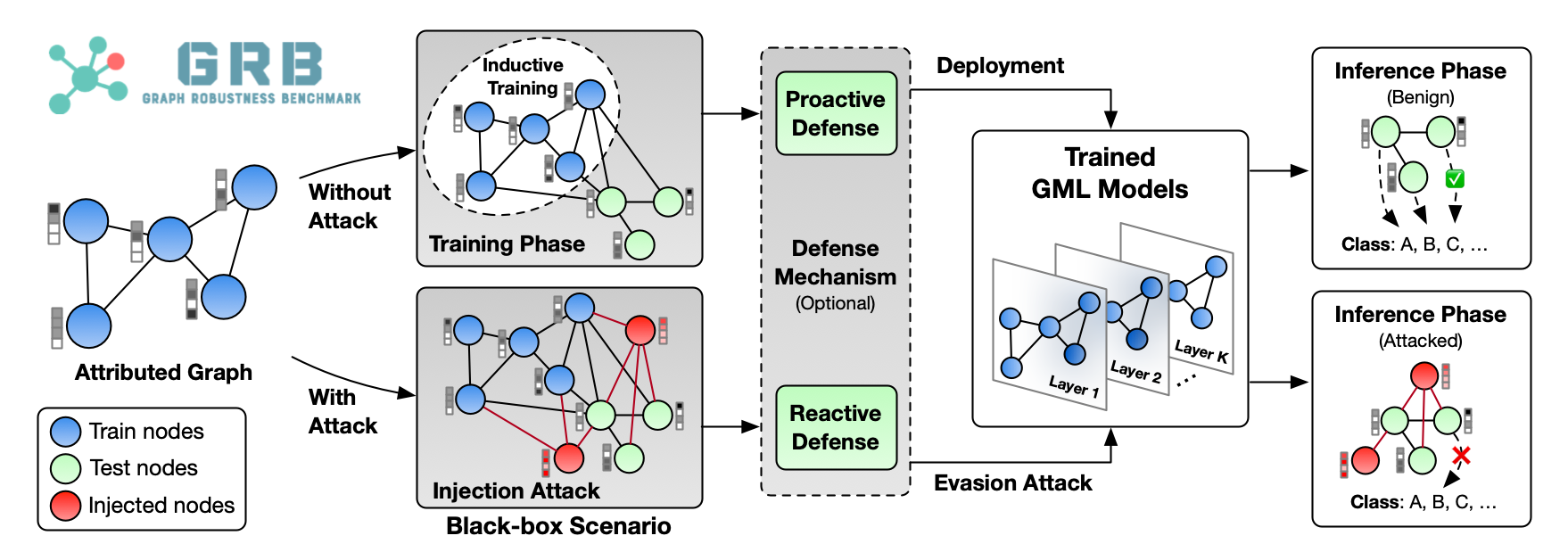}
    \caption{An example of \name's attack vs.\ defense (\textit{graph injection}) scenario:  \textit{Black-box}: attackers only have access to the attributed graph but not the target models; \textit{Inductive}: target models are trained in an inductive setting (test nodes are unseen during training); \textit{Injection}: attackers are allowed to inject new nodes without modifying the existing ones; \textit{Evasion}: attacks happen during model inference. All attacks and defenses are evaluated under unified settings to be fairly compared.}
    \label{fig:grb_process}
    \vspace{-0.5cm}
\end{figure*}

% \outline{Key features or strengths of \name.}

In this paper, we propose the Graph Robustness Benchmark (\name)---the first attempt to benchmark the adversarial robustness of GML models. 
The goal of \name is to provide a reproducible framework that enables a fair evaluation for both adversarial attacks \& defenses on GML models under unified settings. 
To achieve this, \name is designed to have the following properties:

\setlist[enumerate]{leftmargin=5mm}
\begin{enumerate}
    \item \textbf{Refined Attack/Defense Scenarios.} \name includes two refined attack scenarios: \textit{graph modification} and \textit{graph injection}, covering the majority of  works in the field. 
    By revisiting the limitations of previous works, we formalize precise definitions for both attackers' and defenders' capabilities, including available information to use and allowed actions, forming more realistic evaluations. 
   	\item \textbf{Scalable and Unified Evaluations.} \name contains various datasets of different orders of magnitude in size, with a specific robustness-focused splitting scheme for various levels of attacking/defending difficulties. It also provides a unified evaluation pipeline that calibrates all experimental settings, enabling fair comparisons for both attacks and defenses. 
    \item \textbf{Reproducible and Public Leaderboards.} \name offers a modular code framework\footnote{\textcolor{blue}{\url{https://github.com/THUDM/grb}}} that supports the implementations of a diverse set of baseline methods covering GML models, attacks, and defenses. 
    Additionally, it hosts public leaderboards across all evaluation scenarios, which will be continuously updated to track the progress in this community.
\end{enumerate}

Overall, \name serves as a \textit{scalable}, \textit{unified}, \textit{modular}, and \textit{reproducible} benchmark on evaluating the adversarial robustness of GML models. 
It is designed to facilitate the robust developments of graph adversarial learning, summarizing existing progress, and generating insights into future research.

\section{Adversarial Robustness in Graph Machine Learning}\label{sec:revisit}
\subsection{Problem Definition}

In graph machine learning, adversarial robustness refers to the ability of GML models to maintain their performance under potential adversarial attacks. 
Take the task of node classification as an instance, for an undirected attributed graph $\mathcal{G}=(\mathcal{A}, \mathcal{F})$ where $\mathcal{A}\in\mathbb{R}^{N\times N}$ represents the adjacency matrix of $N$ nodes and $\mathcal{F}\in\mathbb{R}^{N\times D}$ denotes the set of node features with $D$ dimensions. 
Define a GML model $\mathcal{M}: \mathcal{G}\rightarrow \mathcal{Z}$ where $\mathcal{Z}\in{[0, 1]}^{N\times L}$, which maps a graph $\mathcal{G}$ to probability vectors with $L$ classes. 
Generally, the objective of adversarial attacks on GML models can be formulated as:
\begin{equation}
    \label{eq:attack}
    \max_{\mathcal{G}'} |\mathop{\arg\max}_{l\in[1, ..., L]}\mathcal{M}(\mathcal{G}') \neq \mathop{\arg\max}_{l\in[1, ..., L]}\mathcal{M}(\mathcal{G})| \ \textrm{s.t.} \ d_{\mathcal{A}}(\mathcal{A}', \mathcal{A})\leq \Delta_{\mathcal{A}} \ \textrm{and} \ d_{\mathcal{F}}(\mathcal{F}', \mathcal{F})\leq \Delta_{\mathcal{F}}
\end{equation}
where $\mathcal{G'}=(\mathcal{A'}, \mathcal{F'})$ is the attacked graph, and $d_{\mathcal{A}}$ and $d_{\mathcal{F}}$ are distance metrics in the metric space $(\mathcal{A}, d_{\mathcal{A}})$ and $(\mathcal{F}, d_{\mathcal{F}})$. 
The attacker tries to maximize the number of incorrect predictions by GML models, under the constraints $\Delta_{\mathcal{A}}$ and $\Delta_{\mathcal{F}}$. 
For instance, $\Delta_{\mathcal{A}}$ can be the limited number of modified edges and $\Delta_{\mathcal{F}}$ can be the limited range of modified features (\emph{Cf.} Section~\ref{sec:overview} for detailed discussions).

\subsection{Revisiting Adversarial Attacks and Defenses in GML} 
\label{sec:attack}

\begin{wraptable}{r}{0.5\textwidth}
\vspace{-0.4cm}
\begin{minipage}{0.49\textwidth}
\centering
\caption{A categorization of graph adversarial attacks. There are mainly two scenarios: \textit{graph modification} and \textit{graph injection}. \name supports the implementation of all types of methods. \textsuperscript{$\dagger$}}
\scalebox{0.55}{
\begin{tabular}{ccccccccc} 
\toprule
\textbf{Adversarial} & \multicolumn{2}{c}{\textbf{Knowledge}} & \multicolumn{2}{c}{\textbf{Objective}} & \multicolumn{2}{c}{\textbf{Approach}} & \multirow{2}{*}{\textbf{Scalability}} \\ 
\textbf{Attack} & \textit{Black.} & \textit{White.} & \textit{Poi.} & \textit{Eva.} & \textit{Mod.} & \textit{Inj.} & \\ 
\midrule
\textbf{DICE}~\cite{waniek2018hiding} & \CheckmarkBold & -- & \CheckmarkBold & -- & \CheckmarkBold & -- & \CheckmarkBold \\
\textbf{FGA}~\cite{chen2018fast} & \CheckmarkBold & -- &  \CheckmarkBold & -- & \CheckmarkBold & -- & \XSolidBrush \\
\textbf{FLIP}~\cite{bojchevski2019adversarial} & \CheckmarkBold & -- & \CheckmarkBold & -- & \CheckmarkBold & -- & \CheckmarkBold \\
\textbf{NEA}~\cite{bojchevski2019adversarial} & \CheckmarkBold & -- & \CheckmarkBold & -- & \CheckmarkBold & -- & \XSolidBrush \\
\textbf{FGSM}~\cite{zugner2018adversarial} & \CheckmarkBold & \CheckmarkBold & \CheckmarkBold & -- & \CheckmarkBold & -- & \CheckmarkBold \\
\textbf{Nettack}~\cite{zugner2018adversarial} & \CheckmarkBold & \CheckmarkBold & \CheckmarkBold & -- & \CheckmarkBold & -- & \XSolidBrush \\
\textbf{RL-S2V}~\cite{dai2018adversarial} & \CheckmarkBold & \CheckmarkBold & \CheckmarkBold & -- & \CheckmarkBold & -- & \XSolidBrush \\
\textbf{Metattack}~\cite{zugner2019adversarial} & \CheckmarkBold & -- & \CheckmarkBold & -- & \CheckmarkBold & -- & \XSolidBrush \\
\textbf{STACK}~\cite{xu2020query} & \CheckmarkBold & -- & \CheckmarkBold & -- & \CheckmarkBold & -- & \XSolidBrush \\
\textbf{AFGSM}~\cite{wang2020scalable} & \CheckmarkBold & -- & \CheckmarkBold & -- & -- & \CheckmarkBold & \CheckmarkBold \\
\textbf{SPEIT}~\cite{zheng2020kdd} & \CheckmarkBold & -- & -- & \CheckmarkBold & -- & \CheckmarkBold & \CheckmarkBold \\
\textbf{TDGIA}~\cite{zou2021effective} & \CheckmarkBold & -- & -- & \CheckmarkBold & -- & \CheckmarkBold & \CheckmarkBold \\ 
\midrule
\textbf{\name Mod. Scenario} & \textcolor{dkgreen}{\CheckmarkBold} & -- & -- & \textcolor{dkgreen}{\CheckmarkBold} & \textcolor{dkgreen}{\CheckmarkBold} & -- & \textcolor{dkgreen}{\CheckmarkBold} \\
\textbf{\name Inj. Scenario} & \textcolor{dkgreen}{\CheckmarkBold} & -- & -- & \textcolor{dkgreen}{\CheckmarkBold} & -- & \textcolor{dkgreen}{\CheckmarkBold} & \textcolor{dkgreen}{\CheckmarkBold} \\
\textbf{\name Support} & \textcolor{dkgreen}{\CheckmarkBold} & \textcolor{dkgreen}{\CheckmarkBold} & \textcolor{dkgreen}{\CheckmarkBold} & \textcolor{dkgreen}{\CheckmarkBold} & \textcolor{dkgreen}{\CheckmarkBold} & \textcolor{dkgreen}{\CheckmarkBold} & \textcolor{dkgreen}{\CheckmarkBold} \\
\bottomrule
\end{tabular}}
\label{tab:attack_summary}
\tiny{\textsuperscript{$\dagger$} The table represents the original settings, while methods can be adapted to other settings by using \name's modualr coding framework.}
\vspace{-0.3cm}
\end{minipage}
\end{wraptable}

In the work of Szegedy \emph{et al.}~\cite{szegedy2013intriguing}, the existence of adversarial examples was revealed for ML models in image classification---imperceptible perturbations on inputs have ineligible impact on outputs of models. 
Recent works (in~\tablename~\ref{tab:attack_summary}) show that GML models are no exception. 
Graph adversarial attacks can mainly be categorized into two types according to the attack approach: \textit{graph modification} attack and \textit{graph injection} attack. 
Graph modification attacks directly modify the existing graph, by adding or removing edges (\emph{e.g.}, DICE~\cite{waniek2018hiding}, FGA~\cite{chen2018fast}, FLIP~\cite{bojchevski2019adversarial}, NEA~\cite{bojchevski2019adversarial}, STACK~\cite{xu2020query}), or further modifying node features (\emph{e.g.}, Nettack~\cite{zugner2018adversarial}, FGSM~\cite{zugner2018adversarial}, RL-S2V~\cite{dai2018adversarial}, Metattack~\cite{zugner2019adversarial}). 
Differently, graph injection attacks add new malicious nodes without modifying the original graph (\emph{e.g.}, AFGSM~\cite{wang2020scalable}, SPEIT~\cite{zheng2020kdd}, TDGIA~\cite{zou2021effective}). 
Facing the problem of scalability, some attacks are not applicable to large graphs due to their high time complexity~\cite{zugner2018adversarial, zugner2019adversarial, dai2018adversarial} or expensive memory consumption~\cite{chen2018fast, bojchevski2019adversarial}. % (requiring the use of dense adjacency matrix rather than the sparse alternative). 

Defenses can mainly be categorized into two types: \textit{preprocess-based} defense and \textit{model-based} defense. 
The first type regards the attacked graphs as noisy ones and defenders can preprocess the adjacency matrix (\emph{e.g.}, GNN-SVD~\cite{entezari2020all}, GNN-Jaccard~\cite{wu2019adversarial}) or the features of nodes (\emph{e.g.}, feature transformation~\cite{zheng2020kdd}), to alleviate the effect of perturbations. 
The second type achieves robustness through \textit{model enhancement}, either by robust training schemes (\emph{e.g.}, adversarial training~\cite{madry2017towards, feng2019graph}) or new model architectures (\emph{e.g.}, RobustGCN~\cite{zhu2019robust}, GNNGuard~\cite{zhang2020gnnguard}). 
Some defenses also suffer from the problem of scalability, due to the need of calculation on large dense matrices~\cite{entezari2020all, wu2019adversarial, zhang2020gnnguard}. 

Notwithstanding the significant progress, existing works share some common limitations: 
(1) {Lack of scalability}: Most works only consider very small graphs and cannot be scaled up to larger ones due to time/memory complexity. 
(2) {Lack of generalization}: Most attacks/defenses are evaluated on very basic GML models, but not on other variants. Meanwhile, some methods are only effective for specific models with ad-hoc designs, which makes the results less generalized and practical. 
(3) {Ill-defined scenarios}: The scenarios and assumptions proposed in some previous works are not realistic, \emph{e.g.}, the \textit{unnoticeability} under \textit{poisoning} setting ignores the real capability of the defenders (\emph{Cf.} \appendixname~\ref{sec:unnoticeablilty} for details). 
Besides, there are no unified standards on evaluating the adversarial robustness. Different settings (\emph{e.g.}, the choice of datasets, random splitting, different constraints) introduce biases, which makes it hard to compare the effectiveness of different methods. 
In light of these challenges, there is an urgent need for benchmarking the adversarial robustness of GML.

\section{\name: Graph Robustness Benchmark}\label{sec:overview}
\subsection{Overview of \name}

\begin{wrapfigure}{r}{0.5\textwidth}
\vspace{-0.4cm}
\centering
\includegraphics[width=0.5\textwidth]{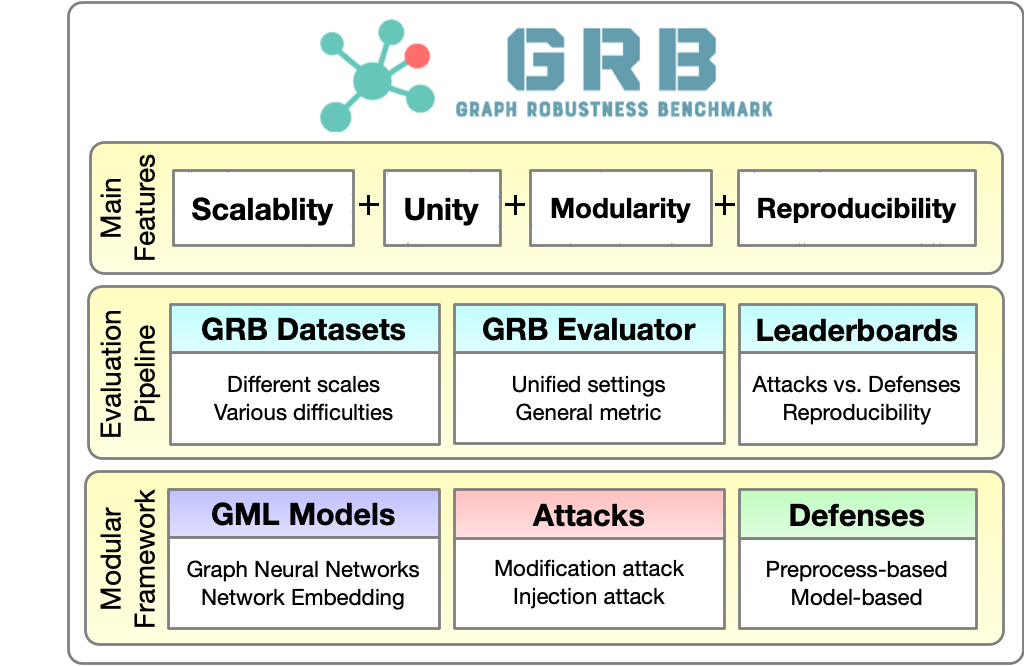}
\caption{\name Framework.}
\label{fig:grb_framework}
\vspace{-0.3cm}
\end{wrapfigure}

To overcome the limitations of previous works, we propose the Graph Robustness Benchmark (\name)---a standardized benchmark for evaluating the adversarial robustness of GML. 
To ensure \name's scalability, we include datasets of different sizes with scalable attack/defense baselines. 
To have a unified process, we standardize the evaluation scenarios with  precise constraints and realistic assumptions on attackers and defenders. 
To make \name easy-to-use, we provide a modular pipeline that facilitates the implementation of GML models, attacks, and defenses. 
To guarantee the reproducibility, we open-source and maintain the \name public leaderboards that are continuously updated to track the progress of the community.

Altogether, \name serves as a \textit{scalable}, \textit{unified}, \textit{modular}, \textit{reproducible} benchmark on evaluating the adversarial robustness of GML models. 
We present the solutions to achieve these goals for \name. 
%In the following, we introduce in detail \name's unified evaluation scenarios, modular coding framework, baseline methods and datasets. 

\subsection{The Unified Evaluation Scenario of GML Adversarial Robustness}

To evaluate the adversarial robustness, it is essential to be aware of the capabilities of potential attackers. 
We categorize attacks into the following aspects (as shown in~\tablename~\ref{tab:attack_summary}):
\setlist[enumerate]{leftmargin=5mm}
\begin{enumerate}
	\item \textbf{Knowledge.} \textit{Black-box}: Attackers do NOT have access to the targeted model (including its architecture, parameters, defense mechanism, etc.). However, they can access the graph data (structure, features, labels of training data, etc.). Additionally, they have limited chances to query the model to get outputs. \textit{White-box}: Attackers have access to {ALL} information. However, if the targeted model has a random process, the run-time randomness is still preserved. 
	\item \textbf{Objective.} \textit{Poisoning}: Attackers generate corrupted graph data and assume that the targeted model is (re)trained on these data to get a worse model. \textit{Evasion}: The target model has already been trained, and attackers can generate corrupted graph data to affect its inference. 
	\item \textbf{Approach.} \textit{Modification}: Attackers modify the original graph (the same one used by defenders for training) by adding/removing edges or perturbing node features. \textit{Injection}: Attackers do not modify the original graph but inject new malicious nodes to influence a set of targeted nodes. 
\end{enumerate}  

In practice, the most common real-world case is that the GML models have already been trained for specific tasks and  deployed in a secret way, i.e., \textit{black-box} and \textit{evasion} settings. Thus, in \name, we propose two unified evaluation scenarios under these settings, \textit{graph modification} and \textit{graph injection}. 
						
\textbf{Graph Modification.} 
This has been the most studied scenario, in which attackers can directly modify the graph (by adding/removing edges or perturbing node attributes) to attack the GML models. 
Under real-world conditions, this is theoretically possible but practically difficult, as the modification attacks require the authority to access the target nodes in order to to change their contents.
Nevertheless, this scenario enables us to understand how the GML models behave under intended modifications.

\textbf{Graph Injection.} 
This scenario was first introduced in the KDDCUP 2020 task of Graph Adversarial Attacks \& Defenses\footnote{\textcolor{blue}{\url{https://www.biendata.xyz/competition/kddcup\_2020\_formal/}}}, which targeted at injecting new nodes to a large-scale academic graph. 
It is more realistic than the modification one since injecting new nodes is more practically possible than modifying the existing ones. 
However, the task in KDDCUP 2020 considers a transductive setting, \emph{i.e.}, test nodes (except for their labels) are available during training. 
In this case, defenders can simply memorize benign nodes and identify the injected nodes, making it an imperfect setting. 

Thus, to further \name's practical usage (\emph{Cf.} ~\appendixname~\ref{sec:unnoticeablilty} for detailed discussions), we make the following assumptions for both scenarios: 
(1) {Black-box}: Both attackers and defenders do \textit{not} have knowledge about the methods each other applied.
(2) {Inductive}: The GML models are trained in trusted data and  used to classify unseen data (\emph{e.g.}, new users), \emph{i.e.}, the validation and test data is unseen during training.
(3) {Evasion}: Attacks will only happen during the inference phase. 
Furthermore, we clarify attackers' and defenders' capabilities in \name: 
\setlist[enumerate]{leftmargin=5mm}
\begin{enumerate}
    \item {For attackers}:
    (a) They have knowledge about the entire graph (including all nodes, edges and labels but {excluding} the labels of the test nodes), but do not have knowledge about the target model or defense mechanism.
    (b) For graph modification, following the most common setting in previous works, attackers are allowed to perturb a limited number of edges in the graph ($\Delta_{\mathcal{A}}$: the number of modified edges less than a ratio $\gamma_e$ of all edges).
    (c) For graph injection, we follow the heuristic setting of KDDCUP 2020, attackers are allowed to inject new nodes with limited edges ($\Delta_{\mathcal{A}}$: less than $N_{n}$ injected nodes each with less than $N_{e}$ edges; $\Delta_{\mathcal{F}}$: constrained range of features $[\mathcal{F}_{min}, \mathcal{F}_{max}]$.).
    (d) They are not allowed to modify the original graph for training.
    (e) They are allowed to get predictions from the target model through a limited number of queries. 
    
    \item {For defenders}: 
    (a) They have knowledge about the graph {excluding} the test nodes %(features, structure, and labels) 
    to be attacked. 
    (b) They are allowed to use any method to increase the adversarial robustness, but do not have prior knowledge about the edges/nodes that are modified/injected.
    
    \item {For both sides}: 
    Attackers/defenders can of course make assumptions even in the {black-box} scenario. 
    For instance, attackers can assume that the target system deploys a certain type of GML models, then it can be used as the surrogate model to conduct transfer attacks.
    Moreover, it is not reasonable to assume that the defense mechanism can be completely held secretly, known as the Kerckhoffs’ principle~\cite{kerckhoffs1883cryptographie}. 
    If a defense wants to be general and universal, it should guarantee part of the robustness even when attackers have some knowledge about it. 
    In \name, we evaluate an attack vs. multiple defenses (vice versa), thus the assumptions can hardly violate the \textit{black-box} conditions. 
    As a result, the objective for both sides is to be generally effective against all potential methods rather than just a single one. 
\end{enumerate}

By following the above rules, we provide unified evaluation scenarios for attacks and defenses in a principled way. It is worth noting that these unified scenarios are not the only valid ones, 
\name will include more scenarios as this field eveloves over time.

\subsection{The Modular \name Pipeline}

\name offers a modular pipeline, which is based on PyTorch~\cite{paszke2019pytorch} as well as other popular GML libraries like CogDL~\cite{cen2021cogdl} and DGL~\cite{wang2019deep}. 
Specifically, it contains the following modules: 
(1) {Dataset}: \name provides data-loaders for \name datasets and applies necessary preprocessing including splitting and feature normalization; it also supports external datasets like OGB~\cite{hu2020open} or user-defined datasets. %, and make them compatible with the \name evaluation pipeline. 
(2) {Model}: The GML models are implemented based on PyTorch, CogDL, and DGL and \name can automatically transform inputs to compatible formats. 
(3) {Attack}: We implement adversarial attacks by abstracting the attack process to different components, e.g., graph injection attacks are decomposed to node injection and feature generation. 
(4) {Defense}: \name engages defense mechanisms to GML models, including \textit{preprocess-based} and \textit{model-based} ones. 
(5) {Evaluator}: The attack or defense methods are evaluated under unified settings and metrics. 
Essentially, \name unifies and modularizes the entire process, including loading datasets, training/loading models, applying attacks/defenses, and generating the evaluation results; it also helps to reproduce the exact results on \name leaderboards. 
In addition to these modules, \name also offers other functions including \textit{Trainer} for model training, \textit{AutoML} for automatic parameter search, and \textit{Visualise} for visualizing the attack process. 

\begin{figure}[!htp]
\centering
    \includegraphics[width=\textwidth]{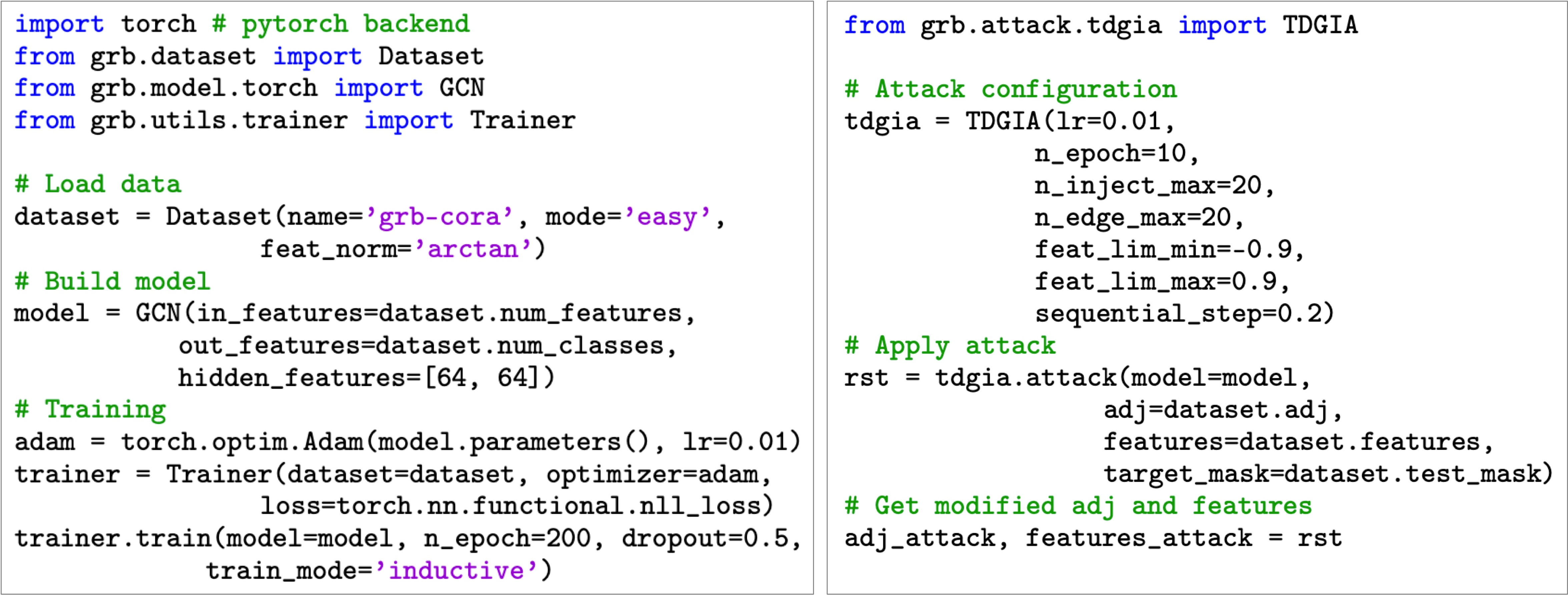}
    \caption{\name usage examples. {Left}: Train GCNs on the \textit{grb-cora} dataset. {Right}: Apply the TDGIA attack on the trained model. \name facilitates the usage of GML models, attacks,and defenses.}
    \label{fig:code_example}
    \vspace{-0.1in}
\end{figure}

The \name  framework has the following features: 
(1) {Easy-to-use}: the baseline methods are easy to use by only a few lines of codes, as shown in~\figurename~\ref{fig:code_example}. 
(2) {Fair-to-compare}: all methods are fairly compared under unified settings. 
(3) {Up-to-date}: the leaderboards for each dataset are maintained to continuously track the progress in the domain. 
(4) {Reproducible}: %unlike some benchmarks that just display results, 
\name prioritizes reproducibility. All necessary materials are made public to reproduce results on leaderboards, including the trained models, generated attack results, etc. 
Users can reproduce results by a single command line (\emph{Cf.} Appendix~\ref{app:reproducible} for \name reproducibility rules). 
All codes are available in \textcolor{blue}{\url{https://github.com/THUDM/grb}}, where the implementation details and examples can be also found. The API documentations are covered in \textcolor{blue}{\url{https://grb.readthedocs.io/en/latest/}}.

\subsection{The \name Baselines}

Currently, \name covers a rich set of baselines for the GML models, attacks, and defenses. 

\textbf{Seven GML models}: GCN~\cite{kipf2016semi}, GAT~\cite{velivckovic2018graph}, GIN~\cite{xu2018powerful}, APPNP~\cite{klicpera2018predict}, TAGCN~\cite{du2017topology}, GraphSAGE~\cite{hamilton2017inductive}, SGCN~\cite{wu2019simplifying}. Note that these models are not originally designed to increase robustness. 

\textbf{Twelve Attacks}: Seven modification attacks---RND~\cite{zugner2018adversarial}, DICE~\cite{waniek2018hiding}, FGA~\cite{chen2018fast}, FLIP~\cite{bojchevski2019adversarial}, NEA~\cite{bojchevski2019adversarial}, STACK~\cite{xu2020query}, and PGD~\cite{madry2017towards}---and five injection attacks---RND, FGSM~\cite{goodfellow2014explaining}, PGD~\cite{madry2017towards}, SPEIT~\cite{zheng2020kdd}, and TDGIA~\cite{zou2021effective}. %We adapt all attacks to the proposed evaluation scenarios.  
More details can be found in~\appendixname~\ref{app:attack}.

\textbf{Five Defenses}: \name adopts RobustGCN (R-GCN)~\cite{zhu2019robust}, GNN-SVD~\cite{entezari2020all}, and GNNGuard~\cite{zhang2020gnnguard}. 
Additionally, we find that techniques like layer normalization (LN)~\cite{ba2016layer} and adversarial training (AT)~\cite{madry2017towards}, if properly used in the proposed evaluation scenarios, can significantly increase the robustness of various GML models. 
The LN can be applied on the input features and after each graph convolutional layer (except for the last one). The idea is to stabilize the dynamics of input and hidden states to alleviate the impact of adversarial perturbations. 
The AT uses modification/injection attacks during training to make GML models more robust. Note that most of previous works only use AT to perturb the existing graph, however, we find that AT also works well by injecting new nodes during training. 
These two defenses are general and scalable, and the experiment results show that they outperform previous dedicated methods. 
Thus, we include them in \name as strong baselines for defenses. More details can be found in~\appendixname~\ref{app:defense}. 

\subsection{The \name Datasets}

\begin{table}[!ht]
\centering
\vspace{-0.1in}
\caption{Statistics of five \name datasets covering from small- to large-scale graphs.}
\label{tab:dataset_stats}
\resizebox{\textwidth}{15mm}{
\begin{tabular}{lcrrrrrr} 
\toprule
\multicolumn{1}{c}{\textbf{Dataset}} & \textbf{Scale} & \multicolumn{1}{c}{\textbf{\#Nodes}} & \multicolumn{1}{c}{\textbf{\#Edges}} & \multicolumn{1}{c}{\textbf{\#Feat.}} & \multicolumn{1}{c}{\textbf{\#Classes}} & \multicolumn{1}{c}{\begin{tabular}[c]{@{}c@{}}\textbf{Feat. Range}\\\textbf{(original)}\end{tabular}} & \multicolumn{1}{c}{\begin{tabular}[c]{@{}c@{}}\textbf{Feat. Range}\\\textbf{(normalized)}\end{tabular}} \\ 
\midrule
\textit{grb-cora} & Small & 2,680 & 5,148 & 302 & 7 & {[}-2.30, 2.40] & {[}-0.94, 0.94] \\
\textit{grb-citeseer} & Small & 3,191 & 4,172 & 768 & 6 & {[}-4.55, 1.67] & {[}-0.96, 0.89] \\
\textit{grb-flickr} & Medium & 89,250 & 449,878 & 500 & 7 & {[}-0.90, 269.96] & {[}-0.47, 1.00] \\
\textit{grb-reddit} & Large & 232,965 & 11,606,919 & 602 & 41 & {[}-28.19, 120.96] & {[}-0.98, 0.99] \\
\textit{grb-aminer} & Large & 659,574 & 2,878,577 & 100 & 18 & {[}-1.74, 1.62] & {[}-0.93, 0.93] \\
\bottomrule
\end{tabular}}
\vspace{-0.1in}
\end{table}

\textbf{Scalability.} \name includes five datasets of different scales, \textit{grb-cora}, \textit{grb-citeseer}, \textit{grb-flickr}, \textit{grb-reddit}, and \textit{grb-aminer}. 
The original datasets are  gathered from previous works~\cite{zou2019dimensional, zeng2019graphsaint, zou2021effective} and  are reprocessed for \name.
The basic statistics of these datasets are shown in~\tablename~\ref{tab:dataset_stats}. 
%Besides small-scale datasets that are common in previous works, \name also includes medium and large-scale datasets with hundreds of thousands of nodes and millions of edges.
More details about datasets can be found in Appendix~\ref{app:dataset}.

\begin{figure}[htbp]
\centering
\includegraphics[width=\textwidth]{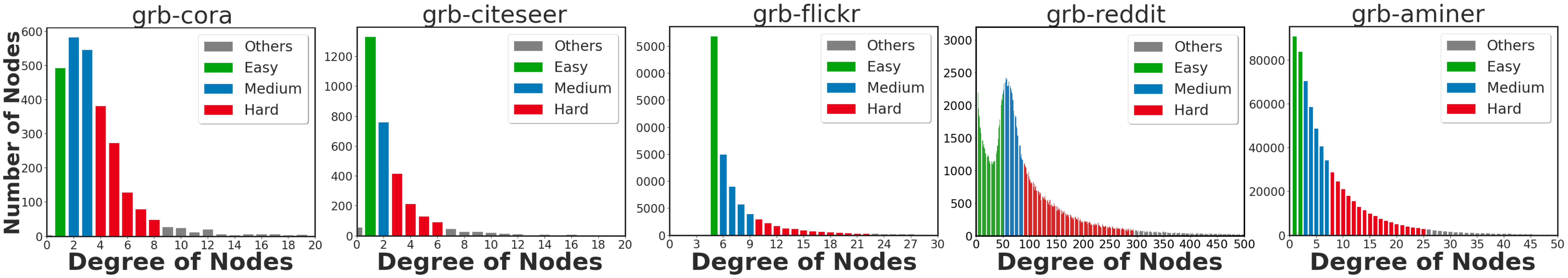}
\caption{\name's splitting scheme. Difficulties are related to the average degree of test nodes.}
\label{fig:splitting}
\end{figure}

\textbf{Data Splitting.}
\name introduces a new splitting data scheme  designed for evaluating the GML adversarial robustness.
Its key idea is based on the assumption that nodes with lower degrees are easier to attack, as demonstrated in~\cite{zou2021effective}. 
If a target node has few neighbors, it is more likely to be influenced by adversarial perturbations aggregated from its neighbors. 
Thus, we construct test subsets with different average degrees to represent different difficulties. 
First, we rank all nodes by their degrees. 
Second, we filter out 5\% nodes with the lowest degrees (\emph{e.g.}, isolated nodes that are too easy to attack) and 5\% nodes with the highest degrees (\emph{e.g.}, nodes connected to hundreds of other nodes that are too hard to attack). 
Third, we divide the rest of nodes into three equal partitions without overlapping, and randomly sample 10\% nodes (without repetition) from each partition. 
Finally, we get three test subsets with different degree distributions as shown in~\figurename~\ref{fig:splitting}, which are defined as Easy/Medium/Hard/Full (`E/M/H/F') with `F' containing all test nodes. 
For the rest of nodes, we divide them into the training set (60\%) and validation set (10\%).

\textbf{Feature Normalization.} 
Initially, the features in each dataset have various ranges.
To unify their constraints and to have values in the same scale (\emph{e.g.}, range $[-1, 1]$), we apply a {standardization} followed by an {arctan} transformation: $\mathcal{F} = \frac{2}{\pi} \arctan(\frac{\mathcal{F} - \tiny{\textrm{mean}}(\mathcal{F})}{\tiny{\textrm{std}}(\mathcal{F})})$. The statistics of datasets after the splitting scheme and the feature normalization can be found in~\appendixname~\ref{app:dataset}.

\section{Experiments}\label{sec:exp}
With the support of \name's modular framework, we conduct extensively experiments to evaluate the adversarial robustness of GML models under the unified evaluation protocol, from which insights are generated into the developments of the field.

\subsection{Experimental Settings}
\textbf{Baselines.} (1) For GML models, we include 7 baselines: GCN~\cite{kipf2016semi}, GAT~\cite{velivckovic2018graph}, GIN~\cite{xu2018powerful}, APPNP~\cite{klicpera2018predict}, TAGCN~\cite{du2017topology}, GraphSAGE~\cite{hamilton2017inductive}, SGCN~\cite{wu2019simplifying}. All models are salable to large graphs.  
(2) For modification attacks, we include 7 baselines: RND, DICE~\cite{waniek2018hiding}, FGA~\cite{chen2018fast}, FLIP~\cite{bojchevski2019adversarial}, NEA~\cite{bojchevski2019adversarial}, STACK~\cite{xu2020query}, and PGD~\cite{madry2017towards}, among which RND, DICE, FLIP, and PGD are scalable to large graphs. FGA, NEA, and PGD need to train a surrogate model to conduct transfer attacks.
(3) For injection attacks, we include 5 baselines: RND, FGSM~\cite{goodfellow2014explaining}, PGD~\cite{madry2017towards}, SPEIT~\cite{zheng2020kdd}, TDGIA~\cite{zou2021effective}. They are all scalable and FGSM, PGD, SPEIT, TDGIA need to train a surrogate model to conduct transfer attacks.
(4) For defenses, we include R-GCN~\cite{zhu2019robust}, GNN-SVD~\cite{entezari2020all}, GNNGuard~\cite{zhang2020gnnguard}. Among which only R-GCN is scalable, since the other two methods require calculation on dense adjacency matrix. Thus, we also adapt two general defense methods, layer normalization (LN)~\cite{ba2016layer} and adversarial training (AT)~\cite{madry2017towards} to the proposed scenarios. More details and hyper-parameter settings can be found in~\appendixname~\ref{app:method}~\ref{app:reproducible}.

\textbf{Evaluation Metrics.} For attacks: (1) {Avg.}: Average accuracy of all defenses (including vanilla GML models). (2) {Avg.\ 3-Max}: Average accuracy for the 3 most robust methods. (3) {Weighted}: Weighted accuracy, calculated by:$s_{w}^{\tiny{\textrm{ATK}}} = \sum_{i=1}^{n} w_i s_i, w_i = \frac{1/i^2}{\sum_{j=1}^n(1/j^2)}, s_i=(S_{descend}^{\tiny{\textrm{DEF}}})_i$ where $S_{descend}^{\tiny{\textrm{DEF}}}$ is the set of defense scores in a descending order. The metric attaches more weight to more robust methods. 
For defenses: (1) {Avg.}: Average accuracy of all attacks. (2) {Avg.\ 3-Min}: Average accuracy of the 3 most effective attacks. (3) {Weighted}: Weighted accuracy across various attacks, calculated by:$s_{w}^{\tiny{\textrm{DEF}}} = \sum_{i=1}^{n} w_i s_i, w_i = \frac{1/i^2}{\sum_{j=1}^n(1/j^2)}, s_i=(S_{ascend}^{\tiny{\textrm{ATK}}})_i$ where $S_{ascend}^{\tiny{\textrm{ATK}}}$ is the set of attack scores in an ascending order. The metric attaches more weight to more effective attacks. 

\subsection{Experimental Results}
We show an example of \name leaderboard, robust ranking of GML models, and various factors that affect the adversarial robustness in GML. 
More results can be found in Appendix and on our website.

\textbf{An Example of the \name Leaderboard.} 
Following the process in~\figurename~\ref{fig:grb_process}, we evaluate the performance of attacks vs.\ defenses in \textit{graph injection} scenario. 
\tablename~\ref{tab:leaderboard_aminer} shows an example of leaderboard for \textit{grb-aminer} dataset. 
Each attack is repeated 10 times to report the error bar. 
Both attacks and defenses are ranked by the weighted accuracy under 'F' difficulty, where {\color{red}red} and {\color{blue}blue} indicate the best results of attacks/defenses in each difficulty. 
Note that the metric is not fixed and will be updated when there are more effective methods. For instance, when there are more powerful attacks, the ranking will change so as the attached weights. It is reasonable that less effective attacks become less important on the final ranking of defenses, the same for defenses. 
As a result, \name leaderboard can indicate the most robust defenses and the most effective attacks. 

\begin{table}[!htp]
\vspace{-0.3cm}
\caption{\textit{grb-aminer} leaderboard (Top 5 ATK.\ vs.\ Top 10 DEF.) in \textit{graph injection} scenario.}
\centering
\scalebox{0.46}{
\begin{tabular}{cccccccccccccccc} 
\toprule
\multicolumn{3}{c}{\multirow{2}{*}{\diagbox{\textbf{Attacks}}{\textbf{Defenses}}}} & \textbf{1} & \textbf{2} & \textbf{3} & \textbf{4} & \textbf{5} & \textbf{6} & \textbf{7} & \textbf{8} & \textbf{9} & \textbf{10} & \multirow{2}{*}{\begin{tabular}[c]{@{}c@{}}\textbf{Avg.}\\\textbf{Accuracy~}\end{tabular}} & \multirow{2}{*}{\begin{tabular}[c]{@{}c@{}}\textbf{Avg. 3-Max}\\\textbf{Accuracy}\end{tabular}} & \multirow{2}{*}{\begin{tabular}[c]{@{}c@{}}\textbf{Weighted}\\\textbf{Accuracy}\end{tabular}} \\
\multicolumn{3}{c}{} & \textbf{GAT\tiny{+AT}} & \textbf{R-GCN\tiny{+AT}} & \textbf{SGCN\tiny{+LN}} & \textbf{R-GCN} & \textbf{GCN\tiny{+LN}} & \textbf{GAT\tiny{LN}} & \textbf{GIN\tiny{+LN}} & \textbf{TAGCN\tiny{+LN}} & \textbf{TAGCN\tiny{+AT}} & \textbf{GAT} &  &  &  \\
\midrule
\multirow{4}{*}{\textbf{1}} & \multirow{4}{*}{\textbf{TDGIA}} & E & 59.54\tiny{{$\pm$}0.05} & 56.83\tiny{{$\pm$}0.06} & 56.73\tiny{{$\pm$}0.06} & 56.12\tiny{{$\pm$}0.07} & 53.51\tiny{{$\pm$}0.21} & 43.93\tiny{{$\pm$}0.41} & 51.10\tiny{{$\pm$}0.12} & 54.63\tiny{{$\pm$}0.20} & 49.59\tiny{{$\pm$}0.50} & 42.40\tiny{{$\pm$}0.52} & {\color{red}52.44\tiny{{$\pm$}0.17}} & {\color{red}57.70\tiny{{$\pm$}1.31}} & {\color{red}58.08\tiny{{$\pm$}0.04}} \\
 &  & M & 68.39\tiny{{$\pm$}0.02} & 65.61\tiny{{$\pm$}0.02} & 66.11\tiny{{$\pm$}0.02} & 65.23\tiny{{$\pm$}0.03} & 66.78\tiny{{$\pm$}0.05} & 61.84\tiny{{$\pm$}1.20} & 64.49\tiny{{$\pm$}0.10} & 64.62\tiny{{$\pm$}0.02} & 67.27\tiny{{$\pm$}0.04} & 62.47\tiny{{$\pm$}1.01} & {\color{red}65.28\tiny{{$\pm$}0.23}} & {\color{red}67.48\tiny{{$\pm$}0.68}} & {\color{red}67.69\tiny{{$\pm$}0.02}} \\
 &  & H & 75.83\tiny{{$\pm$}0.02} & 72.35\tiny{{$\pm$}0.02} & 72.10\tiny{{$\pm$}0.00} & 71.94\tiny{{$\pm$}0.02} & 73.39\tiny{{$\pm$}0.02} & 75.22\tiny{{$\pm$}0.04} & 72.92\tiny{{$\pm$}0.02} & 68.94\tiny{{$\pm$}0.03} & 73.98\tiny{{$\pm$}0.01} & 75.03\tiny{{$\pm$}0.03} & 73.17\tiny{{$\pm$}0.01} & 75.36\tiny{{$\pm$}0.34} & {\color{red}75.33\tiny{{$\pm$}0.01}} \\
 &  & F & 67.69\tiny{{$\pm$}0.03} & 63.62\tiny{{$\pm$}0.32} & 62.20\tiny{{$\pm$}0.15} & 61.99\tiny{{$\pm$}0.22} & 60.38\tiny{{$\pm$}1.46} & 59.69\tiny{{$\pm$}1.57} & 59.59\tiny{{$\pm$}0.42} & 59.06\tiny{{$\pm$}1.75} & 57.24\tiny{{$\pm$}5.04} & 56.63\tiny{{$\pm$}6.75} & 60.81\tiny{{$\pm$}1.71} & {\color{red}64.52\tiny{{$\pm$}2.32}} & {\color{red}65.74\tiny{{$\pm$}0.21}} \\
 \midrule
\multirow{4}{*}{\textbf{2}} & \multirow{4}{*}{\textbf{SPEIT}} & E & 59.54\tiny{{$\pm$}0.07} & 56.80\tiny{{$\pm$}0.05} & 56.94\tiny{{$\pm$}0.10} & 55.64\tiny{{$\pm$}0.10} & 56.15\tiny{{$\pm$}0.06} & 56.13\tiny{{$\pm$}0.07} & 54.24\tiny{{$\pm$}0.09} & 56.61\tiny{{$\pm$}0.06} & 56.59\tiny{{$\pm$}0.08} & 57.36\tiny{{$\pm$}0.09} & 56.60\tiny{{$\pm$}0.04} & 57.95\tiny{{$\pm$}1.14} & 58.62\tiny{{$\pm$}0.05} \\
 &  & M & 68.37\tiny{{$\pm$}0.03} & 65.46\tiny{{$\pm$}0.03} & 66.20\tiny{{$\pm$}0.02} & 65.25\tiny{{$\pm$}0.05} & 66.75\tiny{{$\pm$}0.03} & 67.49\tiny{{$\pm$}0.06} & 65.05\tiny{{$\pm$}0.06} & 64.47\tiny{{$\pm$}0.04} & 66.95\tiny{{$\pm$}0.05} & 66.81\tiny{{$\pm$}0.04} & 66.28\tiny{{$\pm$}0.02} & 67.60\tiny{{$\pm$}0.59} & 67.86\tiny{{$\pm$}0.03} \\
 &  & H & 75.94\tiny{{$\pm$}0.04} & 72.27\tiny{{$\pm$}0.03} & 72.36\tiny{{$\pm$}0.03} & 71.86\tiny{{$\pm$}0.03} & 73.41\tiny{{$\pm$}0.01} & 75.34\tiny{{$\pm$}0.03} & 72.87\tiny{{$\pm$}0.03} & 68.88\tiny{{$\pm$}0.05} & 73.98\tiny{{$\pm$}0.02} & 73.83\tiny{{$\pm$}0.04} & {\color{red}73.07\tiny{{$\pm$}0.01}} & {\color{red}75.08\tiny{{$\pm$}0.82}} & 75.33\tiny{{$\pm$}0.02} \\
 &  & F & 68.04\tiny{{$\pm$}0.03} & 64.05\tiny{{$\pm$}0.04} & 64.84\tiny{{$\pm$}0.04} & 64.06\tiny{{$\pm$}0.04} & 65.51\tiny{{$\pm$}0.02} & 64.02\tiny{{$\pm$}0.04} & 63.11\tiny{{$\pm$}0.02} & 62.59\tiny{{$\pm$}0.04} & 63.77\tiny{{$\pm$}0.06} & 63.58\tiny{{$\pm$}0.06} & 64.36\tiny{{$\pm$}0.02} & 66.13\tiny{{$\pm$}1.38} & 66.89\tiny{{$\pm$}0.02} \\
 \midrule
\multirow{4}{*}{\textbf{3}} & \multirow{4}{*}{\textbf{RND}} & E & 59.56\tiny{{$\pm$}0.06} & 57.53\tiny{{$\pm$}0.06} & 57.41\tiny{{$\pm$}0.06} & 56.38\tiny{{$\pm$}0.11} & 57.76\tiny{{$\pm$}0.05} & 58.83\tiny{{$\pm$}0.10} & 54.41\tiny{{$\pm$}0.13} & 58.07\tiny{{$\pm$}0.12} & 58.14\tiny{{$\pm$}0.04} & 57.46\tiny{{$\pm$}0.10} & 57.55\tiny{{$\pm$}0.03} & 58.85\tiny{{$\pm$}0.57} & 59.09\tiny{{$\pm$}0.05} \\
 &  & M & 68.22\tiny{{$\pm$}0.04} & 65.86\tiny{{$\pm$}0.03} & 66.29\tiny{{$\pm$}0.03} & 65.34\tiny{{$\pm$}0.06} & 67.03\tiny{{$\pm$}0.03} & 68.62\tiny{{$\pm$}0.05} & 65.54\tiny{{$\pm$}0.06} & 64.98\tiny{{$\pm$}0.08} & 67.34\tiny{{$\pm$}0.04} & 67.71\tiny{{$\pm$}0.06} & 66.69\tiny{{$\pm$}0.02} & 68.18\tiny{{$\pm$}0.38} & 68.24\tiny{{$\pm$}0.03} \\
 &  & H & 75.75\tiny{{$\pm$}0.02} & 72.66\tiny{{$\pm$}0.02} & 72.42\tiny{{$\pm$}0.03} & 72.00\tiny{{$\pm$}0.03} & 73.52\tiny{{$\pm$}0.02} & 75.63\tiny{{$\pm$}0.03} & 73.36\tiny{{$\pm$}0.03} & 69.30\tiny{{$\pm$}0.06} & 74.04\tiny{{$\pm$}0.02} & 75.36\tiny{{$\pm$}0.03} & 73.40\tiny{{$\pm$}0.01} & 75.58\tiny{{$\pm$}0.17} & 75.39\tiny{{$\pm$}0.01} \\
 &  & F & 67.72\tiny{{$\pm$}0.04} & 64.98\tiny{{$\pm$}0.02} & 65.31\tiny{{$\pm$}0.04} & 64.45\tiny{{$\pm$}0.04} & 66.17\tiny{{$\pm$}0.02} & 67.54\tiny{{$\pm$}0.04} & 64.36\tiny{{$\pm$}0.06} & 64.33\tiny{{$\pm$}0.03} & 66.42\tiny{{$\pm$}0.03} & 66.23\tiny{{$\pm$}0.04} & 65.75\tiny{{$\pm$}0.02} & 67.23\tiny{{$\pm$}0.58} & 67.34\tiny{{$\pm$}0.03} \\
 \midrule
\multirow{4}{*}{\textbf{4}} & \multirow{4}{*}{\textbf{PGD}} & E & 59.70\tiny{{$\pm$}0.06} & 57.71\tiny{{$\pm$}0.05} & 57.73\tiny{{$\pm$}0.09} & 57.19\tiny{{$\pm$}0.07} & 57.60\tiny{{$\pm$}0.08} & 57.05\tiny{{$\pm$}0.17} & 54.69\tiny{{$\pm$}0.09} & 58.18\tiny{{$\pm$}0.07} & 58.27\tiny{{$\pm$}0.09} & 58.46\tiny{{$\pm$}0.11} & 57.66\tiny{{$\pm$}0.05} & 58.81\tiny{{$\pm$}0.64} & 59.14\tiny{{$\pm$}0.05} \\
 &  & M & 68.40\tiny{{$\pm$}0.05} & 66.12\tiny{{$\pm$}0.02} & 66.39\tiny{{$\pm$}0.04} & 65.67\tiny{{$\pm$}0.04} & 67.04\tiny{{$\pm$}0.03} & 68.24\tiny{{$\pm$}0.04} & 65.64\tiny{{$\pm$}0.08} & 65.17\tiny{{$\pm$}0.05} & 67.32\tiny{{$\pm$}0.03} & 67.85\tiny{{$\pm$}0.05} & 66.78\tiny{{$\pm$}0.02} & 68.16\tiny{{$\pm$}0.23} & 68.12\tiny{{$\pm$}0.03} \\
 &  & H & 75.83\tiny{{$\pm$}0.03} & 72.91\tiny{{$\pm$}0.02} & 72.47\tiny{{$\pm$}0.04} & 72.18\tiny{{$\pm$}0.05} & 73.52\tiny{{$\pm$}0.02} & 75.55\tiny{{$\pm$}0.05} & 73.58\tiny{{$\pm$}0.04} & 69.64\tiny{{$\pm$}0.05} & 73.89\tiny{{$\pm$}0.02} & 74.34\tiny{{$\pm$}0.04} & 73.39\tiny{{$\pm$}0.01} & 75.24\tiny{{$\pm$}0.65} & 75.36\tiny{{$\pm$}0.02} \\
 &  & F & 68.01\tiny{{$\pm$}0.02} & 65.41\tiny{{$\pm$}0.01} & 65.54\tiny{{$\pm$}0.03} & 65.05\tiny{{$\pm$}0.03} & 66.22\tiny{{$\pm$}0.02} & 66.49\tiny{{$\pm$}0.04} & 64.63\tiny{{$\pm$}0.04} & 64.82\tiny{{$\pm$}0.04} & 66.32\tiny{{$\pm$}0.02} & 66.14\tiny{{$\pm$}0.04} & 65.86\tiny{{$\pm$}0.01} & 66.94\tiny{{$\pm$}0.76} & 67.37\tiny{{$\pm$}0.02} \\
 \midrule
\multirow{4}{*}{\textbf{5}} & \multirow{4}{*}{\textbf{FGSM}} & E & 59.71\tiny{{$\pm$}0.05} & 57.69\tiny{{$\pm$}0.08} & 57.62\tiny{{$\pm$}0.06} & 57.16\tiny{{$\pm$}0.08} & 57.60\tiny{{$\pm$}0.06} & 56.97\tiny{{$\pm$}0.09} & 54.67\tiny{{$\pm$}0.08} & 58.20\tiny{{$\pm$}0.10} & 58.23\tiny{{$\pm$}0.06} & 58.46\tiny{{$\pm$}0.07} & 57.63\tiny{{$\pm$}0.05} & 58.81\tiny{{$\pm$}0.65} & 59.15\tiny{{$\pm$}0.04} \\
 &  & M & 68.37\tiny{{$\pm$}0.02} & 66.10\tiny{{$\pm$}0.03} & 66.38\tiny{{$\pm$}0.04} & 65.70\tiny{{$\pm$}0.05} & 67.03\tiny{{$\pm$}0.04} & 68.27\tiny{{$\pm$}0.04} & 65.61\tiny{{$\pm$}0.08} & 65.16\tiny{{$\pm$}0.05} & 67.30\tiny{{$\pm$}0.02} & 67.84\tiny{{$\pm$}0.07} & 66.78\tiny{{$\pm$}0.02} & 68.16\tiny{{$\pm$}0.23} & 68.11\tiny{{$\pm$}0.02} \\
 &  & H & 75.82\tiny{{$\pm$}0.02} & 72.92\tiny{{$\pm$}0.04} & 72.48\tiny{{$\pm$}0.03} & 72.18\tiny{{$\pm$}0.05} & 73.52\tiny{{$\pm$}0.02} & 75.55\tiny{{$\pm$}0.05} & 73.60\tiny{{$\pm$}0.04} & 69.64\tiny{{$\pm$}0.04} & 73.90\tiny{{$\pm$}0.01} & 74.34\tiny{{$\pm$}0.04} & 73.39\tiny{{$\pm$}0.01} & 75.23\tiny{{$\pm$}0.65} & 75.35\tiny{{$\pm$}0.02} \\
 &  & F & 68.00\tiny{{$\pm$}0.02} & 65.41\tiny{{$\pm$}0.02} & 65.54\tiny{{$\pm$}0.04} & 65.05\tiny{{$\pm$}0.04} & 66.22\tiny{{$\pm$}0.02} & 66.50\tiny{{$\pm$}0.06} & 64.65\tiny{{$\pm$}0.04} & 64.82\tiny{{$\pm$}0.03} & 66.34\tiny{{$\pm$}0.03} & 66.15\tiny{{$\pm$}0.06} & 65.87\tiny{{$\pm$}0.01} & 66.95\tiny{{$\pm$}0.75} & 67.37\tiny{{$\pm$}0.01} \\
 \midrule
\multirow{4}{*}{\textbf{6}} & \multirow{4}{*}{\textbf{W/O Attack}} & E & 59.67\tiny{{$\pm$}0.00} & 58.08\tiny{{$\pm$}0.00} & 60.22\tiny{{$\pm$}0.00} & 58.53\tiny{{$\pm$}0.00} & 58.14\tiny{{$\pm$}0.00} & 60.78\tiny{{$\pm$}0.00} & 56.83\tiny{{$\pm$}0.00} & 59.47\tiny{{$\pm$}0.00} & 59.62\tiny{{$\pm$}0.00} & 59.88\tiny{{$\pm$}0.00} & 59.12\tiny{{$\pm$}0.00} & 60.29\tiny{{$\pm$}0.37} & 60.42\tiny{{$\pm$}0.00} \\
 &  & M & 68.28\tiny{{$\pm$}0.00} & 66.14\tiny{{$\pm$}0.00} & 67.11\tiny{{$\pm$}0.00} & 66.35\tiny{{$\pm$}0.00} & 67.00\tiny{{$\pm$}0.00} & 68.98\tiny{{$\pm$}0.00} & 66.26\tiny{{$\pm$}0.00} & 65.41\tiny{{$\pm$}0.00} & 67.53\tiny{{$\pm$}0.00} & 68.41\tiny{{$\pm$}0.00} & 67.15\tiny{{$\pm$}0.00} & 68.56\tiny{{$\pm$}0.30} & 68.59\tiny{{$\pm$}0.00} \\
 &  & H & 75.85\tiny{{$\pm$}0.00} & 73.05\tiny{{$\pm$}0.00} & 72.69\tiny{{$\pm$}0.00} & 72.66\tiny{{$\pm$}0.00} & 73.46\tiny{{$\pm$}0.00} & 75.64\tiny{{$\pm$}0.00} & 73.69\tiny{{$\pm$}0.00} & 69.84\tiny{{$\pm$}0.00} & 74.10\tiny{{$\pm$}0.00} & 75.76\tiny{{$\pm$}0.00} & 73.67\tiny{{$\pm$}0.00} & 75.75\tiny{{$\pm$}0.09} & 75.52\tiny{{$\pm$}0.00} \\
 &  & F & 67.93\tiny{{$\pm$}0.00} & 65.76\tiny{{$\pm$}0.00} & 66.68\tiny{{$\pm$}0.00} & 65.85\tiny{{$\pm$}0.00} & 66.20\tiny{{$\pm$}0.00} & 68.47\tiny{{$\pm$}0.00} & 65.59\tiny{{$\pm$}0.00} & 64.91\tiny{{$\pm$}0.00} & 67.08\tiny{{$\pm$}0.00} & 68.02\tiny{{$\pm$}0.00} & 66.65\tiny{{$\pm$}0.00} & 68.14\tiny{{$\pm$}0.24} & 68.11\tiny{{$\pm$}0.00} \\
\midrule
\multicolumn{2}{c}{\multirow{4}{*}{\begin{tabular}[c]{@{}c@{}}\textbf{Avg.}\\\textbf{Accuracy}\end{tabular}}} & E & {\color{blue}59.62\tiny{{$\pm$}0.02}} & 57.44\tiny{{$\pm$}0.03} & 57.77\tiny{{$\pm$}0.03} & 56.84\tiny{{$\pm$}0.04} & 56.79\tiny{{$\pm$}0.04} & 55.62\tiny{{$\pm$}0.06} & 54.33\tiny{{$\pm$}0.04} & 57.53\tiny{{$\pm$}0.05} & 56.74\tiny{{$\pm$}0.09} & 55.67\tiny{{$\pm$}0.10} & - & - & - \\
\multicolumn{2}{c}{} & M & {\color{blue}68.34\tiny{{$\pm$}0.01}} & 65.88\tiny{{$\pm$}0.01} & 66.41\tiny{{$\pm$}0.01} & 65.59\tiny{{$\pm$}0.02} & 66.94\tiny{{$\pm$}0.02} & 67.24\tiny{{$\pm$}0.19} & 65.43\tiny{{$\pm$}0.03} & 64.97\tiny{{$\pm$}0.02} & 67.28\tiny{{$\pm$}0.01} & 66.85\tiny{{$\pm$}0.18} & - & - & - \\
\multicolumn{2}{c}{} & H & {\color{blue}75.84\tiny{{$\pm$}0.01}} & 72.69\tiny{{$\pm$}0.01} & 72.42\tiny{{$\pm$}0.01} & 72.14\tiny{{$\pm$}0.02} & 73.47\tiny{{$\pm$}0.01} & 75.49\tiny{{$\pm$}0.01} & 73.33\tiny{{$\pm$}0.02} & 69.38\tiny{{$\pm$}0.02} & 73.98\tiny{{$\pm$}0.00} & 74.78\tiny{{$\pm$}0.02} & - & - & - \\
\multicolumn{2}{c}{} & F & {\color{blue}67.90\tiny{{$\pm$}0.01}} & 64.87\tiny{{$\pm$}0.05} & 65.02\tiny{{$\pm$}0.03} & 64.41\tiny{{$\pm$}0.04} & 65.12\tiny{{$\pm$}0.25} & 65.45\tiny{{$\pm$}0.26} & 63.65\tiny{{$\pm$}0.07} & 63.42\tiny{{$\pm$}0.29} & 64.53\tiny{{$\pm$}0.84} & 64.46\tiny{{$\pm$}1.13} & - & - & - \\
\midrule
\multicolumn{2}{c}{\multirow{4}{*}{\begin{tabular}[c]{@{}c@{}}\textbf{Avg. 3-Min} \\\textbf{Accuracy}\end{tabular}}} & E & {\color{blue}59.55\tiny{{$\pm$}0.03}} & 57.05\tiny{{$\pm$}0.04} & 57.02\tiny{{$\pm$}0.03} & 56.05\tiny{{$\pm$}0.07} & 55.73\tiny{{$\pm$}0.07} & 52.33\tiny{{$\pm$}0.12} & 53.25\tiny{{$\pm$}0.07} & 56.43\tiny{{$\pm$}0.07} & 54.77\tiny{{$\pm$}0.16} & 52.41\tiny{{$\pm$}0.17} & - & - & - \\
\multicolumn{2}{c}{} & M & {\color{blue}68.28\tiny{{$\pm$}0.01}} & 65.64\tiny{{$\pm$}0.02} & 66.20\tiny{{$\pm$}0.01} & 65.28\tiny{{$\pm$}0.03} & 66.84\tiny{{$\pm$}0.02} & 65.85\tiny{{$\pm$}0.40} & 65.02\tiny{{$\pm$}0.04} & 64.69\tiny{{$\pm$}0.03} & 67.17\tiny{{$\pm$}0.02} & 65.66\tiny{{$\pm$}0.34} & - & - & - \\
\multicolumn{2}{c}{} & H & {\color{blue}75.80\tiny{{$\pm$}0.02}} & 72.42\tiny{{$\pm$}0.02} & 72.29\tiny{{$\pm$}0.01} & 71.93\tiny{{$\pm$}0.02} & 73.42\tiny{{$\pm$}0.01} & 75.36\tiny{{$\pm$}0.02} & 73.05\tiny{{$\pm$}0.02} & 69.04\tiny{{$\pm$}0.03} & 73.92\tiny{{$\pm$}0.01} & 74.17\tiny{{$\pm$}0.03} & - & - & - \\
\multicolumn{2}{c}{} & F & {\color{blue}67.78\tiny{{$\pm$}0.02}} & 64.22\tiny{{$\pm$}0.11} & 64.12\tiny{{$\pm$}0.06} & 63.50\tiny{{$\pm$}0.08} & 64.02\tiny{{$\pm$}0.49} & 63.39\tiny{{$\pm$}0.53} & 62.35\tiny{{$\pm$}0.14} & 61.99\tiny{{$\pm$}0.58} & 62.44\tiny{{$\pm$}1.69} & 62.11\tiny{{$\pm$}2.26} & - & - & - \\
\midrule
\multicolumn{2}{c}{\multirow{4}{*}{\begin{tabular}[c]{@{}c@{}}\textbf{Weighted} \\\textbf{Accuracy}\end{tabular}}} & E & {\color{blue}59.53\tiny{{$\pm$}0.04}} & 56.93\tiny{{$\pm$}0.04} & 56.94\tiny{{$\pm$}0.04} & 55.93\tiny{{$\pm$}0.08} & 54.63\tiny{{$\pm$}0.14} & 48.21\tiny{{$\pm$}0.27} & 52.23\tiny{{$\pm$}0.08} & 55.55\tiny{{$\pm$}0.14} & 52.18\tiny{{$\pm$}0.33} & 47.45\tiny{{$\pm$}0.35} & - & - & - \\
\multicolumn{2}{c}{} & M & {\color{blue}68.25\tiny{{$\pm$}0.02}} & 65.57\tiny{{$\pm$}0.02} & 66.17\tiny{{$\pm$}0.02} & 65.28\tiny{{$\pm$}0.02} & 66.79\tiny{{$\pm$}0.02} & 63.85\tiny{{$\pm$}0.80} & 64.77\tiny{{$\pm$}0.07} & 64.60\tiny{{$\pm$}0.03} & 67.06\tiny{{$\pm$}0.03} & 64.07\tiny{{$\pm$}0.68} & - & - & - \\
\multicolumn{2}{c}{} & H & {\color{blue}75.78\tiny{{$\pm$}0.02}} & 72.37\tiny{{$\pm$}0.02} & 72.20\tiny{{$\pm$}0.01} & 71.92\tiny{{$\pm$}0.03} & 73.41\tiny{{$\pm$}0.01} & 75.30\tiny{{$\pm$}0.02} & 72.98\tiny{{$\pm$}0.02} & 68.99\tiny{{$\pm$}0.04} & 73.91\tiny{{$\pm$}0.01} & 74.08\tiny{{$\pm$}0.03} & - & - & - \\
\multicolumn{2}{c}{} & F & {\color{blue}67.73\tiny{{$\pm$}0.03}} & 63.96\tiny{{$\pm$}0.21} & 63.19\tiny{{$\pm$}0.10} & 62.80\tiny{{$\pm$}0.15} & 62.18\tiny{{$\pm$}0.98} & 61.58\tiny{{$\pm$}1.05} & 61.00\tiny{{$\pm$}0.28} & 60.54\tiny{{$\pm$}1.18} & 59.82\tiny{{$\pm$}3.38} & 59.37\tiny{{$\pm$}4.53} & - & - & - \\
\bottomrule
\end{tabular}}
\label{tab:leaderboard_aminer}
\end{table}

\begin{figure}[htbp]
\centering
\begin{minipage}[t]{\textwidth}
\centering
\includegraphics[width=\textwidth]{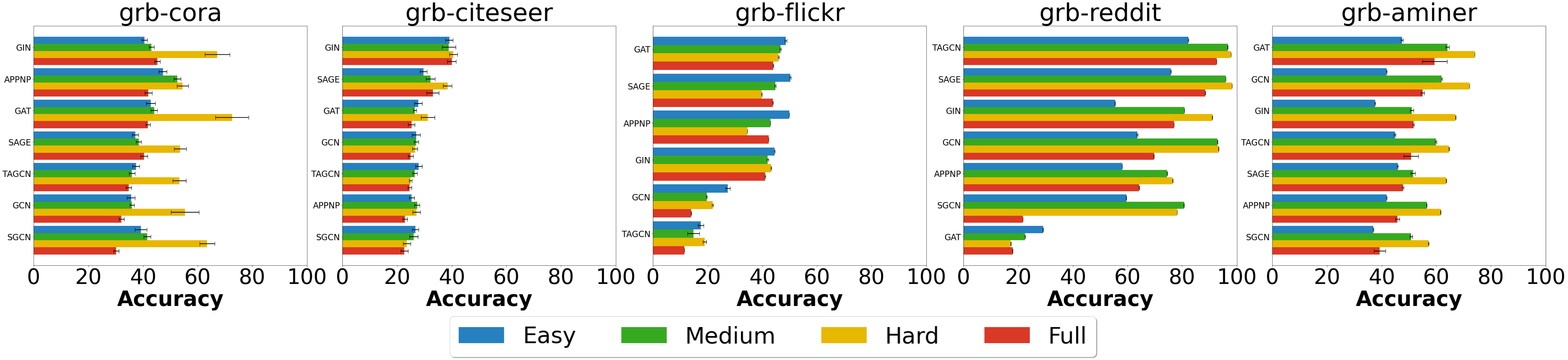}
\caption{Ranking of vanilla GML models in \textit{graph injection} scenario for all five datasets.}
\label{fig:rank_inj_vanilla}
\end{minipage}\vspace{5mm} \\
\begin{minipage}[t]{\textwidth}
\centering
\includegraphics[width=\textwidth]{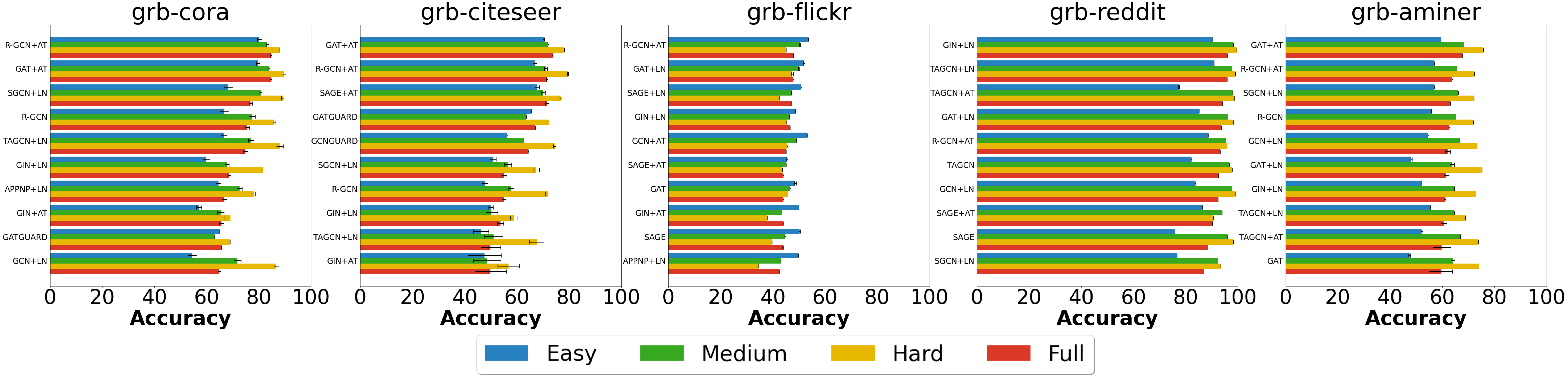}
\caption{Ranking of top 10 defensed GML models in \textit{graph injection} scenario for all five datasets.}
\label{fig:rank_inj}
\end{minipage}
\end{figure}

\textbf{Robust Ranking of GML Models.} 
In~\figurename~\ref{fig:rank_inj_vanilla} and~\ref{fig:rank_inj}, we show the rankings of GML models for all five datasets in \textit{graph injection} scenario. 
The ranking is determined by $s_{w}^{\tiny{\textrm{DEF}}}$ calculated by multiple attacks, which makes the results more general than previous works (only consider very few attacks and vanilla GML models). 
We find that the rankings are different across datasets, indicating that the robustness is related to the properties of graph data. 
Similar situations can be found in other graph benchmarks. For example in OGB, there is no dominant GML model, the performance of certain model architecture may vary a lot across datasets. 
Thus, we suggest that \textit{when giving conclusions about robustness in GML, one should not only consider the model itself but also take the graph data into account}. 
\name provides scalable datasets of various domains, which can help to investigate the robustness of GML models in different situations.
Among current vanilla GML models, we find that GAT and GIN generally perform better under attacks in several datasets, which might be due to the higher expressiveness of model architecture. 
Meanwhile, models like APPNP and SGCN that rely on high-order message propagation seem to be sensible to perturbations on the graph. 
Besides, GML models with defense mechanisms (\emph{i.e.}, R-GCN, GNNGuard) are generally more robust. 
Moreover, we find simple methods like LN can be applied to all GML models to increase robustness. 
In the following, we further analyze some factors that affect the adversarial robustness of GML models. 

\begin{figure}[htbp]
\centering
\includegraphics[width=\textwidth]{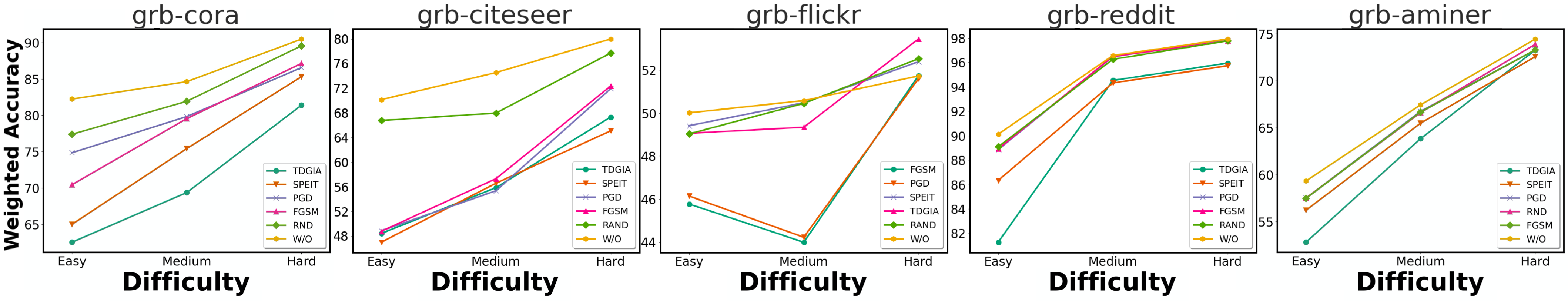}
\caption{Effect of dataset difficulties on the performance of \textit{graph injection} attacks.}
\label{fig:difficulty}
\end{figure}

\textbf{Effect of Difficulties.} 
The new splitting scheme investigates the effect of the average degree of target nodes on the attack performance. 
In ~\figurename~\ref{fig:difficulty}, attacks tend to better decrease the performance on nodes with lower degrees, which confirms the assumption that these low-degree nodes are more vulnerable. 
Moreover, according to~\figurename~\ref{fig:rank_inj_vanilla} and~\ref{fig:rank_inj}, the robustness on these nodes is indeed harder to achieve. 
This phenomenon encourages future work to deal with these vulnerable nodes to design more robust GML models. 

\begin{figure}[htbp]
\centering
\vspace{-0.1in}
\begin{minipage}[t]{0.49\textwidth}
\centering
\includegraphics[width=\textwidth]{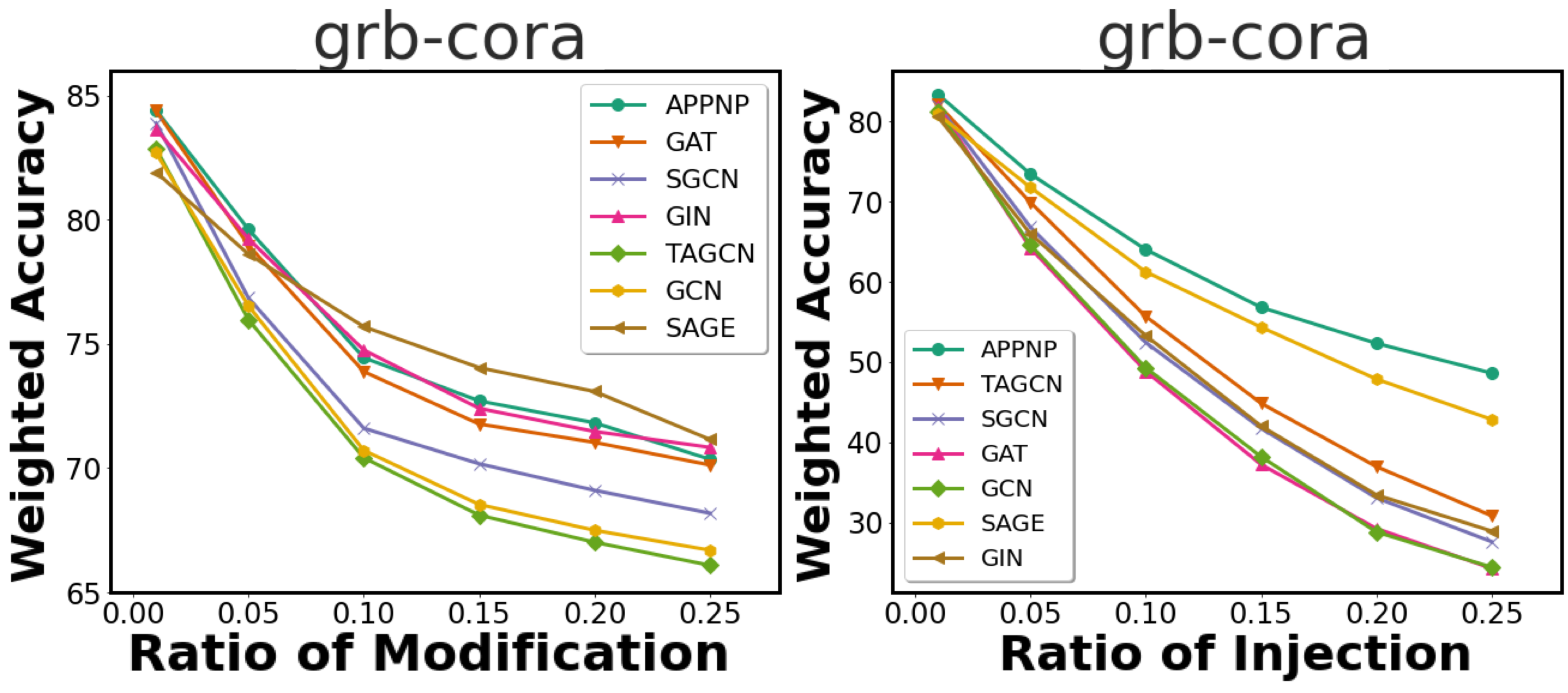}
\caption{Effect of constraints on GML models. Left: \textit{graph modification}. Right: \textit{graph injection}.}
\label{fig:constraint_gml}
\end{minipage}%
\hspace{\fill}
\begin{minipage}[t]{0.49\textwidth}
\centering
\includegraphics[width=\textwidth]{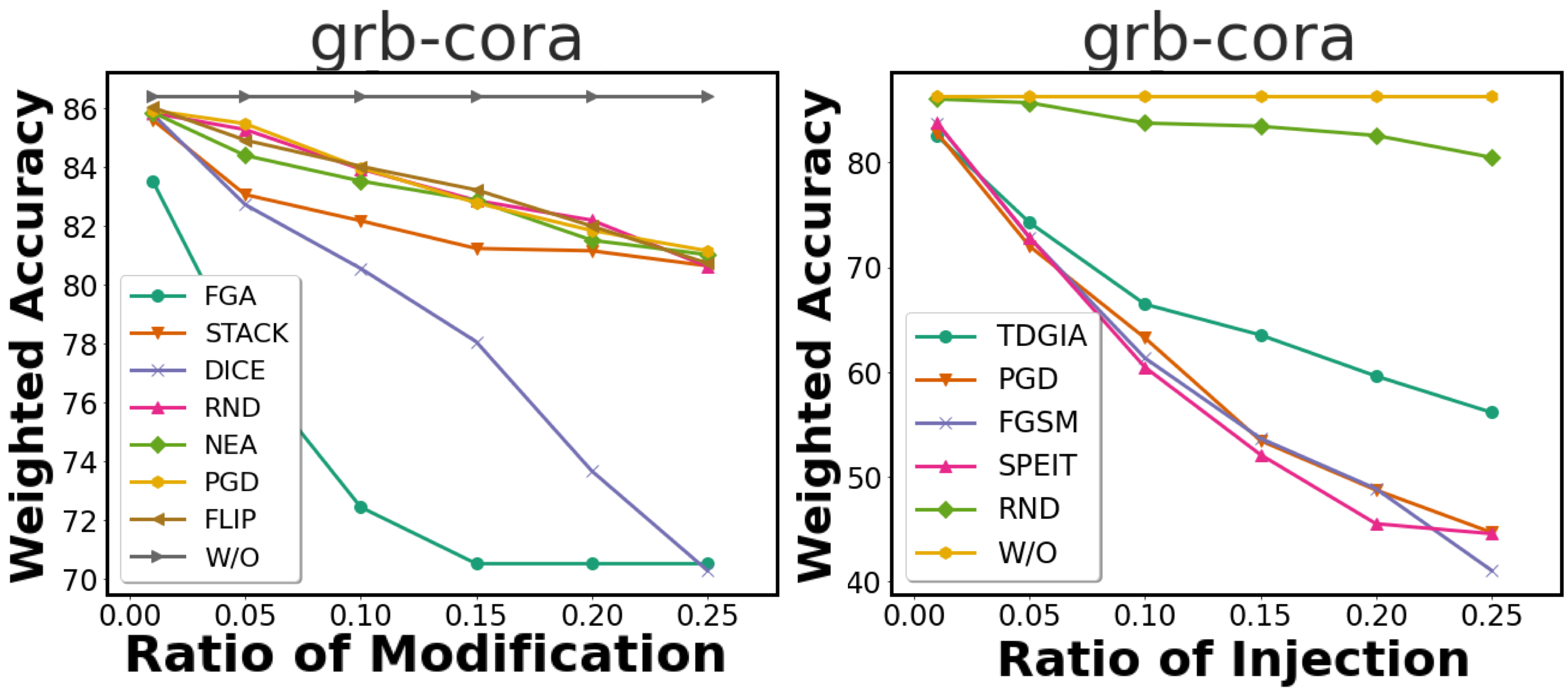}
\caption{Effect of constraints on attacks. Left: \textit{graph modification}. Right: \textit{graph injection}.}
\label{fig:constraint_attack}
\end{minipage}
\vspace{-0.1in}
\end{figure}

\textbf{Effect of Constraints.} 
As shown in~\figurename~\ref{fig:constraint_gml} and~\ref{fig:constraint_attack}, for both \textit{graph modification} and \textit{graph injection} scenarios, the variation of constraints on the ratio of modification/injection affects the effectiveness of attacks. 
Meanwhile, the ranking of methods nearly agrees with different constraints. 
Without loss of generality, it is reasonable to fix a specific constraint to build \name leaderboards, where the relative robustness of GML models will still be indicative.  

\begin{figure}[htbp]
\centering
\begin{minipage}[t]{\textwidth}
\includegraphics[width=\textwidth]{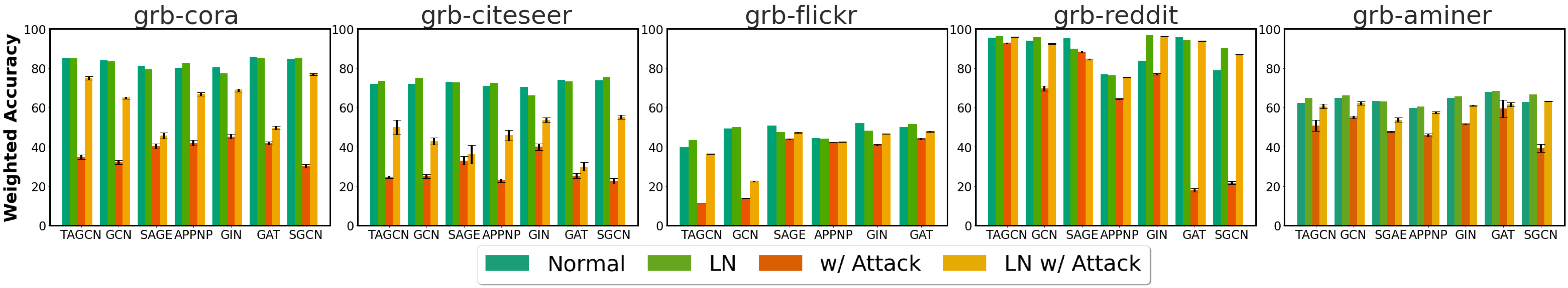}
\caption{Effect of the adapted LN on the adversarial robustness of vanilla GML models for all five datasets. Adding LN can generally increase robustness of GML models.}
\label{fig:layer_norm}
\end{minipage}\vspace{5mm} \\
\begin{minipage}[t]{\textwidth}
\centering
\includegraphics[width=\textwidth]{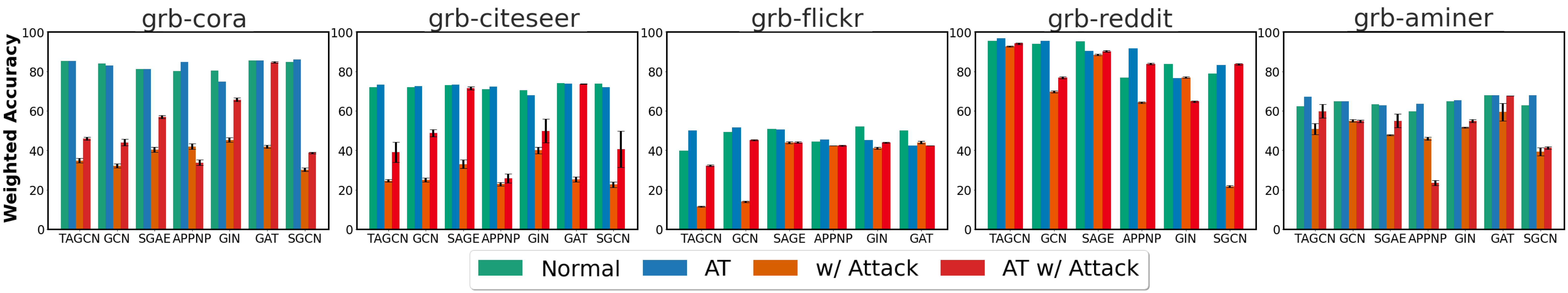}
\caption{Effect of the adapted AT on the adversarial robustness of vanilla GML models for all five datasets. Adding AT can generally increase robustness of GML models.}
\label{fig:adv_train}
\end{minipage}
\end{figure}

\textbf{Effect of General Defenses.} 
\figurename~\ref{fig:layer_norm} and \ref{fig:adv_train} shows the results of the adapted LN and AT for all five datasets. 
LN is a node-wise normalization technique, which can alleviate the perturbations on node features as well as hidden features in each layer of GML models. 
AT applies adversarial attacks during training via modification or injection, which changes the decision boundary of models to tolerate perturbed nodes. 
The results indicate that these approaches can generally increase the robustness of various types of GML models, which can serve as simple but strong baselines for future works.
The details of these algorithms can be found in~\appendixname~\ref{app:defense}.

\section{Conclusion}\label{sec:conclusion}
To improve and facilitate the evaluation of the adversarial robustness in GML, we revisit the limitations of previous works and present the Graph Robustness Benchmark (\name), a \textit{scalable}, \textit{unified}, \textit{modular}, and \textit{reproducible} benchmark.
It has scalable datasets, unified evaluation scenarios, as well as a modular coding framework that ensures the reproducibility and promotes the development of future methods. 
Extensive experiments with \name provide insights on the understanding of the adversarial robustness in GML. 
We welcome the community to contribute more advanced GML models, attacks and defenses to further enrich \name and to promote the research of this field.

\section{Broader Impact}\label{sec:impact}
\textbf{Positive Impact.} 
\name provides a general framework for GML attacks and defenses. On one hand, it will help researchers to develop more robust GML models against attacks. On the other hand, it will also help possible attackers to develop better attack methods to turn down defenses. More public information of potential attacks will make it harder to conduct secret attacks based on private methods. As a result, more generally robust defense mechanisms can be designed. 

\textbf{Negative Impact.} 
By exposing the attack methods widely, the GML models may face more threats. Attackers can use the benchmark to design destructive attacks that may cause damage to GML-based systems. 
Additionally, \name has some limitations. For example, it  only considers homogeneous graphs rather than heterogeneous ones for now. 
It focuses on node classification, while other tasks like link prediction and graph classification are also vulnerable. 
We will regularly update \name (\emph{e.g.}, adding task-specific modules, designing related metrics.) to overcome these limitations. 

\section{Maintenance Plan}\label{sec:maintenance}
\textbf{Open Source.} 
We host the \name homepage (\textcolor{blue}{\url{https://cogdl.ai/grb/home}}) with detailed introduction, leaderboards, and documentations. 
The codes are available in (\textcolor{blue}{\url{https://github.com/THUDM/grb}}). All materials are accessible to ensure reproducibility.

\textbf{Submissions of New Methods.} 
\name will regularly include SOTA methods by updating the "method zoo". To welcome the contribution of the community, we allow submissions through \href{https://docs.google.com/forms/d/e/1FAIpQLSfJaUK-SXYFnlSqTEEwTOwsqzA5JnpXyvZe8E24hlLE7scRcA/viewform}{google form}. There are detailed \href{https://github.com/THUDM/grb/tree/master/examples}{examples} and \href{https://cogdl.ai/grb/intro/rules}{rules} that guide researchers to add new attacks or defenses. Results will be updated on leaderboards to track the progress of the domain. 

\textbf{Extension of Tasks.} 
Due to the modular design, \name can be extended to other tasks. It requires adding task-specific functions in each module (dataset, model, trainer, attack, defense, etc.). 
Other common functions in GML can be reused for different tasks. There are online examples ({\textcolor{blue}{\url{https://github.com/THUDM/grb/tree/master/examples}}}) showing how to use \name for other tasks, \emph{e.g.}, graph classification. In the future, \name will support more GML tasks and define related threat models and metrics to unify the evaluation of adversarial robustness.

\bibliography{reference}
\bibliographystyle{unsrt}

\newpage
\appendix

\section{Appendix}\label{sec:appendix}
\subsection{Datasets}\label{app:dataset}

\begin{table}[!ht]
\centering
\caption{Statistics of \name datasets after new splitting scheme and feature normalization.}
\scalebox{0.85}{
\begin{tabular}{lcrrrc} 
\toprule
\multicolumn{1}{c}{\textbf{Dataset}} & \textbf{Splitting}                                                                                                                       & \multicolumn{1}{c}{\begin{tabular}[c]{@{}c@{}}\textbf{Avg. }\\\textbf{Deg.}\end{tabular}} & \multicolumn{1}{c}{\begin{tabular}[c]{@{}c@{}}\textbf{Avg.Deg.}\\\textbf{(E / M / H / F)}\end{tabular}} & \multicolumn{1}{c}{\begin{tabular}[c]{@{}c@{}}\textbf{Feature range}\\\textbf{(original)}\end{tabular}} & \multicolumn{1}{c}{\begin{tabular}[c]{@{}c@{}}\textbf{Feature range}\\\textbf{(normalized)}\end{tabular}}  \\
\midrule
\textit{grb-cora}                             & \multirow{5}{*}{\begin{tabular}[c]{@{}c@{}}\textbf{Train / Val}\\0.6 / 0.1 \\\textbf{Test: E / M / H / F}\\0.1 / 0.1 / 0.1 / 0.3\end{tabular}} & 3.84                                                                                      & 1.53/2.96/5.23/3.24                                                                                      & {[}-2.30, 2.40]                            & {[}-0.94, 0.94]                                                                                              \\
\textit{grb-citeseer}                              &                                                                                                                                          & 2.61                                                                                     & 1.01/1.74/3.82/2.19                                                                                & {[}-4.55, 1.67]                           & {[}-0.96, 0.89]                                                                                              \\
\textit{grb-flickr}                           &                                                                                                                                          & 10.08                                                                                     & 5.00/6.02/11.03/7.35                                                                                 & {[}-0.90, 269.96]                          & {[}-0.47, 1.00]                                                                                              \\
\textit{grb-reddit}                           &                                                                                                                                          & 99.65                                                                                     & 29.23/68.36/150.99/82.86                                                                                  & {[}-28.19, 120.96]                         & {[}-0.98, 0.99]                                                                                              \\
\textit{grb-aminer}                           &                                                                                                                                          & 8.73                                                                                      & 1.99/5.12/13.25/6.79                                                                                     & {[}-1.74, 1.62]                            & {[}-0.93, 0.93]                                                                                           \\
\bottomrule
\end{tabular}}
\label{tab:dataset_processed}
\end{table}

\name includes five datasets of different scales, the details of them are as following:
\setlist[itemize]{leftmargin=5mm}
\begin{itemize}
    \item \textit{grb-cora}: Small-scale citation networks. Each node represents a research paper, and each edge represents a citation relationship between two papers. Instead of using the popular version of Cora used in Planetoid~\cite{yang2016revisiting}, we use a refined version~\cite{zou2019dimensional}, which removes duplicated nodes and generates indirect pre-trained word embeddings as node features to solve the problem of information leakage in the original version. As a result, the features become 302-dimension continuous features rather than 1433-dimension binary features in the original version. The task is to classify papers into 7 categories.
    \item \textit{grb-citeseer}: Small-scale citation networks. Similar to \textit{grb-cora}, we use a refined version~\cite{zou2019dimensional} of CiteSeer, which eliminates identical papers and generates text embeddings by pre-trained BERT~\cite{devlin2018bert} model. The resulting features are 768-dimension continuous features rather than 3703-dimension binary features in the original version. The task is to classify papers into 6 categories.
    \item \textit{grb-flickr}: Medium-scale social networks. We adopt the Flickr dataset from~\cite{zeng2019graphsaint}, which contains descriptions and common properties of online images. The dataset is processed with a new splitting scheme and feature normalization mentioned in~\ref{sec:exp}. The task is to classify images into 7 categories.
    \item \textit{grn-reddit}: Large-scale social networks. We adopt the Reddit dataset from~\cite{zeng2019graphsaint}, which contains the communities of online posts based on user comments. The task is to classify communities into 41 categories. 
    \item \textit{grb-aminer}: Large-scale citation networks. The papers are collected from the academic searching engine Aminer~\cite{tang2008arnetminer}, and the dataset was used in KDD-CUP 2020 Graph Adversarial Attack \& Defense competition. The task is to classify papers into 18 categories. 
\end{itemize}

All five datasets are processed by the new splitting scheme and feature normalization mentioned in~\ref{sec:exp}. The datasets are saved in the format of numpy~\cite{van2011numpy} zipped format (with \texttt{.npz} extension), and each has four files: \texttt{adj.npz}, \texttt{features.npz}, \texttt{index.npz} and \texttt{labels.npz}. The data can be loaded by using the \textit{Dataset} module in \name. All data are maintained and can be found in {\textcolor{blue}{\url{https://cogdl.ai/grb/datasets}}}, where we will continuously update to ensure the accessibility for a long term. We use MIT license for data and codes.

\subsection{Related Works}\label{app:related}

In other domains like image classification, there are already standards~\cite{carlini2019evaluating} or benchmarks~\cite{dong2020benchmarking}~\cite{croce2020robustbench} for evaluating adversarial robustness. Besides, there exists a toolkit like DeepRobust~\cite{li2020deeprobust} that implements adversarial attacks and defenses for both image classification and GML tasks. There are currently several benchmarks in GML. Open Graph Benchmark (OGB)~\cite{hu2020open} develops a diverse set of scalable and realistic datasets, which facilitates the evaluation of GML models. Dwivedi et al.~\cite{dwivedi2020benchmarking} proposes a reproducible GNN benchmarking framework to facilitate researchers to add new models conveniently for arbitrary datasets. These benchmarks mainly focus on the performance but not the robustness of GNNs. so far, there is no benchmark on evaluating the \textit{adversarial robustness} of GML models, i.e. the robustness in the presence of adversarial attacks. Nevertheless, it is an important but challenging task, which requires avoiding pitfalls in previous works and proposing a better solution. 

\subsection{Rethinking Ill-defined Evaluation Scenarios in Previous Works}
\label{sec:unnoticeablilty}

Many of the previous adversarial attacks~\cite{zugner2018adversarial, dai2018adversarial, zugner2019adversarial} consider the \textit{poisoning} attack and develop the notion of \textit{unnoticeability}, similar to Eq.~\ref{eq:attack}. The initial idea is to imitate the same notion in image classification task: the differences of adversarial examples, compared with clean examples, should be tiny and unnoticeable, so that humans can still easily recognize the objects in images. That's why $l_p$-norm is a widely-used constraint, as it corresponds to the visual sense of humans. 

In the \textit{poisoning} setting of graph modification attacks, the attackers assume that the graph is perturbed with corrupted nodes and edges, in a way that the perturbed graph is close to the original one. However, this assumption is controversial: If defenders have the original graph, they can simply train the model on that one; If defenders do not have the original graph (the general case for data poisoning where defenders can not tell whether the data are benign or not), then it does not make sense to keep \textit{unnoticeability}. In this case, we only have $\mathcal{G'}=(\mathcal{A'}, \mathcal{F'})$ but not $\mathcal{G}=(\mathcal{A}, \mathcal{F})$ in Eq.~\ref{eq:attack}, making it almost impossible to compare them. Previous works propose to compare the graph properties, like degree distribution~\cite{zugner2018adversarial}, feature statistics~\cite{wu2019adversarial} or topological properties~\cite{du2017topology}. However, all these comparisons need to be done in presence of the original graph. This is different from the case of images, where \textit{unnoticeability} can be easily judged by humans even without ground-truth images. %That is to say, for graph adversarial attacks, \textit{unnoticeability} shall be considered from the defenders' view instead of the attackers'.

The attackers may perturb the graph structure or attributes within the scope of \textit{unnoticeability} defined by themselves, while defenders have to depend on their own observations to discover. For example, Nettack~\cite{zugner2018adversarial} proposes to keep the degree distribution of modified graph similar to the original one. However, even if defenders notice that the degree distribution is different, it is still hard to identify specific malicious nodes or edges from the entire graph. On the contrary, defenses like GNNGuard~\cite{zhang2020gnnguard} can use the dissimilarity between features to alleviate effects of perturbations. We argue that it is inadequate to simply adopt the notion from image classification, and to make two graphs ``similar'' in whatever way. Indeed, there is not an absolute definition, but it is recommended that: \textit{``Unnoticeability'' shall be considered from the defenders' view instead of the attackers'.}

As a starting point, we consider very basic constraints in \name (\emph{e.g.}, a limited number of modified edges or nodes). Pre-defined complex constraints might ignore the real capability of attackers and defenders, and might be obsoleted as the research goes. Thus, we do not add too many constraints and we insist that the notion like ``unnoticeability'' will be refined during the arms race between attackers and defenders. For example, if an advanced defense proposes a measure to identify malicious nodes with high probability, then the attackers can decide by themselves to refine the constraints based on this measure. There will be a trade-off, considering more constraints for one specific defense might result in less effectiveness for other methods. That's also why \name considers a general metric across multiple attacks/defenses rather than a single pair of attack/defense. We insist that finding methods that are generally more effective bring much more value in practical applications.

\subsection{Methodology}\label{app:method}

\subsubsection{GML Models}\label{app:gnn}
GCN (Graph Convolutional Networks)~\cite{kipf2016semi} introduces a layer-wise propagation rule for graph-structured data which is motivated from a first-order approximation of spectral graph convolutions. 
GAT (Graph Attention Networks)~\cite{velivckovic2018graph} leverages masked self-attention layers where nodes can attend over their neighborhoods' features with different weights. 
GIN (Graph Isomorphic Networks)~\cite{xu2018powerful} is a theoretically guaranteed framework for analyzing the expressive power of GNNs to capture different graph structures. 
APPNP (Approximated Personalized Propagation of Neural Predictions )~\cite{klicpera2018predict} utilizes an improved propagation scheme based on personalized PageRank to construct a simple model with fast approximation. 
TAGCN (Topological Adaptive Graph Convolutional Networks)~\cite{du2017topology} provides a systematic way to design a
set of fixed-size learnable filters to perform convolutions on graphs. 
GraphSAGE~\cite{hamilton2017inductive} is a general inductive framework that leverages node features to generate node embeddings for previously unseen data. 
SGCN (Simplified Graph Convolutional Networks)~\cite{wu2019simplifying} removes nonlinearities and collapses weight matrices between consecutive layers, resulting in a linear model. 

\subsubsection{Adversarial Attacks}\label{app:attack}

\textbf{Modification Attacks.} RND (Random)~\cite{zugner2018adversarial} is a random attack strategy that only modifies the structure of the graph. DICE (Delete Internally Connect Externally)~\cite{waniek2018hiding} deletes edges with the label, and adds edges with different labels. FGA (Fast Gradient Attack)~\cite{chen2018fast} calculates the gradient of dense adjacency matrix related to the classification loss and identifies the most vulnerable edges to perturb.  FLIP (Flipping attack)~\cite{bojchevski2019adversarial} is a deterministic approach that first ranks all nodes in ascending order according to their degrees, then flips their edges from the lower degree nodes to higher degree nodes. NEA (Network Embedding Attack)~\cite{bojchevski2019adversarial} is a black-box attack originally designed for attacking Deepwalk. STACK (Strict Black-box Attack)~\cite{xu2020query} does not require any knowledge of the target model and does not need training a surrogate model. It uses a generic graph filter unifying different GML models as an estimation for applying optimization attacks. PGD (Projected Gradient Descent)~\cite{madry2017towards} is originally an attack only modifying the features of inputs. Here, we adapt it to first randomly perturb edges of the graph, then optimize the features of target nodes by projected gradient descent. 

\textbf{Injection Attacks.} In \textit{graph injection} scenario, RND (Random) refers to randomly injecting new nodes with features randomly generated from the Gaussian distribution.  
FGSM (Fast Gradient Sign Method)~\cite{goodfellow2014explaining} linearizes the cost function around the current value of parameters, obtaining an optimal max-norm
constrained perturbation, which is called the ``fast gradient sign method'' of generating adversarial examples. We use an iterative version of FGSM to conduct a graph injection attack.
PGD (Projected Gradient Descent)~\cite{madry2017towards} is a universal ``first-order adversary'', \emph{i.e.}, the strongest attack utilizing the local first-order information about the network. The feature initialization is different from FGSM. 
SPEIT~\cite{zheng2020kdd} is the first place solution in KDD-CUP 2020 Graph Adversarial Attack \& Defense competition. It consists of adversarial adjacent matrix generation and enhanced feature gradient attacks, which are designed as a universal black-box graph injection attack.
TDGIA (Topological Defective Graph Injection Attack)~\cite{zou2021effective} is an effective graph injection attack that tackles the topological defectiveness of graphs. By sequentially injecting malicious nodes around nodes that are topologically vulnerable, TDGIA can significantly influence the accuracy of GML models.

Since the proposed scenario in \name is a \textit{black-box} one, all the above attacks are first applied to a surrogate model (trained by the attackers themselves), and then transfer to the target model. As demonstrated in~\cite{zou2021effective}, the choice of surrogate model will influence the transferability of attacks. When using raw GCN as the surrogate model, attacks can generally achieve better performance. Thus in \name experiments, we use GCN as the surrogate model for all attacks. Nevertheless, we encourage future researchers to further investigate the effect of transferability by testing other methods. 

\subsubsection{Defenses}\label{app:defense}

\begin{wrapfigure}{r}{0.3\textwidth}
\vspace{-0.5cm}
\centering
\includegraphics[width=0.3\textwidth]{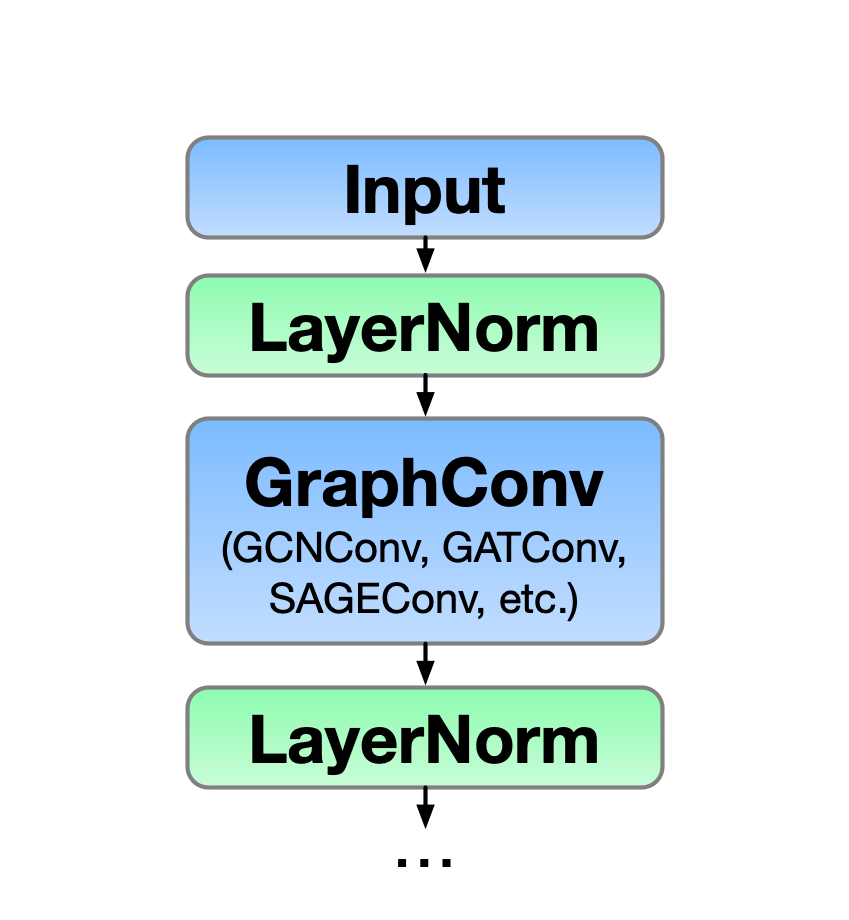}
\caption{The proposed layer normalization in \name. It is applied on the input and after every graph convolutional layer except the last one.}
\label{fig:grb_ln}
% \vspace{-1cm}
\end{wrapfigure}

GNN-SVD~\cite{entezari2020all} utilizes a low-rank approximation of the graph, that uses only the top singular components for its reconstruction. 
GNNGuard~\cite{zhang2020gnnguard} introduces the neighbor importance estimation and the layer-wise graph memory for defenses.
RobustGCN (R-GCN)~\cite{zheng2019robust} is a GCN variant that is specially designed against adversarial attacks on graphs. It adapts the random perturbation of features from VAE~\cite{kingma2013auto} and encodes both the mean and variance of the node representation thus makes the GNNs more robust. However, we found that methods like GNN-SVD and GNNGuard are not scalable to large-scale graphs due to the calculation of large dense matrices. To have stronger baseline defenses, we propose two methods that are scalable and can generally improve the performance of GML models.

\textbf{The adapted layer normalization (LN).} LN~\cite{ba2016layer} computes the mean and variance used for normalization from all of the summed inputs to the neurons in a layer on a single training case. It is originally used to stabilize the hidden state dynamics in recurrent networks. We found that it can also help to improve the adversarial robustness of GML models. Unlike the original version that is only used after hidden layers, we use LN first on the input features, and then after every graph convolutional layer except the last one. The process of the proposed LN is illustrated in~\figurename~\ref{fig:grb_ln}. The experiment results in Section~\ref{sec:exp} show that the proposed LN can generally improve the adversarial robustness of different types of GML models. 

\textbf{The adapted adversarial training (AT) in \textit{graph injection} scenario.} The AT~\cite{madry2017towards} is originally designed for defending adversarial attacks in image classification. The idea is to generate adversarial examples during training to change the classification distribution of models, which makes it difficult to perturb the results. Previous works ~\cite{feng2019graph} show that AT is also helpful for GML models, but it only considers the problem of graph modification attack, where the original graph can be modified. In our case, the defense is against graph injection attack, thus we propose a variant of AT that conduct graph injection attack during training. The procedure of the proposed AT is as following: (1) Initialization: the training graph is first used to train GML models for a few iterations as a warm-up. (2) FGSM attack: we conduct FGSM attack for a few steps on the current model to inject malicious nodes to attack training nodes. (3) Update gradients: we then train on the injected graph and minimize the classification loss of training nodes (excluding the injected nodes). (4) Repetition: we repeat this AT process until the training loss converges. Finally, we are able to construct more robust GML models. Interestingly, we found that AT with FGSM can also defend against other attacks, which shows great generality. Besides, the proposed AT can be easily adapt to any kind of GML models and scalable to large graphs. 

\begin{table}[!ht]
\caption{Hyper-parameters for adversarial training for five datasets.}
\centering
\begin{tabular}{ccccccc} 
\toprule
\textbf{Dataset} & \textbf{Attack} & \textbf{Step size} & \textbf{\# Steps/Iter} & \textbf{\# Injection} & \textbf{\# Edges} & \textbf{Feature range} \\
\textit{grb-cora} & FGSM & 0.01 & 10 & 20 & 20 & {[}-0.94, 0.94] \\
\textit{grb-citeseer} & FGSM & 0.01 & 10 & 30 & 20 & {[}-0.96, 0.89] \\
\textit{grb-flickr} & FGSM & 0.01 & 10 & 200 & 100 & {[}-0.47, 0.99] \\
\textit{grb-reddit} & FGSM & 0.01 & 10 & 500 & 200 & {[}-0.98, 0.99] \\
\textit{grb-aminer} & FGSM & 0.01 & 10 & 500 & 100 & {[}-0.93, 0.93] \\
\bottomrule
\end{tabular}
\label{tab:params_at}
\end{table}

\subsection{Reproducibility}\label{app:reproducible}

Reproducibility is one of the main features of \name. For reproducing results on leaderboards, all necessary components are available, including model weights, attack parameters, generated adversarial results, etc. Besides, \name provides scripts that allow users to reproduce results by a single command line. All codes are available in {\textcolor{blue}{\url{https://github.com/THUDM/grb}}}, where the implementation details and examples can be found. \name also provides full documentation for each module and function. All experiments can be reproduced in a single NVIDIA V100 GPU (with 32 GB memory). 

\subsubsection{Hyper-Parameter Settings}

\textbf{Hyper-Parameters of GML Models and Defenses.} The hyper-parameters of vanilla GML models and defenses are shown in~\tablename~\ref{tab:grb-cora-params}, \ref{tab:grb-citeseer-params}, \ref{tab:grb-flickr-params}, \ref{tab:grb-reddit-params}, \ref{tab:grb-aminer-params}, where GCNGuard stands for GCN+GNNGuard, GATGuard for GAT+GNNGuard, GCN-SVD for GCN+GNN-SVD, and LN for the proposed layer normalization. For the proposed adversarial training (AT), the hyper-parameters are shown in~\tablename~\ref{tab:params_at}. Under the proposed AT, GML models are trained while being continuously attacked by FGSM attack for a few steps per training iterations. Note that in each iteration, the attack is independent of previous iterations, only based on the weights of the model in the current iteration. The other hyper-parameters are exactly the same as training GML models. 

\begin{table}[!htp]
\begin{minipage}[t]{0.49\textwidth}
\caption{Hyper-parameters of GML models for \textit{grb-cora} dataset.}
\centering
\scalebox{0.5}{
\begin{tabular}{ccccccc} 
\toprule
\textbf{Model} & \textbf{\#Params} & \textbf{Hidden sizes} & \textbf{LR} & \textbf{Dropout} & \textbf{Optimizer} & \textbf{Others} \\
\midrule
GCN & 28,167 & 64, 64, 64 & 0.01 & 0.5 & Adam &  \\
GCN+LN & 29,027 & 64, 64, 64 & 0.01 & 0.5 & Adam &  \\
SAGE & 160,320 & 64, 64, 64 & 0.01 & 0.5 & Adam & full-batch \\
SAGE+LN & 161,180 & 64, 64, 64 & 0.01 & 0.5 & Adam & full-batch \\
SGCN & 28,771 & 64, 64, 64 & 0.01 & 0.5 & Adam & k=4 \\
SGCN+LN & 29,027 & 64, 64, 64 & 0.01 & 0.5 & Adam & k=4 \\
R-GCN & 56,334 & 64, 64, 64 & 0.01 & 0.5 & Adam &  \\
TAGCN & 84,103 & 64, 64, 64 & 0.01 & 0.5 & Adam & k=2 \\
TAGCN+LN & 84,963 & 64, 64, 64 & 0.01 & 0.5 & Adam & k=2 \\
GAT & 217,940 & 64, 64, 64 & 0.01 & 0.5 & Adam & num\_heads=4 \\
GAT+LN & 219,568 & 64, 64, 64 & 0.01 & 0.5 & Adam & num\_heads=4 \\
APPNP & 19,847 & 64 & 0.01 & 0.5 & Adam & alpha=0.01, k=10 \\
APPNP+LN & 20,579 & 64 & 0.01 & 0.5 & Adam & alpha=0.01, k=10 \\
GIN & 45,194 & 64, 64, 64 & 0.01 & 0.5 & Adam &  \\
GIN+LN & 46,054 & 64, 64, 64 & 0.01 & 0.5 & Adam &  \\
GCNGuard & 24,010 & 64, 64 & 0.001 & 0.1 & Adam &  \\
GATGuard & 151,639 & 64, 64 & 0.001 & 0.1 & Adam & num\_heads=4 \\
GCN-SVD & 24,007 & 64, 64 & 0.01 & 0.5 & Adam &  \\
\bottomrule
\end{tabular}}
\label{tab:grb-cora-params}
\end{minipage}%
\hspace{\fill}
\begin{minipage}[t]{0.49\textwidth}
\caption{Hyper-parameters of GML models for \textit{grb-citeseer} dataset.}
\centering
\scalebox{0.5}{
\begin{tabular}{ccccccc} 
\toprule
\textbf{Model} & \textbf{\#Params} & \textbf{Hidden sizes} & \textbf{LR} & \textbf{Dropout} & \textbf{Optimizer} & \textbf{Others} \\
\midrule
GCN & 57,926 & 64, 64, 64 & 0.01 & 0.5 & Adam &  \\
GCN+LN & 59,718 & 64, 64, 64 & 0.01 & 0.5 & Adam &  \\
SAGE & 718,924 & 64, 64, 64 & 0.01 & 0.5 & Adam & full batch \\
SAGE+LN & 720,716 & 64, 64, 64 & 0.01 & 0.5 & Adam & full batch \\
SGCN & 59,462 & 64, 64, 64 & 0.01 & 0.5 & Adam & k=4 \\
SGCN+LN & 59,718 & 64, 64, 64 & 0.01 & 0.5 & Adam & k=4 \\
R-GCN & 115,852 & 64, 64, 64 & 0.01 & 0.5 & Adam &  \\
TAGCN & 173,382 & 64, 64, 64 & 0.01 & 0.5 & Adam & k=2 \\
TAGCN+LN & 175,174 & 64, 64, 64 & 0.01 & 0.5 & Adam & k=2 \\
GAT & 336,200 & 64, 64, 64 & 0.01 & 0.5 & Adam & num\_heads=4 \\
GAT+LN & 338,760 & 64, 64, 64 & 0.01 & 0.5 & Adam & num\_heads=4 \\
APPNP & 49,606 & 64 & 0.01 & 0.5 & Adam & alpha=0.01, k=10 \\
APPNP+LN & 51,270 & 64 & 0.01 & 0.5 & Adam & alpha=0.01, k=10 \\
GIN & 74,953 & 64, 64, 64 & 0.01 & 0.5 & Adam &  \\
GIN+LN & 76,745 & 64, 64, 64 & 0.01 & 0.5 & Adam &  \\
GCNGuard & 53,769 & 64, 64 & 0.001 & 0.1 & Adam &  \\
GATGuard & 269,899 & 64, 64 & 0.001 & 0.1 & Adam & num\_heads=4 \\
GCN-SVD & 53,766 & 64, 64 & 0.01 & 0.5 & Adam &  \\
\bottomrule
\end{tabular}}
\label{tab:grb-citeseer-params}
\end{minipage}
\end{table}

\begin{table}[!htp]
\begin{minipage}[t]{0.49\textwidth}
\caption{Hyper-parameters of GML models for \textit{grb-flickr} dataset.}
\centering
\scalebox{0.5}{
\begin{tabular}{ccccccc} 
\toprule
\textbf{Model} & \textbf{\#Params} & \textbf{Hidden sizes} & \textbf{LR} & \textbf{Dropout} & \textbf{Optimizer} & \textbf{Others} \\
\midrule
GCN & 169,863 & 256, 128, 64 & 0.01 & 0.5 & Adam &  \\
GCN+LN & 171,631 & 256, 128, 64 & 0.01 & 0.5 & Adam &  \\
SAGE & 496,146 & 128, 128, 128 & 0.01 & 0.5 & Adam & full batch \\
SAGE+LN & 497,658 & 128, 128, 128 & 0.01 & 0.5 & Adam & full batch \\
R-GCN & 196,110 & 128, 128, 128 & 0.01 & 0.5 & Adam &  \\
TAGCN & 293,383 & 128, 128, 128 & 0.01 & 0.5 & Adam & k=2 \\
TAGCN+LN & 294,895 & 128, 128, 128 & 0.01 & 0.5 & Adam & k=2 \\
GAT & 799,316 & 128, 128, 128 & 0.01 & 0.5 & Adam & num\_heads=4 \\
GAT+LN & 802,364 & 128, 128, 128 & 0.01 & 0.5 & Adam & num\_heads=4 \\
APPNP & 65,031 & 128 & 0.01 & 0.5 & Adam & alpha=0.01, k=10 \\
APPNP+LN & 66,287 & 128 & 0.01 & 0.5 & Adam & alpha=0.01, k=10 \\
GIN & 164,874 & 128, 128, 128 & 0.01 & 0.5 & Adam &  \\
GIN+LN & 166,386 & 128, 128, 128 & 0.01 & 0.5 & Adam &  \\
GCNGuard & 81,546 & 128, 128 & 0.001 & 0.1 & Adam &  \\
\bottomrule
\end{tabular}}
\label{tab:grb-flickr-params}
\end{minipage}%
\hspace{\fill}
\begin{minipage}[t]{0.49\textwidth}
\caption{Hyper-parameters of GML models for \textit{grb-reddit} dataset.}
\centering
\scalebox{0.5}{
\begin{tabular}{ccccccc} 
\toprule
\textbf{Model} & \textbf{\#Params} & \textbf{Hidden sizes} & \textbf{LR} & \textbf{Dropout} & \textbf{Optimizer} & \textbf{Others} \\
\midrule
GCN & 115,497 & 128, 128, 128 & 0.01 & 0.5 & Adam &  \\
GCN+LN & 117,213 & 128, 128, 128 & 0.01 & 0.5 & Adam &  \\
SAGE & 643,536 & 128, 128, 128 & 0.01 & 0.5 & Adam & full batch \\
SAGE+LN & 645,252 & 128, 128, 128 & 0.01 & 0.5 & Adam & full batch \\
SGCN & 116,701 & 128, 128, 128 & 0.01 & 0.5 & Adam & k=4 \\
SGCN+LN & 117,213 & 128, 128, 128 & 0.01 & 0.5 & Adam & k=4 \\
R-GCN & 230,994 & 128, 128, 128 & 0.01 & 0.5 & Adam &  \\
TAGCN & 345,641 & 128, 128, 128 & 0.01 & 0.5 & Adam & k=2 \\
TAGCN+LN & 347,357 & 128, 128, 128 & 0.01 & 0.5 & Adam & k=2 \\
GAT & 104,950 & 64, 64 & 0.01 & 0.5 & Adam & num\_heads=2 \\
GAT+LN & 106,410 & 64, 64 & 0.01 & 0.5 & Adam & num\_heads=2 \\
APPNP & 82,473 & 128 & 0.01 & 0.5 & Adam & alpha=0.01, k=10 \\
APPNP+LN & 83,933 & 128 & 0.01 & 0.5 & Adam & alpha=0.01, k=10 \\
GIN & 182,316 & 128, 128, 128 & 0.01 & 0.5 & Adam &  \\
GIN+LN & 184,032 & 128, 128, 128 & 0.01 & 0.5 & Adam &  \\
\bottomrule
\end{tabular}}
\label{tab:grb-reddit-params}
\end{minipage}
\end{table}

\begin{table}[!htp]
\begin{minipage}[t]{0.49\textwidth}
\caption{Hyper-parameters of GML models for \textit{grb-aminer} dataset.}
\centering
\scalebox{0.5}{
\begin{tabular}{ccccccc} 
\toprule
\textbf{Model} & \textbf{\#Params} & \textbf{Hidden sizes} & \textbf{LR} & \textbf{Dropout} & \textbf{Optimizer} & \textbf{Others} \\
\midrule
GCN & 48,274 & 128, 128, 128 & 0.01 & 0.5 & Adam &  \\
GCN+LN & 48,986 & 128, 128, 128 & 0.01 & 0.5 & Adam &  \\
SAGE & 156,184 & 128, 128, 128 & 0.01 & 0.5 & Adam & full batch \\
SAGE+LN & 156,896 & 128, 128, 128 & 0.01 & 0.5 & Adam & full batch \\
SGCN & 48,474 & 128, 128, 128 & 0.01 & 0.5 & Adam & k=4 \\
SGCN+LN & 48,986 & 128, 128, 128 & 0.01 & 0.5 & Adam & k=4 \\
R-GCN & 96,548 & 128, 128, 128 & 0.01 & 0.5 & Adam &  \\
TAGCN & 144,018 & 128, 128, 128 & 0.01 & 0.5 & Adam & k=2 \\
TAGCN+LN & 144,730 & 128, 128, 128 & 0.01 & 0.5 & Adam & k=2 \\
GAT & 177,624 & 64, 64, 64 & 0.01 & 0.5 & Adam & num\_heads=2 \\
GAT+LN & 178,848 & 64, 64, 64 & 0.01 & 0.5 & Adam & num\_heads=2 \\
APPNP & 15,250 & 128 & 0.01 & 0.5 & Adam & alpha=0.01, k=10 \\
APPNP+LN & 15,706 & 128 & 0.01 & 0.5 & Adam & alpha=0.01, k=10 \\
GIN & 115,093 & 128, 128, 128 & 0.01 & 0.5 & Adam &  \\
GIN+LN & 115,805 & 128, 128, 128 & 0.01 & 0.5 & Adam &  \\
\bottomrule
\end{tabular}}
\label{tab:grb-aminer-params}
\end{minipage}%
\hspace{\fill}
\begin{minipage}[t]{0.49\textwidth}
\centering
\caption{Runtime (/s) of \textit{graph injection} attacks on large-scale graphs.}
\scalebox{0.6}{
\begin{tabular}{ccccccc} 
\toprule
\multicolumn{1}{l}{} & \textbf{Difficulty} & \multicolumn{1}{l}{\textbf{RAND}} & \multicolumn{1}{l}{\textbf{FGSM}} & \multicolumn{1}{l}{\textbf{PGD}} & \multicolumn{1}{l}{\textbf{SPEIT}} & \multicolumn{1}{l}{\textbf{TDGIA}} \\ 
\hline
\multirow{4}{*}{\textit{grb-reddit}} & F & 10.32 & 1110.92 & 1112.95 & 1119.10 & 6892.12 \\
 & H & 10.14 & 243.32 & 244.10 & 263.65 & 3179.26 \\
 & M & 9.97 & 1067.18 & 801.40 & 950.90 & 3126.48 \\
 & E & 11.64 & 304.03 & 305.44 & 319.04 & 4053.48 \\ 
\hline
\multirow{4}{*}{\textit{grb-aminer}} & F & 12.93 & 954.53 & 953.14 & 961.66 & 4079.35 \\
 & H & 12.87 & 212.41 & 213.44 & 233.08 & 2673.79 \\
 & M & 11.62 & 932.04 & 926.10 & 372.64 & 2612.44 \\
 & E & 12.61 & 218.53 & 219.93 & 239.57 & 2640.93 \\
\bottomrule
\end{tabular}}
\label{app:runtime}
\end{minipage}
\end{table}

\begin{table}
\centering
\caption{Hyper-parameters for attacks for five datasets in \textit{graph injection} scenario.}
\scalebox{0.9}{
\begin{tabular}{cccccccc} 
\toprule
\textbf{Dataset} & \textbf{Attack} & Step size & \textbf{\# Iter} & \begin{tabular}[c]{@{}c@{}}\textbf{\# Injection }\\\textbf{(E/M/H/F)}\end{tabular} & \textbf{\# Edges} & \begin{tabular}[c]{@{}c@{}}\textbf{Feature }\\\textbf{Range} \end{tabular} & \textbf{Others} \\
\midrule
\multirow{5}{*}{\textit{grb-cora}} & RND & - & 1 & \multirow{5}{*}{20/20/20/60} & \multirow{5}{*}{20} & \multirow{5}{*}{{[}-0.94, 0.94]} & \tiny{Random features} \\
 & PGD & 0.01 & 1000 &  &  &  & - \\
 & FGSM & 0.01 & 1000 &  &  &  & - \\
 & SPEIT & 0.01 & 1000 &  &  &  & - \\
 & TDGIA & 0.01 & 1000 &  &  &  & \tiny{Sequential} \\
\midrule
\multirow{5}{*}{\textit{grb-citeseer}} & RND & - & 1 & \multirow{5}{*}{30/30/30/90} & \multirow{5}{*}{20} & \multirow{5}{*}{{[}-0.96, 0.89]} & \tiny{Random features} \\
 & PGD & 0.01 & 1000 &  &  &  & - \\
 & FGSM & 0.01 & 1000 &  &  &  & - \\
 & SPEIT & 0.01 & 1000 &  &  &  & - \\
 & TDGIA & 0.01 & 1000 &  &  &  & \tiny{Sequential} \\
\midrule
\multirow{5}{*}{\textit{grb-flickr}} & RND & - & 1 & \multirow{5}{*}{200/200/200/600} & \multirow{5}{*}{100} & \multirow{5}{*}{{[}-0.47, 0.99]} & \tiny{Random features} \\
 & PGD & 0.01 & 2000 &  &  &  & - \\
 & FGSM & 0.01 & 2000 &  &  &  & - \\
 & SPEIT & 0.01 & 2000 &  &  &  & - \\
 & TDGIA & 0.01 & 2000 &  &  &  & \tiny{Sequential} \\
\midrule
\multirow{5}{*}{\textit{grb-reddit}} & RND & - & 1 & \multirow{5}{*}{500/500/500/1500} & \multirow{5}{*}{200} & \multirow{5}{*}{{[}-0.98, 0.99]} & \tiny{Random features} \\
 & PGD & 0.01 & 2000 &  &  &  & - \\
 & FGSM & 0.01 & 2000 &  &  &  & - \\
 & SPEIT & 0.01 & 2000 &  &  &  & - \\
 & TDGIA & 0.01 & 2000 &  &  &  & \tiny{Sequential} \\
\midrule
\multirow{5}{*}{\textit{grb-aminer}} & RND & - & 1 & \multirow{5}{*}{500/500/500/1500} & \multirow{5}{*}{100} & \multirow{5}{*}{{[}-0.93, 0.93]} & \tiny{Random features} \\
 & PGD & 0.01 & 5000 &  &  &  & - \\
 & FGSM & 0.01 & 5000 &  &  &  & - \\
 & SPEIT & 0.01 & 5000 &  &  &  & - \\
 & TDGIA & 0.01 & 5000 &  &  &  & \tiny{Sequential} \\
\bottomrule
\end{tabular}}
\label{tab:params_attack}
\end{table}

\textbf{Hyper-Parameters for Adversarial Attacks.} The hyper-parameters of attacks are shown in~\tablename~\ref{tab:params_attack}. For graph modification, following the most common setting in previous works, attackers are allowed to perturb a limited number of edges in the graph ($\Delta_{\mathcal{A}}$: the number of modified edges less than a ratio $\gamma_e$ of all edges). For graph injection, we follow the heuristic setting of KDDCUP 2020, attackers are allowed to inject new nodes with limited edges ($\Delta_{\mathcal{A}}$: less than $N_{n}$ injected nodes each with less than $N_{e}$ edges; $\Delta_{\mathcal{F}}$: constrained range of features $[\mathcal{F}_{min}, \mathcal{F}_{max}]$.). Nevertheless, more definitions of \textit{unnoticeabilty} can be developed by attackers and defenders when using \name. Since the proposed scenario in \name is a \textit{black-box} one, all the above attacks are first applied to a surrogate model (trained by the attackers themselves), and then transfer to the target model. As demonstrated in~\cite{zou2021effective}, the choice of surrogate model will influence the transferability of attacks. When using raw GCN as the surrogate model, attacks can generally achieve better performance. Thus in \name experiments, we use GCN as the surrogate model for all attacks. 

\subsection{Detailed Experiment Results}\label{app:result}

We conduct extensive experiments on all datasets and build a leaderboard for each dataset. Here we show the results of \textit{graph injection} scenario with Top-5 attacks vs. Top-10 defenses, full leaderboards can be found in {\textcolor{blue}{\url{https://cogdl.ai/grb/leaderboard}}}. Both attacks and defenses are ranked by the weighted accuracy, where {\color{red}red} and {\color{blue}blue} indicated the best results in each difficulty. It can be seen that different methods vary performance in different datasets. And it is also hard for attacks to be generally effective, especially in the presence of the proposed strong defense baselines. The runtime of attacks on large-scale graphs (\textit{grb-aminer}, \textit{grb-reddit}) can be found in~\tablename~\ref{app:runtime}.

\begin{table}[!htp]
\caption{\name leaderboard (Top 5 Attacks vs. Top 10 Defenses) for \textit{grb-cora} dataset.}
\centering
\scalebox{0.46}{
\begin{tabular}{cccccccccccccccc} 
\toprule
\multicolumn{3}{c}{\multirow{2}{*}{\diagbox{\textbf{Attack}}{\textbf{Defenses}}}} & \textbf{1} & \textbf{2} & \textbf{3} & \textbf{4} & \textbf{5} & \textbf{6} & \textbf{7} & \textbf{8} & \textbf{9} & \textbf{10} & \multirow{2}{*}{\begin{tabular}[c]{@{}c@{}}\textbf{Avg.}\\\textbf{Accuracy~}\end{tabular}} & \multirow{2}{*}{\begin{tabular}[c]{@{}c@{}}\textbf{Avg. 3-Max}\\\textbf{Accuracy}\end{tabular}} & \multirow{2}{*}{\begin{tabular}[c]{@{}c@{}}\textbf{Weighted}\\\textbf{Accuracy}\end{tabular}} \\
\multicolumn{3}{c}{} & \textbf{R-GCN\tiny{+AT}} & \textbf{GAT\tiny{+AT}} & \textbf{SGCN\tiny{+LN}} & \textbf{R-GCN} & \textbf{TAGCN\tiny{+LN}} & \textbf{GIN\tiny{+LN}} & \textbf{APPNP\tiny{+LN}} & \textbf{GIN\tiny{+AT}} & \textbf{GATGuard} & \textbf{GCN\tiny{+LN}} &  &  &  \\
\midrule
\multirow{4}{*}{\textbf{1}} & \multirow{4}{*}{\textbf{SPEIT}} & E & 79.97\tiny{{$\pm$}1.56} & 80.97\tiny{{$\pm$}1.23} & 66.16\tiny{{$\pm$}1.96} & 74.62\tiny{{$\pm$}2.80} & 63.28\tiny{{$\pm$}1.76} & 55.60\tiny{{$\pm$}2.13} & 71.57\tiny{{$\pm$}0.86} & 54.48\tiny{{$\pm$}1.76} & 64.93\tiny{{$\pm$}0.00} & 48.06\tiny{{$\pm$}2.38} & 52.01\tiny{{$\pm$}0.09} & 53.48\tiny{{$\pm$}0.25} & 53.44\tiny{{$\pm$}0.17} \\
 &  & M & 84.11\tiny{{$\pm$}0.38} & 84.55\tiny{{$\pm$}0.44} & 80.60\tiny{{$\pm$}1.18} & 81.16\tiny{{$\pm$}1.39} & 74.59\tiny{{$\pm$}1.86} & 64.10\tiny{{$\pm$}1.05} & 75.67\tiny{{$\pm$}1.12} & 63.36\tiny{{$\pm$}1.39} & 63.06\tiny{{$\pm$}0.00} & 69.18\tiny{{$\pm$}2.24} & 47.31\tiny{{$\pm$}0.09} & {\color{red}50.35\tiny{{$\pm$}1.25}} & {\color{red}50.89\tiny{{$\pm$}0.24}} \\
 &  & H & 88.51\tiny{{$\pm$}0.80} & 89.78\tiny{{$\pm$}1.00} & 89.74\tiny{{$\pm$}0.71} & 88.47\tiny{{$\pm$}0.54} & 89.85\tiny{{$\pm$}0.55} & 80.37\tiny{{$\pm$}1.10} & 80.64\tiny{{$\pm$}1.14} & 63.51\tiny{{$\pm$}3.22} & 69.03\tiny{{$\pm$}0.00} & 88.55\tiny{{$\pm$}1.32} & {\color{red}42.72\tiny{{$\pm$}0.07}} & {\color{red}47.34\tiny{{$\pm$}2.27}} & {\color{red}48.44\tiny{{$\pm$}0.12}} \\
 &  & F & 85.21\tiny{{$\pm$}0.41} & 85.35\tiny{{$\pm$}0.19} & 75.65\tiny{{$\pm$}0.87} & 79.85\tiny{{$\pm$}0.48} & 71.73\tiny{{$\pm$}1.14} & 64.75\tiny{{$\pm$}0.62} & 73.11\tiny{{$\pm$}0.76} & 63.05\tiny{{$\pm$}1.37} & 65.67\tiny{{$\pm$}0.00} & 59.45\tiny{{$\pm$}0.64} & {\color{red}45.86\tiny{{$\pm$}0.05}} & {\color{red}48.56\tiny{{$\pm$}0.98}} & {\color{red}48.95\tiny{{$\pm$}0.12}} \\
\midrule
\multirow{4}{*}{\textbf{2}} & \multirow{4}{*}{\textbf{TDGIA}} & E & 81.68\tiny{{$\pm$}1.81} & 80.52\tiny{{$\pm$}0.72} & 72.05\tiny{{$\pm$}3.19} & 68.13\tiny{{$\pm$}3.50} & 73.36\tiny{{$\pm$}3.86} & 68.77\tiny{{$\pm$}2.56} & 64.18\tiny{{$\pm$}1.71} & 63.02\tiny{{$\pm$}3.37} & 64.93\tiny{{$\pm$}0.00} & 65.93\tiny{{$\pm$}4.07} & {\color{red}50.12\tiny{{$\pm$}0.14}} & {\color{red}53.22\tiny{{$\pm$}1.01}} & {\color{red}53.30\tiny{{$\pm$}0.12}} \\
 &  & M & 83.96\tiny{{$\pm$}0.53} & 84.25\tiny{{$\pm$}0.40} & 81.49\tiny{{$\pm$}0.71} & 77.57\tiny{{$\pm$}1.86} & 82.17\tiny{{$\pm$}3.29} & 73.58\tiny{{$\pm$}2.81} & 74.44\tiny{{$\pm$}1.52} & 70.19\tiny{{$\pm$}1.84} & 63.06\tiny{{$\pm$}0.00} & 76.34\tiny{{$\pm$}2.16} & {\color{red}47.21\tiny{{$\pm$}0.06}} & 51.58\tiny{{$\pm$}0.22} & 51.06\tiny{{$\pm$}0.10} \\
 &  & H & 88.21\tiny{{$\pm$}0.18} & 90.30\tiny{{$\pm$}0.00} & 88.92\tiny{{$\pm$}0.80} & 86.83\tiny{{$\pm$}0.85} & 87.39\tiny{{$\pm$}1.81} & 84.52\tiny{{$\pm$}1.28} & 78.58\tiny{{$\pm$}1.30} & 80.63\tiny{{$\pm$}3.06} & 69.03\tiny{{$\pm$}0.00} & 86.64\tiny{{$\pm$}1.63} & 45.17\tiny{{$\pm$}0.07} & 48.53\tiny{{$\pm$}1.14} & 48.68\tiny{{$\pm$}0.06} \\
 &  & F & 84.43\tiny{{$\pm$}0.27} & 84.55\tiny{{$\pm$}0.50} & 77.39\tiny{{$\pm$}1.05} & 74.58\tiny{{$\pm$}1.76} & 79.67\tiny{{$\pm$}1.53} & 76.14\tiny{{$\pm$}1.80} & 68.16\tiny{{$\pm$}2.10} & 70.51\tiny{{$\pm$}1.58} & 65.67\tiny{{$\pm$}0.00} & 72.58\tiny{{$\pm$}2.71} & 46.24\tiny{{$\pm$}0.04} & 49.75\tiny{{$\pm$}0.84} & 49.73\tiny{{$\pm$}0.09} \\
\midrule
\multirow{4}{*}{\textbf{3}} & \multirow{4}{*}{\textbf{PGD}} & E & 83.02\tiny{{$\pm$}1.26} & 80.60\tiny{{$\pm$}1.04} & 73.88\tiny{{$\pm$}2.41} & 67.80\tiny{{$\pm$}2.24} & 74.78\tiny{{$\pm$}2.36} & 70.07\tiny{{$\pm$}1.79} & 66.19\tiny{{$\pm$}1.50} & 62.65\tiny{{$\pm$}1.33} & 64.93\tiny{{$\pm$}0.00} & 68.58\tiny{{$\pm$}3.00} & 50.11\tiny{{$\pm$}0.11} & 53.19\tiny{{$\pm$}0.98} & 53.29\tiny{{$\pm$}0.14} \\
 &  & M & 83.84\tiny{{$\pm$}1.15} & 84.81\tiny{{$\pm$}0.67} & 82.20\tiny{{$\pm$}1.18} & 78.51\tiny{{$\pm$}1.57} & 83.36\tiny{{$\pm$}1.39} & 77.35\tiny{{$\pm$}1.02} & 73.21\tiny{{$\pm$}1.83} & 69.25\tiny{{$\pm$}2.18} & 63.06\tiny{{$\pm$}0.00} & 77.80\tiny{{$\pm$}1.26} & 47.24\tiny{{$\pm$}0.08} & 51.64\tiny{{$\pm$}0.20} & 51.11\tiny{{$\pm$}0.09} \\
 &  & H & 88.88\tiny{{$\pm$}0.50} & 90.48\tiny{{$\pm$}0.67} & 89.96\tiny{{$\pm$}0.75} & 85.86\tiny{{$\pm$}0.83} & 89.48\tiny{{$\pm$}1.07} & 85.97\tiny{{$\pm$}0.82} & 78.47\tiny{{$\pm$}1.29} & 80.82\tiny{{$\pm$}0.89} & 69.03\tiny{{$\pm$}0.00} & 88.32\tiny{{$\pm$}0.56} & 45.18\tiny{{$\pm$}0.05} & 48.50\tiny{{$\pm$}1.14} & 48.68\tiny{{$\pm$}0.06} \\
 &  & F & 85.60\tiny{{$\pm$}0.38} & 85.43\tiny{{$\pm$}0.34} & 81.90\tiny{{$\pm$}0.90} & 76.77\tiny{{$\pm$}0.74} & 83.22\tiny{{$\pm$}0.60} & 78.58\tiny{{$\pm$}0.51} & 66.49\tiny{{$\pm$}0.68} & 71.30\tiny{{$\pm$}0.94} & 65.67\tiny{{$\pm$}0.00} & 78.21\tiny{{$\pm$}0.68} & 46.26\tiny{{$\pm$}0.04} & 49.81\tiny{{$\pm$}0.89} & 49.83\tiny{{$\pm$}0.08} \\
\midrule
\multirow{4}{*}{\textbf{4}} & \multirow{4}{*}{\textbf{FGSM}} & E & 82.35\tiny{{$\pm$}0.95} & 80.19\tiny{{$\pm$}0.97} & 73.95\tiny{{$\pm$}2.46} & 67.01\tiny{{$\pm$}1.40} & 74.03\tiny{{$\pm$}1.72} & 70.48\tiny{{$\pm$}1.60} & 64.67\tiny{{$\pm$}1.71} & 62.46\tiny{{$\pm$}2.30} & 64.93\tiny{{$\pm$}0.00} & 68.39\tiny{{$\pm$}1.27} & 52.72\tiny{{$\pm$}0.03} & 54.71\tiny{{$\pm$}0.33} & 54.71\tiny{{$\pm$}0.07} \\
 &  & M & 84.25\tiny{{$\pm$}1.24} & 85.11\tiny{{$\pm$}0.64} & 82.46\tiny{{$\pm$}0.93} & 77.87\tiny{{$\pm$}1.43} & 84.29\tiny{{$\pm$}1.02} & 78.77\tiny{{$\pm$}1.09} & 73.10\tiny{{$\pm$}1.13} & 69.74\tiny{{$\pm$}1.60} & 63.06\tiny{{$\pm$}0.00} & 78.47\tiny{{$\pm$}1.30} & 48.71\tiny{{$\pm$}0.16} & 51.81\tiny{{$\pm$}0.67} & 51.93\tiny{{$\pm$}0.09} \\
 &  & H & 89.22\tiny{{$\pm$}0.59} & 90.59\tiny{{$\pm$}0.40} & 90.30\tiny{{$\pm$}0.50} & 86.23\tiny{{$\pm$}1.16} & 89.55\tiny{{$\pm$}0.67} & 86.08\tiny{{$\pm$}0.78} & 78.80\tiny{{$\pm$}1.45} & 80.93\tiny{{$\pm$}0.91} & 69.03\tiny{{$\pm$}0.00} & 87.13\tiny{{$\pm$}0.96} & 43.58\tiny{{$\pm$}0.08} & 48.20\tiny{{$\pm$}1.74} & 48.84\tiny{{$\pm$}0.08} \\
 &  & F & 85.01\tiny{{$\pm$}0.41} & 85.33\tiny{{$\pm$}0.72} & 81.53\tiny{{$\pm$}1.27} & 76.62\tiny{{$\pm$}0.90} & 83.00\tiny{{$\pm$}0.61} & 78.26\tiny{{$\pm$}0.88} & 67.09\tiny{{$\pm$}1.18} & 71.64\tiny{{$\pm$}0.78} & 65.67\tiny{{$\pm$}0.00} & 77.60\tiny{{$\pm$}1.14} & 48.26\tiny{{$\pm$}0.03} & 51.45\tiny{{$\pm$}0.77} & 51.61\tiny{{$\pm$}0.06} \\
\midrule
\multirow{4}{*}{\textbf{5}} & \multirow{4}{*}{\textbf{RND}} & E & 82.28\tiny{{$\pm$}0.93} & 80.26\tiny{{$\pm$}1.24} & 76.57\tiny{{$\pm$}2.44} & 73.54\tiny{{$\pm$}1.44} & 78.36\tiny{{$\pm$}1.74} & 68.81\tiny{{$\pm$}0.82} & 67.95\tiny{{$\pm$}1.68} & 67.69\tiny{{$\pm$}1.05} & 64.93\tiny{{$\pm$}0.00} & 74.29\tiny{{$\pm$}1.86} & 52.31\tiny{{$\pm$}0.06} & 53.65\tiny{{$\pm$}0.24} & 53.65\tiny{{$\pm$}0.14} \\
 &  & M & 84.18\tiny{{$\pm$}1.00} & 84.44\tiny{{$\pm$}0.88} & 82.05\tiny{{$\pm$}0.72} & 79.33\tiny{{$\pm$}1.24} & 84.11\tiny{{$\pm$}0.34} & 76.68\tiny{{$\pm$}1.05} & 73.88\tiny{{$\pm$}1.64} & 72.72\tiny{{$\pm$}1.68} & 63.06\tiny{{$\pm$}0.00} & 80.11\tiny{{$\pm$}1.08} & 48.89\tiny{{$\pm$}0.05} & 51.70\tiny{{$\pm$}0.32} & 51.57\tiny{{$\pm$}0.11} \\
 &  & H & 88.99\tiny{{$\pm$}0.51} & 90.71\tiny{{$\pm$}0.31} & 90.19\tiny{{$\pm$}0.41} & 87.20\tiny{{$\pm$}0.75} & 90.04\tiny{{$\pm$}0.24} & 84.52\tiny{{$\pm$}0.73} & 80.03\tiny{{$\pm$}1.11} & 82.87\tiny{{$\pm$}0.83} & 69.03\tiny{{$\pm$}0.00} & 89.29\tiny{{$\pm$}0.75} & 44.74\tiny{{$\pm$}0.07} & 49.19\tiny{{$\pm$}0.26} & 48.77\tiny{{$\pm$}0.18} \\
 &  & F & 85.36\tiny{{$\pm$}0.41} & 84.95\tiny{{$\pm$}0.58} & 82.85\tiny{{$\pm$}1.29} & 79.53\tiny{{$\pm$}0.74} & 84.22\tiny{{$\pm$}0.58} & 76.75\tiny{{$\pm$}0.89} & 68.93\tiny{{$\pm$}0.92} & 74.11\tiny{{$\pm$}0.71} & 65.67\tiny{{$\pm$}0.00} & 81.34\tiny{{$\pm$}0.60} & 48.32\tiny{{$\pm$}0.04} & 50.98\tiny{{$\pm$}0.17} & 50.74\tiny{{$\pm$}0.07} \\
\midrule
\multirow{4}{*}{\textbf{6}} & \multirow{4}{*}{\textbf{W/O Attack}} & E & 84.70\tiny{{$\pm$}0.00} & 81.34\tiny{{$\pm$}0.00} & 81.72\tiny{{$\pm$}0.00} & 82.09\tiny{{$\pm$}0.00} & 79.10\tiny{{$\pm$}0.00} & 70.15\tiny{{$\pm$}0.00} & 77.99\tiny{{$\pm$}0.00} & 68.28\tiny{{$\pm$}0.00} & 64.93\tiny{{$\pm$}0.00} & 79.10\tiny{{$\pm$}0.00} & 52.04\tiny{{$\pm$}0.01} & 53.70\tiny{{$\pm$}0.88} & 54.15\tiny{{$\pm$}0.08} \\
 &  & M & 83.96\tiny{{$\pm$}0.00} & 84.33\tiny{{$\pm$}0.00} & 82.84\tiny{{$\pm$}0.00} & 83.21\tiny{{$\pm$}0.00} & 85.45\tiny{{$\pm$}0.00} & 76.87\tiny{{$\pm$}0.00} & 82.46\tiny{{$\pm$}0.00} & 73.51\tiny{{$\pm$}0.00} & 63.06\tiny{{$\pm$}0.00} & 81.72\tiny{{$\pm$}0.00} & 49.67\tiny{{$\pm$}0.01} & 52.67\tiny{{$\pm$}0.50} & 52.65\tiny{{$\pm$}0.09} \\
 &  & H & 89.55\tiny{{$\pm$}0.00} & 91.04\tiny{{$\pm$}0.00} & 91.04\tiny{{$\pm$}0.00} & 89.18\tiny{{$\pm$}0.00} & 90.67\tiny{{$\pm$}0.00} & 84.70\tiny{{$\pm$}0.00} & 88.06\tiny{{$\pm$}0.00} & 82.84\tiny{{$\pm$}0.00} & 69.03\tiny{{$\pm$}0.00} & 89.93\tiny{{$\pm$}0.00} & 45.94\tiny{{$\pm$}0.01} & 50.01\tiny{{$\pm$}0.56} & 49.99\tiny{{$\pm$}0.00} \\
 &  & F & 86.07\tiny{{$\pm$}0.00} & 85.57\tiny{{$\pm$}0.00} & 85.20\tiny{{$\pm$}0.00} & 84.83\tiny{{$\pm$}0.00} & 85.07\tiny{{$\pm$}0.00} & 77.24\tiny{{$\pm$}0.00} & 82.84\tiny{{$\pm$}0.00} & 74.88\tiny{{$\pm$}0.00} & 65.67\tiny{{$\pm$}0.00} & 83.58\tiny{{$\pm$}0.00} & 49.21\tiny{{$\pm$}0.01} & 51.82\tiny{{$\pm$}0.33} & 51.77\tiny{{$\pm$}0.04} \\
\midrule
\multicolumn{2}{c}{\multirow{4}{*}{\begin{tabular}[c]{@{}c@{}}\textbf{Avg.}\\\textbf{Accuracy}\end{tabular}}} & E & {\color{blue}82.33\tiny{{$\pm$}0.40}} & 80.65\tiny{{$\pm$}0.54} & 74.05\tiny{{$\pm$}1.05} & 72.20\tiny{{$\pm$}1.01} & 73.82\tiny{{$\pm$}0.78} & 67.31\tiny{{$\pm$}0.56} & 68.76\tiny{{$\pm$}0.54} & 63.10\tiny{{$\pm$}0.38} & 64.93\tiny{{$\pm$}0.00} & 67.39\tiny{{$\pm$}0.84} & - & - & - \\
\multicolumn{2}{c}{} & M & 84.05\tiny{{$\pm$}0.34} & {\color{blue}84.58\tiny{{$\pm$}0.21}} & 81.94\tiny{{$\pm$}0.30} & 79.61\tiny{{$\pm$}0.60} & 82.33\tiny{{$\pm$}0.49} & 74.56\tiny{{$\pm$}0.45} & 75.46\tiny{{$\pm$}0.50} & 69.80\tiny{{$\pm$}0.92} & 63.06\tiny{{$\pm$}0.00} & 77.27\tiny{{$\pm$}0.49} & - & - & - \\
\multicolumn{2}{c}{} & H & 88.89\tiny{{$\pm$}0.21} & {\color{blue}90.49\tiny{{$\pm$}0.18}} & 90.02\tiny{{$\pm$}0.27} & 87.29\tiny{{$\pm$}0.33} & 89.50\tiny{{$\pm$}0.41} & 84.36\tiny{{$\pm$}0.36} & 80.77\tiny{{$\pm$}0.23} & 78.60\tiny{{$\pm$}0.96} & 69.03\tiny{{$\pm$}0.00} & 88.31\tiny{{$\pm$}0.38} & - & - & - \\
\multicolumn{2}{c}{} & F & {\color{blue}85.28\tiny{{$\pm$}0.15}} & 85.20\tiny{{$\pm$}0.21} & 80.75\tiny{{$\pm$}0.41} & 78.69\tiny{{$\pm$}0.27} & 81.15\tiny{{$\pm$}0.42} & 75.29\tiny{{$\pm$}0.49} & 71.10\tiny{{$\pm$}0.40} & 70.91\tiny{{$\pm$}0.43} & 65.67\tiny{{$\pm$}0.00} & 75.46\tiny{{$\pm$}0.49} & - & - & - \\
\midrule
\multicolumn{2}{c}{\multirow{4}{*}{\begin{tabular}[c]{@{}c@{}}\textbf{Avg. 3-Min} \\\textbf{Accuracy}\end{tabular}}} & E & {\color{blue}80.93\tiny{{$\pm$}0.61}} & 79.91\tiny{{$\pm$}0.61} & 70.14\tiny{{$\pm$}1.74} & 67.61\tiny{{$\pm$}1.32} & 69.58\tiny{{$\pm$}1.09} & 63.92\tiny{{$\pm$}0.83} & 64.89\tiny{{$\pm$}0.88} & 59.17\tiny{{$\pm$}0.66} & 64.93\tiny{{$\pm$}0.00} & 60.34\tiny{{$\pm$}1.48} & - & - & - \\
\multicolumn{2}{c}{} & M & 83.46\tiny{{$\pm$}0.31} & {\color{blue}84.14\tiny{{$\pm$}0.22}} & 81.07\tiny{{$\pm$}0.36} & 77.74\tiny{{$\pm$}0.93} & 79.90\tiny{{$\pm$}0.82} & 71.29\tiny{{$\pm$}1.07} & 73.07\tiny{{$\pm$}0.83} & 67.10\tiny{{$\pm$}1.28} & 63.06\tiny{{$\pm$}0.00} & 74.28\tiny{{$\pm$}0.90} & - & - & - \\
\multicolumn{2}{c}{} & H & 88.36\tiny{{$\pm$}0.28} & {\color{blue}90.06\tiny{{$\pm$}0.34}} & 89.39\tiny{{$\pm$}0.37} & 86.17\tiny{{$\pm$}0.53} & 88.58\tiny{{$\pm$}0.80} & 82.97\tiny{{$\pm$}0.42} & 78.41\tiny{{$\pm$}0.38} & 74.66\tiny{{$\pm$}1.77} & 69.03\tiny{{$\pm$}0.00} & 87.16\tiny{{$\pm$}0.61} & - & - & - \\
\multicolumn{2}{c}{} & F & {\color{blue}84.79\tiny{{$\pm$}0.23}} & 84.77\tiny{{$\pm$}0.34} & 78.01\tiny{{$\pm$}0.50} & 75.99\tiny{{$\pm$}0.42} & 78.07\tiny{{$\pm$}0.78} & 72.46\tiny{{$\pm$}0.81} & 67.10\tiny{{$\pm$}0.76} & 68.12\tiny{{$\pm$}0.65} & 65.67\tiny{{$\pm$}0.00} & 69.83\tiny{{$\pm$}0.88} & - & - & - \\
\midrule
\multicolumn{2}{c}{\multirow{4}{*}{\begin{tabular}[c]{@{}c@{}}\textbf{Weighted} \\\textbf{Accuracy}\end{tabular}}} & E & {\color{blue}80.19\tiny{{$\pm$}0.86}} & 79.62\tiny{{$\pm$}0.61} & 68.40\tiny{{$\pm$}1.64} & 66.85\tiny{{$\pm$}1.66} & 66.67\tiny{{$\pm$}1.12} & 59.85\tiny{{$\pm$}1.37} & 64.50\tiny{{$\pm$}0.99} & 57.05\tiny{{$\pm$}1.00} & 64.93\tiny{{$\pm$}0.00} & 54.51\tiny{{$\pm$}1.73} & - & - & - \\
\multicolumn{2}{c}{} & M & 83.27\tiny{{$\pm$}0.41} & {\color{blue}84.00\tiny{{$\pm$}0.30}} & 80.70\tiny{{$\pm$}0.44} & 77.37\tiny{{$\pm$}1.24} & 77.02\tiny{{$\pm$}1.13} & 67.76\tiny{{$\pm$}0.85} & 72.63\tiny{{$\pm$}1.04} & 65.40\tiny{{$\pm$}1.27} & 63.06\tiny{{$\pm$}0.00} & 71.83\tiny{{$\pm$}1.38} & - & - & - \\
\multicolumn{2}{c}{} & H & 88.22\tiny{{$\pm$}0.31} & {\color{blue}89.80\tiny{{$\pm$}0.60}} & 89.15\tiny{{$\pm$}0.49} & 85.91\tiny{{$\pm$}0.56} & 88.05\tiny{{$\pm$}1.28} & 81.70\tiny{{$\pm$}0.69} & 78.00\tiny{{$\pm$}0.67} & 69.13\tiny{{$\pm$}2.45} & 69.03\tiny{{$\pm$}0.00} & 86.71\tiny{{$\pm$}0.94} & - & - & - \\
\multicolumn{2}{c}{} & F & {\color{blue}84.65\tiny{{$\pm$}0.23}} & 84.62\tiny{{$\pm$}0.33} & 76.86\tiny{{$\pm$}0.58} & 75.45\tiny{{$\pm$}0.96} & 74.96\tiny{{$\pm$}0.91} & 68.64\tiny{{$\pm$}0.68} & 66.79\tiny{{$\pm$}0.98} & 65.73\tiny{{$\pm$}0.89} & 65.67\tiny{{$\pm$}0.00} & 64.82\tiny{{$\pm$}0.60} & - & - & - \\
\bottomrule
\end{tabular}}
\label{tab:leaderboard_cora}
\end{table}

\begin{table}[!htp]
\caption{\textit{grb-citeseer} leaderboard (Top 5 ATK.\ vs.\ Top 10 DEF.) in \textit{graph injection} scenario.}
\centering
\scalebox{0.46}{
\begin{tabular}{cccccccccccccccc} 
\toprule
\multicolumn{3}{c}{\multirow{2}{*}{\diagbox{\textbf{Attack}}{\textbf{Defenses}}}} & \textbf{1} & \textbf{2} & \textbf{3} & \textbf{4} & \textbf{5} & \textbf{6} & \textbf{7} & \textbf{8} & \textbf{9} & \textbf{10} & \multirow{2}{*}{\begin{tabular}[c]{@{}c@{}}\textbf{Avg.}\\\textbf{Accuracy~}\end{tabular}} & \multirow{2}{*}{\begin{tabular}[c]{@{}c@{}}\textbf{Avg. 3-Max}\\\textbf{Accuracy}\end{tabular}} & \multirow{2}{*}{\begin{tabular}[c]{@{}c@{}}\textbf{Weighted}\\\textbf{Accuracy}\end{tabular}} \\
\multicolumn{3}{c}{} & \textbf{GAT\tiny{+AT}} & \textbf{R-GCN\tiny{+AT}} & \textbf{SAGE\tiny{AT}} & \textbf{GATGuard} & \textbf{GCNGuard} & \textbf{SGCN\tiny{+LN}} & \textbf{R-GCN} & \textbf{GIN\tiny{+LN}} & \textbf{TAGCN\tiny{+LN}} & \textbf{GIN\tiny{+AT}} &  &  &  \\
\midrule
\multirow{4}{*}{\textbf{1}} & \multirow{4}{*}{\textbf{SPEIT}} & E & 70.88\tiny{{$\pm$}0.83} & 67.49\tiny{{$\pm$}1.01} & 68.68\tiny{{$\pm$}1.64} & 65.52\tiny{{$\pm$}0.00} & 56.43\tiny{{$\pm$}0.00} & 49.78\tiny{{$\pm$}2.57} & 52.29\tiny{{$\pm$}2.76} & 48.53\tiny{{$\pm$}1.60} & 44.89\tiny{{$\pm$}4.98} & 47.30\tiny{{$\pm$}8.94} & 52.01\tiny{{$\pm$}0.09} & 53.48\tiny{{$\pm$}0.25} & 53.44\tiny{{$\pm$}0.17} \\
 &  & M & 72.16\tiny{{$\pm$}0.44} & 71.00\tiny{{$\pm$}1.12} & 70.81\tiny{{$\pm$}1.79} & 63.64\tiny{{$\pm$}0.00} & 62.70\tiny{{$\pm$}0.00} & 55.05\tiny{{$\pm$}1.74} & 61.63\tiny{{$\pm$}3.24} & 49.78\tiny{{$\pm$}2.12} & 52.66\tiny{{$\pm$}4.65} & 47.34\tiny{{$\pm$}9.06} & 47.31\tiny{{$\pm$}0.09} & 50.35\tiny{{$\pm$}1.25} & {\color{red}50.89\tiny{{$\pm$}0.24}} \\
 &  & H & 78.18\tiny{{$\pm$}0.49} & 79.69\tiny{{$\pm$}0.37} & 77.46\tiny{{$\pm$}0.67} & 72.10\tiny{{$\pm$}0.00} & 74.64\tiny{{$\pm$}0.09} & 66.87\tiny{{$\pm$}1.73} & 77.08\tiny{{$\pm$}1.91} & 56.86\tiny{{$\pm$}1.31} & 69.03\tiny{{$\pm$}4.80} & 51.10\tiny{{$\pm$}6.38} & {\color{red}42.72\tiny{{$\pm$}0.07}} & {\color{red}47.34\tiny{{$\pm$}2.27}} & {\color{red}48.44\tiny{{$\pm$}0.12}} \\
 &  & F & 73.85\tiny{{$\pm$}0.22} & 71.30\tiny{{$\pm$}0.64} & 71.62\tiny{{$\pm$}1.39} & 67.08\tiny{{$\pm$}0.00} & 64.54\tiny{{$\pm$}0.13} & 53.64\tiny{{$\pm$}1.43} & 59.32\tiny{{$\pm$}3.65} & 52.82\tiny{{$\pm$}0.93} & 51.09\tiny{{$\pm$}6.38} & 44.71\tiny{{$\pm$}9.62} & {\color{red}45.86\tiny{{$\pm$}0.05}} & {\color{red}48.56\tiny{{$\pm$}0.98}} & {\color{red}48.95\tiny{{$\pm$}0.12}} \\
\midrule
\multirow{4}{*}{\textbf{2}} & \multirow{4}{*}{\textbf{TDGIA}} & E & 71.00\tiny{{$\pm$}0.67} & 66.68\tiny{{$\pm$}1.30} & 68.18\tiny{{$\pm$}1.33} & 65.52\tiny{{$\pm$}0.00} & 56.43\tiny{{$\pm$}0.00} & 57.15\tiny{{$\pm$}1.87} & 48.81\tiny{{$\pm$}3.56} & 53.92\tiny{{$\pm$}2.98} & 47.27\tiny{{$\pm$}2.58} & 51.95\tiny{{$\pm$}10.85} & {\color{red}50.12\tiny{{$\pm$}0.14}} & {\color{red}53.22\tiny{{$\pm$}1.01}} & {\color{red}53.30\tiny{{$\pm$}0.12}} \\
 &  & M & 72.54\tiny{{$\pm$}0.47} & 71.76\tiny{{$\pm$}1.21} & 71.13\tiny{{$\pm$}1.10} & 63.64\tiny{{$\pm$}0.00} & 62.70\tiny{{$\pm$}0.00} & 60.91\tiny{{$\pm$}0.95} & 60.97\tiny{{$\pm$}2.82} & 49.22\tiny{{$\pm$}3.91} & 49.09\tiny{{$\pm$}6.61} & 51.69\tiny{{$\pm$}7.68} & {\color{red}47.21\tiny{{$\pm$}0.06}} & 51.58\tiny{{$\pm$}0.22} & 51.06\tiny{{$\pm$}0.10} \\
 &  & H & 78.28\tiny{{$\pm$}0.37} & 79.72\tiny{{$\pm$}0.49} & 77.27\tiny{{$\pm$}0.87} & 72.10\tiny{{$\pm$}0.00} & 74.61\tiny{{$\pm$}0.00} & 70.50\tiny{{$\pm$}1.89} & 73.92\tiny{{$\pm$}1.91} & 57.80\tiny{{$\pm$}4.13} & 65.61\tiny{{$\pm$}3.54} & 66.08\tiny{{$\pm$}6.43} & 45.17\tiny{{$\pm$}0.07} & 48.53\tiny{{$\pm$}1.14} & 48.68\tiny{{$\pm$}0.06} \\
 &  & F & 73.89\tiny{{$\pm$}0.40} & 71.75\tiny{{$\pm$}0.78} & 72.25\tiny{{$\pm$}0.62} & 67.08\tiny{{$\pm$}0.00} & 64.58\tiny{{$\pm$}0.00} & 59.76\tiny{{$\pm$}2.10} & 56.69\tiny{{$\pm$}1.77} & 52.16\tiny{{$\pm$}2.47} & 46.87\tiny{{$\pm$}4.99} & 59.01\tiny{{$\pm$}5.47} & 46.24\tiny{{$\pm$}0.04} & 49.75\tiny{{$\pm$}0.84} & 49.73\tiny{{$\pm$}0.09} \\
\midrule
\multirow{4}{*}{\textbf{3}} & \multirow{4}{*}{\textbf{FGSM}} & E & 71.44\tiny{{$\pm$}1.03} & 69.00\tiny{{$\pm$}0.75} & 69.28\tiny{{$\pm$}0.73} & 65.52\tiny{{$\pm$}0.00} & 56.46\tiny{{$\pm$}0.17} & 52.38\tiny{{$\pm$}1.56} & 49.87\tiny{{$\pm$}2.20} & 52.38\tiny{{$\pm$}2.41} & 50.88\tiny{{$\pm$}4.01} & 56.61\tiny{{$\pm$}2.45} & 50.11\tiny{{$\pm$}0.11} & 53.19\tiny{{$\pm$}0.98} & 53.29\tiny{{$\pm$}0.14} \\
 &  & M & 72.47\tiny{{$\pm$}0.48} & 73.23\tiny{{$\pm$}0.51} & 71.25\tiny{{$\pm$}0.67} & 63.64\tiny{{$\pm$}0.00} & 62.67\tiny{{$\pm$}0.10} & 60.22\tiny{{$\pm$}1.19} & 58.53\tiny{{$\pm$}1.76} & 57.43\tiny{{$\pm$}1.53} & 61.76\tiny{{$\pm$}2.49} & 61.85\tiny{{$\pm$}1.49} & 47.24\tiny{{$\pm$}0.08} & 51.64\tiny{{$\pm$}0.20} & 51.11\tiny{{$\pm$}0.09} \\
 &  & H & 78.40\tiny{{$\pm$}0.38} & 79.94\tiny{{$\pm$}0.31} & 77.24\tiny{{$\pm$}0.96} & 72.10\tiny{{$\pm$}0.00} & 74.36\tiny{{$\pm$}0.75} & 71.03\tiny{{$\pm$}1.09} & 71.94\tiny{{$\pm$}1.26} & 70.72\tiny{{$\pm$}1.29} & 77.90\tiny{{$\pm$}1.40} & 71.66\tiny{{$\pm$}2.18} & 45.18\tiny{{$\pm$}0.05} & 48.50\tiny{{$\pm$}1.14} & 48.68\tiny{{$\pm$}0.06} \\
 &  & F & 73.86\tiny{{$\pm$}0.28} & 73.63\tiny{{$\pm$}0.50} & 72.40\tiny{{$\pm$}0.60} & 67.08\tiny{{$\pm$}0.00} & 64.56\tiny{{$\pm$}0.04} & 58.05\tiny{{$\pm$}0.88} & 55.07\tiny{{$\pm$}1.43} & 61.36\tiny{{$\pm$}1.03} & 63.39\tiny{{$\pm$}1.21} & 62.54\tiny{{$\pm$}0.81} & 46.26\tiny{{$\pm$}0.04} & 49.81\tiny{{$\pm$}0.89} & 49.83\tiny{{$\pm$}0.08} \\
\midrule
\multirow{4}{*}{\textbf{4}} & \multirow{4}{*}{\textbf{PGD}} & E & 71.22\tiny{{$\pm$}0.75} & 69.19\tiny{{$\pm$}0.66} & 69.06\tiny{{$\pm$}0.79} & 65.52\tiny{{$\pm$}0.00} & 56.40\tiny{{$\pm$}0.10} & 53.39\tiny{{$\pm$}1.94} & 47.77\tiny{{$\pm$}1.29} & 54.70\tiny{{$\pm$}1.99} & 51.16\tiny{{$\pm$}2.93} & 58.02\tiny{{$\pm$}2.31} & 52.72\tiny{{$\pm$}0.03} & 54.71\tiny{{$\pm$}0.33} & 54.71\tiny{{$\pm$}0.07} \\
 &  & M & 72.60\tiny{{$\pm$}0.67} & 72.91\tiny{{$\pm$}0.61} & 70.91\tiny{{$\pm$}0.64} & 63.64\tiny{{$\pm$}0.00} & 62.70\tiny{{$\pm$}0.00} & 60.34\tiny{{$\pm$}1.00} & 57.77\tiny{{$\pm$}1.58} & 58.78\tiny{{$\pm$}1.70} & 62.10\tiny{{$\pm$}1.96} & 60.75\tiny{{$\pm$}3.33} & 48.71\tiny{{$\pm$}0.16} & 51.81\tiny{{$\pm$}0.67} & 51.93\tiny{{$\pm$}0.09} \\
 &  & H & 78.18\tiny{{$\pm$}0.40} & 79.94\tiny{{$\pm$}0.34} & 77.53\tiny{{$\pm$}0.93} & 72.10\tiny{{$\pm$}0.00} & 74.61\tiny{{$\pm$}0.00} & 70.69\tiny{{$\pm$}1.56} & 71.79\tiny{{$\pm$}1.67} & 71.57\tiny{{$\pm$}1.84} & 78.21\tiny{{$\pm$}0.97} & 71.41\tiny{{$\pm$}1.80} & 43.58\tiny{{$\pm$}0.08} & 48.20\tiny{{$\pm$}1.74} & 48.84\tiny{{$\pm$}0.08} \\
 &  & F & 73.84\tiny{{$\pm$}0.26} & 73.58\tiny{{$\pm$}0.36} & 72.38\tiny{{$\pm$}0.54} & 67.08\tiny{{$\pm$}0.00} & 64.46\tiny{{$\pm$}0.16} & 58.31\tiny{{$\pm$}0.62} & 54.90\tiny{{$\pm$}1.58} & 61.60\tiny{{$\pm$}0.95} & 64.25\tiny{{$\pm$}1.48} & 63.21\tiny{{$\pm$}1.19} & 48.26\tiny{{$\pm$}0.03} & 51.45\tiny{{$\pm$}0.77} & 51.61\tiny{{$\pm$}0.06} \\
\midrule
\multirow{4}{*}{\textbf{5}} & \multirow{4}{*}{\textbf{RND}} & E & 71.07\tiny{{$\pm$}0.47} & 67.56\tiny{{$\pm$}1.04} & 68.34\tiny{{$\pm$}0.54} & 65.52\tiny{{$\pm$}0.00} & 56.34\tiny{{$\pm$}0.28} & 54.64\tiny{{$\pm$}2.04} & 51.94\tiny{{$\pm$}1.64} & 60.88\tiny{{$\pm$}1.22} & 69.06\tiny{{$\pm$}1.48} & 60.66\tiny{{$\pm$}1.42} & 52.31\tiny{{$\pm$}0.06} & 53.65\tiny{{$\pm$}0.24} & 53.65\tiny{{$\pm$}0.14} \\
 &  & M & 72.48\tiny{{$\pm$}0.34} & 71.82\tiny{{$\pm$}0.41} & 71.13\tiny{{$\pm$}0.57} & 63.64\tiny{{$\pm$}0.00} & 62.73\tiny{{$\pm$}0.09} & 57.21\tiny{{$\pm$}1.82} & 61.38\tiny{{$\pm$}1.08} & 63.10\tiny{{$\pm$}1.93} & 70.53\tiny{{$\pm$}1.23} & 62.79\tiny{{$\pm$}1.02} & 48.89\tiny{{$\pm$}0.05} & 51.70\tiny{{$\pm$}0.32} & 51.57\tiny{{$\pm$}0.11} \\
 &  & H & 78.37\tiny{{$\pm$}0.44} & 79.84\tiny{{$\pm$}0.42} & 77.93\tiny{{$\pm$}0.72} & 72.10\tiny{{$\pm$}0.00} & 74.58\tiny{{$\pm$}0.22} & 68.18\tiny{{$\pm$}1.85} & 75.24\tiny{{$\pm$}1.81} & 72.41\tiny{{$\pm$}1.34} & 78.37\tiny{{$\pm$}0.87} & 72.41\tiny{{$\pm$}0.40} & 44.74\tiny{{$\pm$}0.07} & 49.19\tiny{{$\pm$}0.26} & 48.77\tiny{{$\pm$}0.18} \\
 &  & F & 74.00\tiny{{$\pm$}0.45} & 72.95\tiny{{$\pm$}0.68} & 72.70\tiny{{$\pm$}0.47} & 67.08\tiny{{$\pm$}0.00} & 64.55\tiny{{$\pm$}0.16} & 55.79\tiny{{$\pm$}0.95} & 60.43\tiny{{$\pm$}0.71} & 65.66\tiny{{$\pm$}0.43} & 73.06\tiny{{$\pm$}0.58} & 65.24\tiny{{$\pm$}0.51} & 48.32\tiny{{$\pm$}0.04} & 50.98\tiny{{$\pm$}0.17} & 50.74\tiny{{$\pm$}0.07} \\
\midrule
\multirow{4}{*}{\textbf{6}} & \multirow{4}{*}{\textbf{W/O Attack}} & E & 70.22\tiny{{$\pm$}0.00} & 71.16\tiny{{$\pm$}0.00} & 70.22\tiny{{$\pm$}0.00} & 65.52\tiny{{$\pm$}0.00} & 56.43\tiny{{$\pm$}0.00} & 71.16\tiny{{$\pm$}0.00} & 68.75\tiny{{$\pm$}0.15} & 63.95\tiny{{$\pm$}0.00} & 69.59\tiny{{$\pm$}0.00} & 64.26\tiny{{$\pm$}0.00} & 52.04\tiny{{$\pm$}0.01} & 53.70\tiny{{$\pm$}0.88} & 54.15\tiny{{$\pm$}0.08} \\
 &  & M & 73.04\tiny{{$\pm$}0.00} & 73.04\tiny{{$\pm$}0.00} & 71.79\tiny{{$\pm$}0.00} & 63.64\tiny{{$\pm$}0.00} & 62.70\tiny{{$\pm$}0.00} & 75.55\tiny{{$\pm$}0.00} & 70.16\tiny{{$\pm$}0.19} & 64.58\tiny{{$\pm$}0.00} & 73.67\tiny{{$\pm$}0.00} & 65.20\tiny{{$\pm$}0.00} & 49.67\tiny{{$\pm$}0.01} & 52.67\tiny{{$\pm$}0.50} & 52.65\tiny{{$\pm$}0.09} \\
 &  & H & 78.37\tiny{{$\pm$}0.00} & 80.88\tiny{{$\pm$}0.00} & 78.06\tiny{{$\pm$}0.00} & 72.10\tiny{{$\pm$}0.00} & 74.61\tiny{{$\pm$}0.00} & 79.62\tiny{{$\pm$}0.00} & 80.41\tiny{{$\pm$}0.16} & 69.59\tiny{{$\pm$}0.00} & 77.43\tiny{{$\pm$}0.00} & 73.98\tiny{{$\pm$}0.00} & 45.94\tiny{{$\pm$}0.01} & 50.01\tiny{{$\pm$}0.56} & 49.99\tiny{{$\pm$}0.00} \\
 &  & F & 73.88\tiny{{$\pm$}0.00} & 75.03\tiny{{$\pm$}0.00} & 73.35\tiny{{$\pm$}0.00} & 67.08\tiny{{$\pm$}0.00} & 64.58\tiny{{$\pm$}0.00} & 75.44\tiny{{$\pm$}0.00} & 73.16\tiny{{$\pm$}0.09} & 66.04\tiny{{$\pm$}0.00} & 73.56\tiny{{$\pm$}0.00} & 67.82\tiny{{$\pm$}0.00} & 49.21\tiny{{$\pm$}0.01} & 51.82\tiny{{$\pm$}0.33} & 51.77\tiny{{$\pm$}0.04} \\
\midrule
\multicolumn{2}{c}{\multirow{4}{*}{\begin{tabular}[c]{@{}c@{}}\textbf{Avg.}\\\textbf{Accuracy}\end{tabular}}} & E & {\color{blue}70.97\tiny{{$\pm$}0.26}} & 68.51\tiny{{$\pm$}0.41} & 68.96\tiny{{$\pm$}0.47} & 65.52\tiny{{$\pm$}0.00} & 56.41\tiny{{$\pm$}0.06} & 56.42\tiny{{$\pm$}0.53} & 53.24\tiny{{$\pm$}0.73} & 55.73\tiny{{$\pm$}0.42} & 55.48\tiny{{$\pm$}1.32} & 56.47\tiny{{$\pm$}2.73} & - & - & - \\
\multicolumn{2}{c}{} & M & {\color{blue}72.55\tiny{{$\pm$}0.21}} & 72.29\tiny{{$\pm$}0.26} & 71.17\tiny{{$\pm$}0.34} & 63.64\tiny{{$\pm$}0.00} & 62.70\tiny{{$\pm$}0.02} & 61.55\tiny{{$\pm$}0.65} & 61.74\tiny{{$\pm$}0.93} & 57.15\tiny{{$\pm$}1.00} & 61.64\tiny{{$\pm$}1.06} & 58.27\tiny{{$\pm$}1.47} & - & - & - \\
\multicolumn{2}{c}{} & H & 78.30\tiny{{$\pm$}0.17} & {\color{blue}80.00\tiny{{$\pm$}0.13}} & 77.58\tiny{{$\pm$}0.29} & 72.10\tiny{{$\pm$}0.00} & 74.57\tiny{{$\pm$}0.13} & 71.15\tiny{{$\pm$}0.50} & 75.06\tiny{{$\pm$}0.71} & 66.49\tiny{{$\pm$}0.67} & 74.42\tiny{{$\pm$}1.31} & 67.78\tiny{{$\pm$}1.47} & - & - & - \\
\multicolumn{2}{c}{} & F & {\color{blue}73.89\tiny{{$\pm$}0.11}} & 73.04\tiny{{$\pm$}0.17} & 72.45\tiny{{$\pm$}0.17} & 67.08\tiny{{$\pm$}0.00} & 64.54\tiny{{$\pm$}0.04} & 60.16\tiny{{$\pm$}0.55} & 59.93\tiny{{$\pm$}0.57} & 59.94\tiny{{$\pm$}0.63} & 62.04\tiny{{$\pm$}1.70} & 60.42\tiny{{$\pm$}1.30} & - & - & - \\
\midrule
\multicolumn{2}{c}{\multirow{4}{*}{\begin{tabular}[c]{@{}c@{}}\textbf{Avg. 3-Min} \\\textbf{Accuracy}\end{tabular}}} & E & 70.37\tiny{{$\pm$}0.29} & 67.19\tiny{{$\pm$}0.80} & 68.08\tiny{{$\pm$}0.64} & 65.52\tiny{{$\pm$}0.00} & 56.38\tiny{{$\pm$}0.10} & 51.58\tiny{{$\pm$}0.94} & 48.25\tiny{{$\pm$}0.95} & 51.09\tiny{{$\pm$}0.74} & 47.07\tiny{{$\pm$}1.85} & 51.32\tiny{{$\pm$}4.97} & - & - & - \\
\multicolumn{2}{c}{} & M & {\color{blue}72.17\tiny{{$\pm$}0.26}} & 71.44\tiny{{$\pm$}0.37} & 70.50\tiny{{$\pm$}0.56} & 63.64\tiny{{$\pm$}0.00} & 62.69\tiny{{$\pm$}0.03} & 57.22\tiny{{$\pm$}1.16} & 58.43\tiny{{$\pm$}1.16} & 51.98\tiny{{$\pm$}1.52} & 54.06\tiny{{$\pm$}2.00} & 52.96\tiny{{$\pm$}3.03} & - & - & - \\
\multicolumn{2}{c}{} & H & 78.05\tiny{{$\pm$}0.20} & {\color{blue}79.59\tiny{{$\pm$}0.21}} & 77.00\tiny{{$\pm$}0.40} & 72.10\tiny{{$\pm$}0.00} & 74.51\tiny{{$\pm$}0.25} & 68.16\tiny{{$\pm$}0.90} & 72.31\tiny{{$\pm$}0.96} & 61.27\tiny{{$\pm$}1.55} & 70.49\tiny{{$\pm$}2.41} & 62.42\tiny{{$\pm$}2.81} & - & - & - \\
\multicolumn{2}{c}{} & F & {\color{blue}73.68\tiny{{$\pm$}0.12}} & 71.94\tiny{{$\pm$}0.36} & 71.86\tiny{{$\pm$}0.30} & 67.08\tiny{{$\pm$}0.00} & 64.49\tiny{{$\pm$}0.08} & 55.61\tiny{{$\pm$}0.58} & 55.19\tiny{{$\pm$}0.62} & 55.26\tiny{{$\pm$}1.07} & 53.69\tiny{{$\pm$}3.32} & 55.31\tiny{{$\pm$}2.84} & - & - & - \\
\midrule
\multicolumn{2}{c}{\multirow{4}{*}{\begin{tabular}[c]{@{}c@{}}\textbf{Weighted} \\\textbf{Accuracy}\end{tabular}}} & E & {\color{blue}70.26\tiny{{$\pm$}0.29}} & 66.91\tiny{{$\pm$}0.80} & 67.73\tiny{{$\pm$}0.82} & 65.52\tiny{{$\pm$}0.00} & 56.33\tiny{{$\pm$}0.19} & 50.83\tiny{{$\pm$}1.23} & 47.83\tiny{{$\pm$}1.09} & 49.98\tiny{{$\pm$}0.93} & 46.26\tiny{{$\pm$}2.89} & 47.56\tiny{{$\pm$}6.43} & - & - & - \\
\multicolumn{2}{c}{} & M & {\color{blue}72.04\tiny{{$\pm$}0.27}} & 71.03\tiny{{$\pm$}0.58} & 70.09\tiny{{$\pm$}0.74} & 63.64\tiny{{$\pm$}0.00} & 62.68\tiny{{$\pm$}0.06} & 56.51\tiny{{$\pm$}1.39} & 57.80\tiny{{$\pm$}1.03} & 50.29\tiny{{$\pm$}2.25} & 51.08\tiny{{$\pm$}3.58} & 48.58\tiny{{$\pm$}5.29} & - & - & - \\
\multicolumn{2}{c}{} & H & 77.94\tiny{{$\pm$}0.22} & {\color{blue}79.53\tiny{{$\pm$}0.22}} & 76.79\tiny{{$\pm$}0.41} & 72.10\tiny{{$\pm$}0.00} & 74.40\tiny{{$\pm$}0.51} & 67.47\tiny{{$\pm$}1.09} & 71.97\tiny{{$\pm$}1.10} & 58.78\tiny{{$\pm$}1.41} & 67.53\tiny{{$\pm$}2.85} & 56.76\tiny{{$\pm$}4.16} & - & - & - \\
\multicolumn{2}{c}{} & F & {\color{blue}73.64\tiny{{$\pm$}0.15}} & 71.59\tiny{{$\pm$}0.34} & 71.48\tiny{{$\pm$}0.66} & 67.08\tiny{{$\pm$}0.00} & 64.44\tiny{{$\pm$}0.12} & 55.05\tiny{{$\pm$}1.00} & 54.95\tiny{{$\pm$}0.76} & 53.66\tiny{{$\pm$}1.23} & 49.94\tiny{{$\pm$}3.72} & 49.84\tiny{{$\pm$}6.03} & - & - & - \\
\bottomrule
\end{tabular}}
\label{tab:leaderboard_citeseer_top10}
\end{table}

\begin{table}[!htp]
\caption{\textit{grb-flickr} leaderboard (Top 5 ATK.\ vs.\ Top 10 DEF.) in \textit{graph injection} scenario.}
\centering
\scalebox{0.46}{
\begin{tabular}{cccccccccccccccc} 
\toprule
\multicolumn{3}{c}{\multirow{2}{*}{\diagbox{\textbf{Attack}}{\textbf{Defenses}}}} & \textbf{1} & \textbf{2} & \textbf{3} & \textbf{4} & \textbf{5} & \textbf{6} & \textbf{7} & \textbf{8} & \textbf{9} & \textbf{10} & \multirow{2}{*}{\begin{tabular}[c]{@{}c@{}}\textbf{Avg.}\\\textbf{Accuracy~}\end{tabular}} & \multirow{2}{*}{\begin{tabular}[c]{@{}c@{}}\textbf{Avg. 3-Max}\\\textbf{Accuracy}\end{tabular}} & \multirow{2}{*}{\begin{tabular}[c]{@{}c@{}}\textbf{Weighted}\\\textbf{Accuracy}\end{tabular}} \\
\multicolumn{3}{c}{} & \multicolumn{1}{c}{\textbf{R-GCN\tiny{+AT}}} & \multicolumn{1}{c}{\textbf{GAT\tiny{+LN}}} & \multicolumn{1}{c}{\textbf{SAGE\tiny{+LN}}} & \multicolumn{1}{c}{\textbf{GIN\tiny{LN}}} & \multicolumn{1}{c}{\textbf{GCN\tiny{+AT}}} & \multicolumn{1}{c}{\textbf{SAGE\tiny{+AT}}} & \multicolumn{1}{c}{\textbf{GAT}} & \multicolumn{1}{c}{\textbf{GIN\tiny{+AT}}} & \multicolumn{1}{c}{\textbf{SAGE}} & \multicolumn{1}{c}{\textbf{APPNP\tiny{+LN}}} & \multicolumn{1}{c}{} & \multicolumn{1}{c}{} & \multicolumn{1}{c}{} \\
\midrule
\multirow{4}{*}{\textbf{1}} & \multirow{4}{*}{\textbf{SPEIT}} & E & 53.41\tiny{{$\pm$}0.12} & 53.36\tiny{{$\pm$}0.22} & 51.43\tiny{{$\pm$}0.15} & 50.96\tiny{{$\pm$}0.29} & 52.89\tiny{{$\pm$}0.22} & 52.87\tiny{{$\pm$}0.13} & 53.62\tiny{{$\pm$}0.36} & 50.15\tiny{{$\pm$}0.14} & 51.65\tiny{{$\pm$}0.45} & 49.78\tiny{{$\pm$}0.03} & 52.01\tiny{{$\pm$}0.09} & 53.48\tiny{{$\pm$}0.25} & 53.44\tiny{{$\pm$}0.17} \\
 &  & M & 49.79\tiny{{$\pm$}0.26} & 49.21\tiny{{$\pm$}0.26} & 47.83\tiny{{$\pm$}0.09} & 46.19\tiny{{$\pm$}0.14} & 48.05\tiny{{$\pm$}0.24} & 48.91\tiny{{$\pm$}0.12} & 52.04\tiny{{$\pm$}0.33} & 43.62\tiny{{$\pm$}0.10} & 44.54\tiny{{$\pm$}0.12} & 42.94\tiny{{$\pm$}0.03} & 47.31\tiny{{$\pm$}0.09} & {\color{red}50.35\tiny{{$\pm$}1.25}} & {\color{red}50.89\tiny{{$\pm$}0.24}} \\
 &  & H & 44.72\tiny{{$\pm$}0.15} & 46.78\tiny{{$\pm$}0.54} & 43.41\tiny{{$\pm$}0.12} & 44.79\tiny{{$\pm$}0.09} & 44.83\tiny{{$\pm$}0.14} & 43.51\tiny{{$\pm$}0.15} & 50.32\tiny{{$\pm$}0.20} & 37.33\tiny{{$\pm$}0.12} & 37.17\tiny{{$\pm$}0.17} & 34.33\tiny{{$\pm$}0.02} & {\color{red}42.72\tiny{{$\pm$}0.07}} & {\color{red}47.34\tiny{{$\pm$}2.27}} & {\color{red}48.44\tiny{{$\pm$}0.12}} \\
 &  & F & 47.14\tiny{{$\pm$}0.13} & 46.39\tiny{{$\pm$}0.13} & 47.62\tiny{{$\pm$}0.04} & 46.18\tiny{{$\pm$}0.09} & 43.52\tiny{{$\pm$}0.07} & 48.17\tiny{{$\pm$}0.14} & 49.91\tiny{{$\pm$}0.17} & 43.73\tiny{{$\pm$}0.05} & 43.66\tiny{{$\pm$}0.07} & 42.30\tiny{{$\pm$}0.01} & {\color{red}45.86\tiny{{$\pm$}0.05}} & {\color{red}48.56\tiny{{$\pm$}0.98}} & {\color{red}48.95\tiny{{$\pm$}0.12}} \\
\midrule
\multirow{4}{*}{\textbf{2}} & \multirow{4}{*}{\textbf{FGSM}} & E & 53.99\tiny{{$\pm$}0.13} & 51.87\tiny{{$\pm$}0.49} & 50.89\tiny{{$\pm$}0.15} & 48.24\tiny{{$\pm$}0.15} & 53.79\tiny{{$\pm$}0.17} & 44.06\tiny{{$\pm$}0.23} & 48.23\tiny{{$\pm$}0.56} & 50.08\tiny{{$\pm$}0.16} & 49.99\tiny{{$\pm$}0.33} & 50.07\tiny{{$\pm$}0.05} & {\color{red}50.12\tiny{{$\pm$}0.14}} & {\color{red}53.22\tiny{{$\pm$}1.01}} & {\color{red}53.30\tiny{{$\pm$}0.12}} \\
 &  & M & 51.79\tiny{{$\pm$}0.13} & 51.42\tiny{{$\pm$}0.20} & 47.89\tiny{{$\pm$}0.13} & 46.43\tiny{{$\pm$}0.14} & 51.54\tiny{{$\pm$}0.13} & 44.24\tiny{{$\pm$}0.17} & 46.17\tiny{{$\pm$}0.41} & 44.81\tiny{{$\pm$}0.14} & 44.24\tiny{{$\pm$}0.22} & 43.59\tiny{{$\pm$}0.06} & {\color{red}47.21\tiny{{$\pm$}0.06}} & 51.58\tiny{{$\pm$}0.22} & 51.06\tiny{{$\pm$}0.10} \\
 &  & H & 46.87\tiny{{$\pm$}0.17} & 49.32\tiny{{$\pm$}0.09} & 43.69\tiny{{$\pm$}0.11} & 46.76\tiny{{$\pm$}0.16} & 49.35\tiny{{$\pm$}0.08} & 46.46\tiny{{$\pm$}0.22} & 45.24\tiny{{$\pm$}0.29} & 42.58\tiny{{$\pm$}0.30} & 45.90\tiny{{$\pm$}0.16} & 35.56\tiny{{$\pm$}0.08} & 45.17\tiny{{$\pm$}0.07} & 48.53\tiny{{$\pm$}1.14} & 48.68\tiny{{$\pm$}0.06} \\
 &  & F & 50.20\tiny{{$\pm$}0.08} & 50.47\tiny{{$\pm$}0.15} & 47.58\tiny{{$\pm$}0.08} & 46.94\tiny{{$\pm$}0.07} & 48.58\tiny{{$\pm$}0.09} & 43.10\tiny{{$\pm$}0.14} & 42.78\tiny{{$\pm$}0.27} & 45.28\tiny{{$\pm$}0.10} & 43.33\tiny{{$\pm$}0.19} & 44.16\tiny{{$\pm$}0.09} & 46.24\tiny{{$\pm$}0.04} & 49.75\tiny{{$\pm$}0.84} & 49.73\tiny{{$\pm$}0.09} \\
\midrule
\multirow{4}{*}{\textbf{3}} & \multirow{4}{*}{\textbf{PGD}} & E & 54.02\tiny{{$\pm$}0.18} & 51.88\tiny{{$\pm$}0.40} & 50.77\tiny{{$\pm$}0.19} & 48.36\tiny{{$\pm$}0.19} & 53.68\tiny{{$\pm$}0.18} & 43.95\tiny{{$\pm$}0.24} & 48.33\tiny{{$\pm$}0.71} & 50.04\tiny{{$\pm$}0.15} & 49.95\tiny{{$\pm$}0.42} & 50.05\tiny{{$\pm$}0.07} & 50.11\tiny{{$\pm$}0.11} & 53.19\tiny{{$\pm$}0.98} & 53.29\tiny{{$\pm$}0.14} \\
 &  & M & 51.74\tiny{{$\pm$}0.22} & 51.65\tiny{{$\pm$}0.21} & 47.77\tiny{{$\pm$}0.16} & 46.69\tiny{{$\pm$}0.23} & 51.53\tiny{{$\pm$}0.10} & 44.26\tiny{{$\pm$}0.17} & 45.91\tiny{{$\pm$}0.40} & 44.95\tiny{{$\pm$}0.27} & 44.38\tiny{{$\pm$}0.43} & 43.56\tiny{{$\pm$}0.10} & 47.24\tiny{{$\pm$}0.08} & 51.64\tiny{{$\pm$}0.20} & 51.11\tiny{{$\pm$}0.09} \\
 &  & H & 46.81\tiny{{$\pm$}0.13} & 49.28\tiny{{$\pm$}0.13} & 43.75\tiny{{$\pm$}0.17} & 46.75\tiny{{$\pm$}0.22} & 49.32\tiny{{$\pm$}0.14} & 46.45\tiny{{$\pm$}0.18} & 45.24\tiny{{$\pm$}0.26} & 42.47\tiny{{$\pm$}0.27} & 46.17\tiny{{$\pm$}0.19} & 35.55\tiny{{$\pm$}0.08} & 45.18\tiny{{$\pm$}0.05} & 48.50\tiny{{$\pm$}1.14} & 48.68\tiny{{$\pm$}0.06} \\
 &  & F & 50.22\tiny{{$\pm$}0.11} & 50.62\tiny{{$\pm$}0.12} & 47.56\tiny{{$\pm$}0.07} & 46.94\tiny{{$\pm$}0.10} & 48.58\tiny{{$\pm$}0.09} & 43.09\tiny{{$\pm$}0.15} & 42.90\tiny{{$\pm$}0.24} & 45.24\tiny{{$\pm$}0.10} & 43.32\tiny{{$\pm$}0.14} & 44.14\tiny{{$\pm$}0.07} & 46.26\tiny{{$\pm$}0.04} & 49.81\tiny{{$\pm$}0.89} & 49.83\tiny{{$\pm$}0.08} \\
\midrule
\multirow{4}{*}{\textbf{4}} & \multirow{4}{*}{\textbf{TDGIA}} & E & 55.13\tiny{{$\pm$}0.08} & 54.17\tiny{{$\pm$}0.17} & 51.83\tiny{{$\pm$}0.08} & 52.29\tiny{{$\pm$}0.06} & 54.35\tiny{{$\pm$}0.07} & 53.61\tiny{{$\pm$}0.05} & 54.66\tiny{{$\pm$}0.12} & 49.95\tiny{{$\pm$}0.02} & 51.50\tiny{{$\pm$}0.12} & 49.72\tiny{{$\pm$}0.01} & 52.72\tiny{{$\pm$}0.03} & 54.71\tiny{{$\pm$}0.33} & 54.71\tiny{{$\pm$}0.07} \\
 &  & M & 51.45\tiny{{$\pm$}0.12} & 51.30\tiny{{$\pm$}0.44} & 46.86\tiny{{$\pm$}0.08} & 48.54\tiny{{$\pm$}0.17} & 50.86\tiny{{$\pm$}0.31} & 50.03\tiny{{$\pm$}0.16} & 52.67\tiny{{$\pm$}0.24} & 43.20\tiny{{$\pm$}0.07} & 49.31\tiny{{$\pm$}1.04} & 42.88\tiny{{$\pm$}0.00} & 48.71\tiny{{$\pm$}0.16} & 51.81\tiny{{$\pm$}0.67} & 51.93\tiny{{$\pm$}0.09} \\
 &  & H & 45.26\tiny{{$\pm$}0.10} & 48.19\tiny{{$\pm$}0.17} & 42.04\tiny{{$\pm$}0.09} & 46.09\tiny{{$\pm$}0.07} & 44.97\tiny{{$\pm$}0.30} & 43.16\tiny{{$\pm$}0.09} & 50.33\tiny{{$\pm$}0.13} & 37.33\tiny{{$\pm$}0.05} & 44.04\tiny{{$\pm$}0.08} & 34.41\tiny{{$\pm$}0.01} & 43.58\tiny{{$\pm$}0.08} & 48.20\tiny{{$\pm$}1.74} & 48.84\tiny{{$\pm$}0.08} \\
 &  & F & 48.97\tiny{{$\pm$}0.05} & 51.25\tiny{{$\pm$}0.07} & 47.13\tiny{{$\pm$}0.07} & 49.05\tiny{{$\pm$}0.05} & 47.54\tiny{{$\pm$}0.06} & 49.52\tiny{{$\pm$}0.05} & 52.46\tiny{{$\pm$}0.09} & 43.77\tiny{{$\pm$}0.01} & 50.63\tiny{{$\pm$}0.08} & 42.30\tiny{{$\pm$}0.00} & 48.26\tiny{{$\pm$}0.03} & 51.45\tiny{{$\pm$}0.77} & 51.61\tiny{{$\pm$}0.06} \\
\multirow{4}{*}{\textbf{5}} & \multirow{4}{*}{\textbf{RND}} & E & 53.90\tiny{{$\pm$}0.20} & 53.31\tiny{{$\pm$}0.18} & 51.73\tiny{{$\pm$}0.14} & 51.16\tiny{{$\pm$}0.15} & 53.62\tiny{{$\pm$}0.10} & 53.19\tiny{{$\pm$}0.12} & 52.77\tiny{{$\pm$}0.22} & 50.36\tiny{{$\pm$}0.08} & 53.25\tiny{{$\pm$}0.23} & 49.83\tiny{{$\pm$}0.03} & 52.31\tiny{{$\pm$}0.06} & 53.65\tiny{{$\pm$}0.24} & 53.65\tiny{{$\pm$}0.14} \\
 &  & M & 51.22\tiny{{$\pm$}0.15} & 52.01\tiny{{$\pm$}0.15} & 47.82\tiny{{$\pm$}0.12} & 48.21\tiny{{$\pm$}0.15} & 51.78\tiny{{$\pm$}0.14} & 49.84\tiny{{$\pm$}0.19} & 51.06\tiny{{$\pm$}0.30} & 44.11\tiny{{$\pm$}0.03} & 49.85\tiny{{$\pm$}0.18} & 43.04\tiny{{$\pm$}0.06} & 48.89\tiny{{$\pm$}0.05} & 51.70\tiny{{$\pm$}0.32} & 51.57\tiny{{$\pm$}0.11} \\
 &  & H & 46.38\tiny{{$\pm$}0.15} & 49.33\tiny{{$\pm$}0.18} & 42.90\tiny{{$\pm$}0.13} & 46.79\tiny{{$\pm$}0.14} & 48.95\tiny{{$\pm$}0.12} & 45.03\tiny{{$\pm$}0.15} & 49.29\tiny{{$\pm$}0.25} & 39.10\tiny{{$\pm$}0.09} & 45.13\tiny{{$\pm$}0.28} & 34.50\tiny{{$\pm$}0.03} & 44.74\tiny{{$\pm$}0.07} & 49.19\tiny{{$\pm$}0.26} & 48.77\tiny{{$\pm$}0.18} \\
 &  & F & 49.32\tiny{{$\pm$}0.10} & 51.09\tiny{{$\pm$}0.08} & 47.47\tiny{{$\pm$}0.07} & 48.58\tiny{{$\pm$}0.06} & 51.00\tiny{{$\pm$}0.09} & 49.18\tiny{{$\pm$}0.07} & 50.84\tiny{{$\pm$}0.20} & 44.36\tiny{{$\pm$}0.03} & 49.00\tiny{{$\pm$}0.08} & 42.31\tiny{{$\pm$}0.01} & 48.32\tiny{{$\pm$}0.04} & 50.98\tiny{{$\pm$}0.17} & 50.74\tiny{{$\pm$}0.07} \\
\midrule
\multirow{4}{*}{\textbf{6}} & \multirow{4}{*}{\textbf{W/O Attack}} & E & 54.94\tiny{{$\pm$}0.12} & 52.58\tiny{{$\pm$}0.00} & 51.68\tiny{{$\pm$}0.00} & 50.48\tiny{{$\pm$}0.00} & 52.97\tiny{{$\pm$}0.00} & 53.18\tiny{{$\pm$}0.00} & 49.50\tiny{{$\pm$}0.00} & 50.72\tiny{{$\pm$}0.00} & 52.77\tiny{{$\pm$}0.00} & 51.55\tiny{{$\pm$}0.00} & 52.04\tiny{{$\pm$}0.01} & 53.70\tiny{{$\pm$}0.88} & 54.15\tiny{{$\pm$}0.08} \\
 &  & M & 53.23\tiny{{$\pm$}0.13} & 52.75\tiny{{$\pm$}0.00} & 47.83\tiny{{$\pm$}0.00} & 48.03\tiny{{$\pm$}0.00} & 52.03\tiny{{$\pm$}0.00} & 51.27\tiny{{$\pm$}0.00} & 50.06\tiny{{$\pm$}0.00} & 44.87\tiny{{$\pm$}0.00} & 51.63\tiny{{$\pm$}0.00} & 45.01\tiny{{$\pm$}0.00} & 49.67\tiny{{$\pm$}0.01} & 52.67\tiny{{$\pm$}0.50} & 52.65\tiny{{$\pm$}0.09} \\
 &  & H & 48.70\tiny{{$\pm$}0.08} & 49.65\tiny{{$\pm$}0.00} & 42.90\tiny{{$\pm$}0.00} & 46.43\tiny{{$\pm$}0.00} & 49.59\tiny{{$\pm$}0.00} & 47.10\tiny{{$\pm$}0.00} & 50.80\tiny{{$\pm$}0.00} & 40.10\tiny{{$\pm$}0.00} & 47.96\tiny{{$\pm$}0.00} & 36.12\tiny{{$\pm$}0.00} & 45.94\tiny{{$\pm$}0.01} & 50.01\tiny{{$\pm$}0.56} & 49.99\tiny{{$\pm$}0.00} \\
 &  & F & 52.28\tiny{{$\pm$}0.06} & 51.66\tiny{{$\pm$}0.00} & 47.47\tiny{{$\pm$}0.00} & 48.31\tiny{{$\pm$}0.00} & 51.53\tiny{{$\pm$}0.00} & 50.52\tiny{{$\pm$}0.00} & 50.12\tiny{{$\pm$}0.00} & 45.23\tiny{{$\pm$}0.00} & 50.79\tiny{{$\pm$}0.00} & 44.23\tiny{{$\pm$}0.00} & 49.21\tiny{{$\pm$}0.01} & 51.82\tiny{{$\pm$}0.33} & 51.77\tiny{{$\pm$}0.04} \\
\midrule
\multicolumn{2}{c}{\multirow{4}{*}{\begin{tabular}[c]{@{}c@{}}\textbf{Avg.}\\\textbf{Accuracy}\end{tabular}}} & E & 54.23\tiny{{$\pm$}0.07} & 52.86\tiny{{$\pm$}0.16} & 51.38\tiny{{$\pm$}0.05} & 50.25\tiny{{$\pm$}0.09} & 53.55\tiny{{$\pm$}0.07} & 50.14\tiny{{$\pm$}0.07} & 51.19\tiny{{$\pm$}0.15} & 50.22\tiny{{$\pm$}0.04} & 51.52\tiny{{$\pm$}0.12} & 50.17\tiny{{$\pm$}0.02} & \multicolumn{1}{c}{-} & \multicolumn{1}{c}{-} & \multicolumn{1}{c}{-} \\
\multicolumn{2}{c}{} & M & 51.54\tiny{{$\pm$}0.07} & 51.39\tiny{{$\pm$}0.10} & 47.67\tiny{{$\pm$}0.05} & 47.35\tiny{{$\pm$}0.06} & 50.96\tiny{{$\pm$}0.04} & 48.09\tiny{{$\pm$}0.07} & 49.65\tiny{{$\pm$}0.09} & 44.26\tiny{{$\pm$}0.06} & 47.32\tiny{{$\pm$}0.19} & 43.50\tiny{{$\pm$}0.02} & \multicolumn{1}{c}{-} & \multicolumn{1}{c}{-} & \multicolumn{1}{c}{-} \\
\multicolumn{2}{c}{} & H & 46.46\tiny{{$\pm$}0.05} & 48.76\tiny{{$\pm$}0.11} & 43.12\tiny{{$\pm$}0.05} & 46.27\tiny{{$\pm$}0.04} & 47.84\tiny{{$\pm$}0.08} & 45.29\tiny{{$\pm$}0.04} & 48.54\tiny{{$\pm$}0.08} & 39.82\tiny{{$\pm$}0.06} & 44.40\tiny{{$\pm$}0.10} & 35.08\tiny{{$\pm$}0.02} & \multicolumn{1}{c}{-} & \multicolumn{1}{c}{-} & \multicolumn{1}{c}{-} \\
\multicolumn{2}{c}{} & F & 49.69\tiny{{$\pm$}0.04} & 50.25\tiny{{$\pm$}0.03} & 47.47\tiny{{$\pm$}0.03} & 47.67\tiny{{$\pm$}0.03} & 48.46\tiny{{$\pm$}0.02} & 47.26\tiny{{$\pm$}0.04} & 48.17\tiny{{$\pm$}0.09} & 44.60\tiny{{$\pm$}0.03} & 46.79\tiny{{$\pm$}0.06} & 43.24\tiny{{$\pm$}0.01} & \multicolumn{1}{c}{-} & \multicolumn{1}{c}{-} & \multicolumn{1}{c}{-} \\
\midrule
\multicolumn{2}{c}{\multirow{4}{*}{\begin{tabular}[c]{@{}c@{}}\textbf{Avg. 3-Min} \\\textbf{Accuracy}\end{tabular}}} & E & {\color{blue}53.74\tiny{{$\pm$}0.09}} & 52.11\tiny{{$\pm$}0.26} & 51.03\tiny{{$\pm$}0.10} & 49.03\tiny{{$\pm$}0.07} & 53.14\tiny{{$\pm$}0.09} & 46.96\tiny{{$\pm$}0.14} & 48.69\tiny{{$\pm$}0.25} & 50.00\tiny{{$\pm$}0.07} & 50.46\tiny{{$\pm$}0.22} & 49.78\tiny{{$\pm$}0.02} & \multicolumn{1}{c}{-} & \multicolumn{1}{c}{-} & \multicolumn{1}{c}{-} \\
\multicolumn{2}{c}{} & M & {\color{blue}50.81\tiny{{$\pm$}0.10}} & 50.62\tiny{{$\pm$}0.17} & 47.44\tiny{{$\pm$}0.07} & 46.44\tiny{{$\pm$}0.11} & 50.13\tiny{{$\pm$}0.09} & 45.81\tiny{{$\pm$}0.09} & 47.38\tiny{{$\pm$}0.19} & 43.65\tiny{{$\pm$}0.04} & 44.38\tiny{{$\pm$}0.21} & 42.95\tiny{{$\pm$}0.02} & \multicolumn{1}{c}{-} & \multicolumn{1}{c}{-} & \multicolumn{1}{c}{-} \\
\multicolumn{2}{c}{} & H & 45.45\tiny{{$\pm$}0.07} & {\color{blue}48.05\tiny{{$\pm$}0.21}} & 42.62\tiny{{$\pm$}0.05} & 45.77\tiny{{$\pm$}0.04} & 46.25\tiny{{$\pm$}0.15} & 43.90\tiny{{$\pm$}0.08} & 46.59\tiny{{$\pm$}0.17} & 37.92\tiny{{$\pm$}0.04} & 42.11\tiny{{$\pm$}0.15} & 34.41\tiny{{$\pm$}0.01} & \multicolumn{1}{c}{-} & \multicolumn{1}{c}{-} & \multicolumn{1}{c}{-} \\
\multicolumn{2}{c}{} & F & 48.48\tiny{{$\pm$}0.06} & {\color{blue}49.16\tiny{{$\pm$}0.04}} & 47.35\tiny{{$\pm$}0.04} & 46.69\tiny{{$\pm$}0.04} & 46.52\tiny{{$\pm$}0.04} & 44.79\tiny{{$\pm$}0.07} & 45.19\tiny{{$\pm$}0.14} & 43.95\tiny{{$\pm$}0.02} & 43.44\tiny{{$\pm$}0.10} & 42.30\tiny{{$\pm$}0.00} & \multicolumn{1}{c}{-} & \multicolumn{1}{c}{-} & \multicolumn{1}{c}{-} \\
\midrule
\multicolumn{2}{c}{\multirow{4}{*}{\begin{tabular}[c]{@{}c@{}}\textbf{Weighted} \\\textbf{Accuracy}\end{tabular}}} & E & {\color{blue}53.62\tiny{{$\pm$}0.08}} & 51.97\tiny{{$\pm$}0.33} & 50.92\tiny{{$\pm$}0.12} & 48.67\tiny{{$\pm$}0.09} & 53.02\tiny{{$\pm$}0.12} & 45.43\tiny{{$\pm$}0.18} & 48.62\tiny{{$\pm$}0.42} & 49.97\tiny{{$\pm$}0.06} & 50.22\tiny{{$\pm$}0.26} & 49.80\tiny{{$\pm$}0.01} & \multicolumn{1}{c}{-} & \multicolumn{1}{c}{-} & \multicolumn{1}{c}{-} \\
\multicolumn{2}{c}{} & M & {\color{blue}50.35\tiny{{$\pm$}0.18}} & 49.95\tiny{{$\pm$}0.18} & 47.16\tiny{{$\pm$}0.07} & 46.44\tiny{{$\pm$}0.10} & 49.10\tiny{{$\pm$}0.14} & 45.06\tiny{{$\pm$}0.09} & 46.70\tiny{{$\pm$}0.28} & 43.48\tiny{{$\pm$}0.04} & 44.73\tiny{{$\pm$}0.21} & 42.99\tiny{{$\pm$}0.01} & \multicolumn{1}{c}{-} & \multicolumn{1}{c}{-} & \multicolumn{1}{c}{-} \\
\multicolumn{2}{c}{} & H & 45.15\tiny{{$\pm$}0.10} & {\color{blue}47.43\tiny{{$\pm$}0.37}} & 42.37\tiny{{$\pm$}0.06} & 45.30\tiny{{$\pm$}0.07} & 45.55\tiny{{$\pm$}0.13} & 43.66\tiny{{$\pm$}0.07} & 45.92\tiny{{$\pm$}0.18} & 37.78\tiny{{$\pm$}0.04} & 39.72\tiny{{$\pm$}0.13} & 34.47\tiny{{$\pm$}0.01} & \multicolumn{1}{c}{-} & \multicolumn{1}{c}{-} & \multicolumn{1}{c}{-} \\
\multicolumn{2}{c}{} & F & {\color{blue}47.91\tiny{{$\pm$}0.09}} & 47.81\tiny{{$\pm$}0.08} & 47.25\tiny{{$\pm$}0.05} & 46.57\tiny{{$\pm$}0.06} & 45.13\tiny{{$\pm$}0.05} & 43.99\tiny{{$\pm$}0.09} & 43.98\tiny{{$\pm$}0.18} & 43.91\tiny{{$\pm$}0.03} & 43.88\tiny{{$\pm$}0.10} & 42.46\tiny{{$\pm$}0.01} & \multicolumn{1}{c}{-} & \multicolumn{1}{c}{-} & \multicolumn{1}{c}{-} \\
\bottomrule
\end{tabular}}
\label{tab:leaderboard_flickr_top10}
\end{table}

\begin{table}[!htp]
\caption{\textit{grb-reddit} leaderboard (Top 5 ATK.\ vs.\ Top 10 DEF.) in \textit{graph injection} scenario.}
\centering
\scalebox{0.46}{
\begin{tabular}{cccccccccccccccc} 
\toprule
\multicolumn{3}{c}{\multirow{2}{*}{\diagbox{\textbf{Attack}}{\textbf{Defenses}}}} & \textbf{1} & \textbf{2} & \textbf{3} & \textbf{4} & \textbf{5} & \textbf{6} & \textbf{7} & \textbf{8} & \textbf{9} & \textbf{10} & \multirow{2}{*}{\begin{tabular}[c]{@{}c@{}}\textbf{Avg.}\\\textbf{Accuracy~}\end{tabular}} & \multirow{2}{*}{\begin{tabular}[c]{@{}c@{}}\textbf{Avg. 3-Max}\\\textbf{Accuracy}\end{tabular}} & \multirow{2}{*}{\begin{tabular}[c]{@{}c@{}}\textbf{Weighted}\\\textbf{Accuracy}\end{tabular}} \\
\multicolumn{3}{c}{} & \textbf{GIN\tiny{+LN}} & \textbf{TAGCN\tiny{+LN}} & \textbf{TAGCN\tiny{+AT}} & \textbf{GAT\tiny{+LN}} & \textbf{R-GCN\tiny{+AT}} & \textbf{TAGCN} & \textbf{GCN\tiny{+LN}} & \textbf{SAGE\tiny{+AT}} & \textbf{SAGE} & \textbf{SGCN\tiny{+LN}} &  &  &  \\
\midrule
\multirow{4}{*}{\textbf{1}} & \multirow{4}{*}{\textbf{TDGIA}} & E & 89.57\tiny{{$\pm$}0.05} & 90.59\tiny{{$\pm$}0.02} & 70.17\tiny{{$\pm$}0.28} & 84.18\tiny{{$\pm$}0.03} & 88.38\tiny{{$\pm$}0.02} & 78.83\tiny{{$\pm$}0.18} & 80.79\tiny{{$\pm$}0.17} & 86.29\tiny{{$\pm$}0.03} & 71.00\tiny{{$\pm$}0.11} & 76.36\tiny{{$\pm$}0.03} & {\color{red}81.62\tiny{{$\pm$}0.08}} & {\color{red}89.52\tiny{{$\pm$}0.90}} & {\color{red}89.17\tiny{{$\pm$}0.03}} \\
 &  & M & 98.29\tiny{{$\pm$}0.01} & 97.79\tiny{{$\pm$}0.00} & 98.02\tiny{{$\pm$}0.01} & 96.20\tiny{{$\pm$}0.01} & 95.50\tiny{{$\pm$}0.01} & 96.62\tiny{{$\pm$}0.01} & 97.66\tiny{{$\pm$}0.00} & 94.07\tiny{{$\pm$}0.01} & 95.66\tiny{{$\pm$}0.02} & 92.27\tiny{{$\pm$}0.01} & {\color{red}96.21\tiny{{$\pm$}0.00}} & {\color{red}98.03\tiny{{$\pm$}0.21}} & {\color{red}97.97\tiny{{$\pm$}0.01}} \\
 &  & H & 99.54\tiny{{$\pm$}0.00} & 99.16\tiny{{$\pm$}0.00} & 98.54\tiny{{$\pm$}0.01} & 98.17\tiny{{$\pm$}0.01} & 95.68\tiny{{$\pm$}0.02} & 97.77\tiny{{$\pm$}0.01} & 99.14\tiny{{$\pm$}0.01} & 90.60\tiny{{$\pm$}0.01} & 98.43\tiny{{$\pm$}0.02} & 93.33\tiny{{$\pm$}0.01} & {\color{red}97.04\tiny{{$\pm$}0.01}} & {\color{red}99.28\tiny{{$\pm$}0.18}} & {\color{red}99.18\tiny{{$\pm$}0.00}} \\
 &  & F & 95.92\tiny{{$\pm$}0.01} & 95.89\tiny{{$\pm$}0.01} & 93.12\tiny{{$\pm$}0.01} & 93.73\tiny{{$\pm$}0.01} & 93.08\tiny{{$\pm$}0.01} & 91.77\tiny{{$\pm$}0.01} & 91.09\tiny{{$\pm$}0.02} & 90.16\tiny{{$\pm$}0.01} & 85.98\tiny{{$\pm$}0.03} & 86.61\tiny{{$\pm$}0.01} & {\color{red}91.74\tiny{{$\pm$}0.00}} & {\color{red}95.18\tiny{{$\pm$}1.03}} & {\color{red}95.24\tiny{{$\pm$}0.01}} \\
\midrule
\multirow{4}{*}{\textbf{2}} & \multirow{4}{*}{\textbf{SPEIT}} & E & 91.92\tiny{{$\pm$}0.04} & 91.58\tiny{{$\pm$}0.03} & 91.73\tiny{{$\pm$}0.05} & 87.18\tiny{{$\pm$}0.08} & 88.81\tiny{{$\pm$}0.02} & 88.41\tiny{{$\pm$}0.05} & 89.81\tiny{{$\pm$}0.05} & 86.65\tiny{{$\pm$}0.03} & 83.10\tiny{{$\pm$}0.07} & 76.55\tiny{{$\pm$}0.10} & 87.57\tiny{{$\pm$}0.02} & 91.74\tiny{{$\pm$}0.14} & 91.35\tiny{{$\pm$}0.03} \\
 &  & M & 98.27\tiny{{$\pm$}0.01} & 97.84\tiny{{$\pm$}0.01} & 97.98\tiny{{$\pm$}0.02} & 96.19\tiny{{$\pm$}0.02} & 95.28\tiny{{$\pm$}0.01} & 96.60\tiny{{$\pm$}0.02} & 97.77\tiny{{$\pm$}0.01} & 94.03\tiny{{$\pm$}0.02} & 96.74\tiny{{$\pm$}0.02} & 92.26\tiny{{$\pm$}0.03} & 96.30\tiny{{$\pm$}0.00} & 98.03\tiny{{$\pm$}0.18} & 97.97\tiny{{$\pm$}0.01} \\
 &  & H & 99.54\tiny{{$\pm$}0.01} & 99.20\tiny{{$\pm$}0.01} & 98.81\tiny{{$\pm$}0.01} & 98.17\tiny{{$\pm$}0.01} & 96.04\tiny{{$\pm$}0.01} & 97.97\tiny{{$\pm$}0.01} & 99.22\tiny{{$\pm$}0.01} & 90.69\tiny{{$\pm$}0.03} & 98.22\tiny{{$\pm$}0.03} & 93.43\tiny{{$\pm$}0.02} & 97.13\tiny{{$\pm$}0.00} & 99.32\tiny{{$\pm$}0.16} & 99.21\tiny{{$\pm$}0.01} \\
 &  & F & 96.31\tiny{{$\pm$}0.02} & 96.31\tiny{{$\pm$}0.02} & 95.77\tiny{{$\pm$}0.02} & 93.76\tiny{{$\pm$}0.02} & 93.59\tiny{{$\pm$}0.01} & 93.76\tiny{{$\pm$}0.03} & 95.57\tiny{{$\pm$}0.01} & 90.21\tiny{{$\pm$}0.02} & 92.96\tiny{{$\pm$}0.04} & 87.02\tiny{{$\pm$}0.03} & 93.53\tiny{{$\pm$}0.01} & 96.13\tiny{{$\pm$}0.26} & 95.96\tiny{{$\pm$}0.01} \\
\midrule
\multirow{4}{*}{\textbf{3}} & \multirow{4}{*}{\textbf{FGSM}} & E & 91.77\tiny{{$\pm$}0.06} & 91.53\tiny{{$\pm$}0.05} & 92.60\tiny{{$\pm$}0.04} & 87.38\tiny{{$\pm$}0.05} & 89.06\tiny{{$\pm$}0.01} & 90.52\tiny{{$\pm$}0.05} & 89.70\tiny{{$\pm$}0.03} & 86.78\tiny{{$\pm$}0.03} & 88.24\tiny{{$\pm$}0.06} & 78.51\tiny{{$\pm$}0.12} & 88.61\tiny{{$\pm$}0.02} & 91.97\tiny{{$\pm$}0.46} & 91.92\tiny{{$\pm$}0.03} \\
 &  & M & 98.26\tiny{{$\pm$}0.01} & 97.74\tiny{{$\pm$}0.01} & 98.13\tiny{{$\pm$}0.01} & 96.18\tiny{{$\pm$}0.01} & 95.47\tiny{{$\pm$}0.01} & 96.94\tiny{{$\pm$}0.03} & 97.68\tiny{{$\pm$}0.01} & 94.01\tiny{{$\pm$}0.02} & 96.52\tiny{{$\pm$}0.04} & 92.65\tiny{{$\pm$}0.01} & 96.36\tiny{{$\pm$}0.01} & 98.04\tiny{{$\pm$}0.22} & 97.99\tiny{{$\pm$}0.01} \\
 &  & H & 99.55\tiny{{$\pm$}0.01} & 99.02\tiny{{$\pm$}0.01} & 99.16\tiny{{$\pm$}0.01} & 98.17\tiny{{$\pm$}0.02} & 95.94\tiny{{$\pm$}0.01} & 98.32\tiny{{$\pm$}0.01} & 99.08\tiny{{$\pm$}0.01} & 91.01\tiny{{$\pm$}0.02} & 98.48\tiny{{$\pm$}0.02} & 93.65\tiny{{$\pm$}0.02} & 97.24\tiny{{$\pm$}0.00} & 99.27\tiny{{$\pm$}0.21} & 99.23\tiny{{$\pm$}0.01} \\
 &  & F & 96.48\tiny{{$\pm$}0.01} & 95.90\tiny{{$\pm$}0.01} & 96.74\tiny{{$\pm$}0.01} & 93.91\tiny{{$\pm$}0.01} & 93.57\tiny{{$\pm$}0.01} & 95.24\tiny{{$\pm$}0.02} & 95.22\tiny{{$\pm$}0.01} & 90.54\tiny{{$\pm$}0.02} & 93.78\tiny{{$\pm$}0.03} & 89.01\tiny{{$\pm$}0.03} & 94.04\tiny{{$\pm$}0.01} & 96.37\tiny{{$\pm$}0.35} & 96.32\tiny{{$\pm$}0.01} \\
\midrule
\multirow{4}{*}{\textbf{4}} & \multirow{4}{*}{\textbf{RND}} & E & 92.04\tiny{{$\pm$}0.04} & 91.75\tiny{{$\pm$}0.04} & 92.60\tiny{{$\pm$}0.03} & 87.09\tiny{{$\pm$}0.04} & 88.87\tiny{{$\pm$}0.03} & 89.55\tiny{{$\pm$}0.05} & 90.00\tiny{{$\pm$}0.04} & 86.71\tiny{{$\pm$}0.03} & 88.12\tiny{{$\pm$}0.09} & 77.27\tiny{{$\pm$}0.10} & 88.40\tiny{{$\pm$}0.02} & 92.13\tiny{{$\pm$}0.36} & 91.94\tiny{{$\pm$}0.02} \\
 &  & M & 98.28\tiny{{$\pm$}0.02} & 97.82\tiny{{$\pm$}0.01} & 98.13\tiny{{$\pm$}0.01} & 96.17\tiny{{$\pm$}0.02} & 95.42\tiny{{$\pm$}0.02} & 96.84\tiny{{$\pm$}0.03} & 97.75\tiny{{$\pm$}0.01} & 94.07\tiny{{$\pm$}0.02} & 97.14\tiny{{$\pm$}0.02} & 92.43\tiny{{$\pm$}0.03} & 96.40\tiny{{$\pm$}0.01} & 98.08\tiny{{$\pm$}0.19} & 98.02\tiny{{$\pm$}0.01} \\
 &  & H & 99.54\tiny{{$\pm$}0.01} & 99.13\tiny{{$\pm$}0.01} & 99.00\tiny{{$\pm$}0.01} & 98.17\tiny{{$\pm$}0.01} & 96.21\tiny{{$\pm$}0.02} & 98.15\tiny{{$\pm$}0.01} & 99.12\tiny{{$\pm$}0.00} & 90.89\tiny{{$\pm$}0.03} & 98.34\tiny{{$\pm$}0.03} & 93.50\tiny{{$\pm$}0.01} & 97.21\tiny{{$\pm$}0.01} & 99.27\tiny{{$\pm$}0.19} & 99.21\tiny{{$\pm$}0.00} \\
 &  & F & 96.60\tiny{{$\pm$}0.01} & 96.30\tiny{{$\pm$}0.02} & 96.54\tiny{{$\pm$}0.01} & 93.84\tiny{{$\pm$}0.02} & 93.23\tiny{{$\pm$}0.02} & 94.88\tiny{{$\pm$}0.03} & 95.63\tiny{{$\pm$}0.01} & 90.37\tiny{{$\pm$}0.03} & 94.59\tiny{{$\pm$}0.02} & 87.72\tiny{{$\pm$}0.03} & 93.97\tiny{{$\pm$}0.01} & 96.48\tiny{{$\pm$}0.13} & 96.27\tiny{{$\pm$}0.01} \\
\midrule
\multirow{4}{*}{\textbf{5}} & \multirow{4}{*}{\textbf{PGD}} & E & 91.74\tiny{{$\pm$}0.06} & 91.56\tiny{{$\pm$}0.04} & 92.63\tiny{{$\pm$}0.04} & 87.39\tiny{{$\pm$}0.04} & 89.06\tiny{{$\pm$}0.02} & 90.58\tiny{{$\pm$}0.06} & 89.72\tiny{{$\pm$}0.04} & 86.77\tiny{{$\pm$}0.03} & 88.25\tiny{{$\pm$}0.10} & 78.53\tiny{{$\pm$}0.07} & 88.62\tiny{{$\pm$}0.02} & 91.98\tiny{{$\pm$}0.47} & 91.94\tiny{{$\pm$}0.03} \\
 &  & M & 98.26\tiny{{$\pm$}0.01} & 97.74\tiny{{$\pm$}0.01} & 98.12\tiny{{$\pm$}0.02} & 96.19\tiny{{$\pm$}0.01} & 95.46\tiny{{$\pm$}0.02} & 96.94\tiny{{$\pm$}0.01} & 97.68\tiny{{$\pm$}0.01} & 94.01\tiny{{$\pm$}0.03} & 96.52\tiny{{$\pm$}0.05} & 92.65\tiny{{$\pm$}0.04} & 96.36\tiny{{$\pm$}0.01} & 98.04\tiny{{$\pm$}0.22} & 97.98\tiny{{$\pm$}0.01} \\
 &  & H & 99.55\tiny{{$\pm$}0.01} & 99.02\tiny{{$\pm$}0.01} & 99.16\tiny{{$\pm$}0.01} & 98.18\tiny{{$\pm$}0.01} & 95.94\tiny{{$\pm$}0.01} & 98.32\tiny{{$\pm$}0.01} & 99.08\tiny{{$\pm$}0.01} & 91.00\tiny{{$\pm$}0.03} & 98.47\tiny{{$\pm$}0.03} & 93.66\tiny{{$\pm$}0.02} & 97.24\tiny{{$\pm$}0.00} & 99.26\tiny{{$\pm$}0.21} & 99.23\tiny{{$\pm$}0.00} \\
 &  & F & 96.48\tiny{{$\pm$}0.02} & 95.91\tiny{{$\pm$}0.01} & 96.74\tiny{{$\pm$}0.01} & 93.91\tiny{{$\pm$}0.02} & 93.56\tiny{{$\pm$}0.01} & 95.24\tiny{{$\pm$}0.03} & 95.23\tiny{{$\pm$}0.02} & 90.54\tiny{{$\pm$}0.02} & 93.75\tiny{{$\pm$}0.02} & 88.99\tiny{{$\pm$}0.03} & 94.03\tiny{{$\pm$}0.01} & 96.38\tiny{{$\pm$}0.35} & 96.32\tiny{{$\pm$}0.01} \\
\midrule
\multirow{4}{*}{\textbf{6}} & \multirow{4}{*}{\textbf{W/O Attack}} & E & 92.19\tiny{{$\pm$}0.00} & 92.11\tiny{{$\pm$}0.00} & 93.05\tiny{{$\pm$}0.00} & 87.97\tiny{{$\pm$}0.00} & 89.14\tiny{{$\pm$}0.00} & 90.99\tiny{{$\pm$}0.00} & 90.17\tiny{{$\pm$}0.00} & 86.86\tiny{{$\pm$}0.00} & 89.96\tiny{{$\pm$}0.00} & 83.12\tiny{{$\pm$}0.00} & 89.56\tiny{{$\pm$}0.00} & 92.45\tiny{{$\pm$}0.43} & 92.41\tiny{{$\pm$}0.00} \\
 &  & M & 98.30\tiny{{$\pm$}0.00} & 97.82\tiny{{$\pm$}0.00} & 98.23\tiny{{$\pm$}0.00} & 96.27\tiny{{$\pm$}0.00} & 95.75\tiny{{$\pm$}0.00} & 97.07\tiny{{$\pm$}0.00} & 97.74\tiny{{$\pm$}0.00} & 93.86\tiny{{$\pm$}0.00} & 97.21\tiny{{$\pm$}0.00} & 93.48\tiny{{$\pm$}0.00} & 96.57\tiny{{$\pm$}0.00} & 98.12\tiny{{$\pm$}0.21} & 98.06\tiny{{$\pm$}0.00} \\
 &  & H & 99.55\tiny{{$\pm$}0.00} & 99.16\tiny{{$\pm$}0.00} & 99.03\tiny{{$\pm$}0.00} & 98.20\tiny{{$\pm$}0.00} & 96.46\tiny{{$\pm$}0.00} & 98.12\tiny{{$\pm$}0.00} & 99.12\tiny{{$\pm$}0.00} & 90.72\tiny{{$\pm$}0.00} & 98.30\tiny{{$\pm$}0.00} & 93.84\tiny{{$\pm$}0.00} & 97.25\tiny{{$\pm$}0.00} & 99.28\tiny{{$\pm$}0.19} & 99.23\tiny{{$\pm$}0.00} \\
 &  & F & 96.68\tiny{{$\pm$}0.00} & 96.37\tiny{{$\pm$}0.00} & 96.77\tiny{{$\pm$}0.00} & 94.15\tiny{{$\pm$}0.00} & 93.78\tiny{{$\pm$}0.00} & 95.39\tiny{{$\pm$}0.00} & 95.68\tiny{{$\pm$}0.00} & 90.48\tiny{{$\pm$}0.00} & 95.16\tiny{{$\pm$}0.00} & 90.15\tiny{{$\pm$}0.00} & 94.46\tiny{{$\pm$}0.00} & 96.61\tiny{{$\pm$}0.17} & 96.46\tiny{{$\pm$}0.00} \\
\midrule
\multicolumn{2}{c}{\multirow{4}{*}{\begin{tabular}[c]{@{}c@{}}\textbf{Avg.}\\\textbf{Accuracy}\end{tabular}}} & E & {\color{blue}91.53\tiny{{$\pm$}0.01}} & 91.52\tiny{{$\pm$}0.02} & 88.80\tiny{{$\pm$}0.05} & 86.87\tiny{{$\pm$}0.02} & 88.89\tiny{{$\pm$}0.01} & 88.15\tiny{{$\pm$}0.03} & 88.37\tiny{{$\pm$}0.04} & 86.68\tiny{{$\pm$}0.01} & 84.78\tiny{{$\pm$}0.03} & 78.39\tiny{{$\pm$}0.03} & - & - & - \\
\multicolumn{2}{c}{} & M & {\color{blue}98.28\tiny{{$\pm$}0.00}} & 97.79\tiny{{$\pm$}0.01} & 98.10\tiny{{$\pm$}0.01} & 96.20\tiny{{$\pm$}0.01} & 95.48\tiny{{$\pm$}0.01} & 96.84\tiny{{$\pm$}0.01} & 97.71\tiny{{$\pm$}0.00} & 94.01\tiny{{$\pm$}0.01} & 96.63\tiny{{$\pm$}0.01} & 92.62\tiny{{$\pm$}0.01} & - & - & - \\
\multicolumn{2}{c}{} & H & {\color{blue}99.54\tiny{{$\pm$}0.00}} & 99.12\tiny{{$\pm$}0.00} & 98.95\tiny{{$\pm$}0.00} & 98.17\tiny{{$\pm$}0.01} & 96.05\tiny{{$\pm$}0.01} & 98.11\tiny{{$\pm$}0.01} & 99.13\tiny{{$\pm$}0.00} & 90.82\tiny{{$\pm$}0.01} & 98.37\tiny{{$\pm$}0.01} & 93.57\tiny{{$\pm$}0.01} & - & - & - \\
\multicolumn{2}{c}{} & F & {\color{blue}96.41\tiny{{$\pm$}0.01}} & 96.11\tiny{{$\pm$}0.01} & 95.95\tiny{{$\pm$}0.01} & 93.88\tiny{{$\pm$}0.01} & 93.47\tiny{{$\pm$}0.01} & 94.38\tiny{{$\pm$}0.01} & 94.74\tiny{{$\pm$}0.01} & 90.38\tiny{{$\pm$}0.01} & 92.70\tiny{{$\pm$}0.01} & 88.25\tiny{{$\pm$}0.01} & - & - & - \\
\midrule
\multicolumn{2}{c}{\multirow{4}{*}{\begin{tabular}[c]{@{}c@{}}\textbf{Avg. 3-Min} \\\textbf{Accuracy}\end{tabular}}} & E & 91.03\tiny{{$\pm$}0.02} & {\color{blue}91.22\tiny{{$\pm$}0.02}} & 84.82\tiny{{$\pm$}0.10} & 86.15\tiny{{$\pm$}0.03} & 88.69\tiny{{$\pm$}0.01} & 85.59\tiny{{$\pm$}0.07} & 86.73\tiny{{$\pm$}0.07} & 86.55\tiny{{$\pm$}0.01} & 80.73\tiny{{$\pm$}0.05} & 76.72\tiny{{$\pm$}0.05} & - & - & - \\
\multicolumn{2}{c}{} & M & {\color{blue}98.26\tiny{{$\pm$}0.01}} & 97.75\tiny{{$\pm$}0.01} & 98.04\tiny{{$\pm$}0.01} & 96.17\tiny{{$\pm$}0.01} & 95.38\tiny{{$\pm$}0.01} & 96.69\tiny{{$\pm$}0.01} & 97.67\tiny{{$\pm$}0.00} & 93.96\tiny{{$\pm$}0.01} & 96.23\tiny{{$\pm$}0.03} & 92.32\tiny{{$\pm$}0.02} & - & - & - \\
\multicolumn{2}{c}{} & H & {\color{blue}99.54\tiny{{$\pm$}0.00}} & 99.06\tiny{{$\pm$}0.00} & 98.78\tiny{{$\pm$}0.01} & 98.16\tiny{{$\pm$}0.01} & 95.86\tiny{{$\pm$}0.01} & 97.95\tiny{{$\pm$}0.01} & 99.09\tiny{{$\pm$}0.00} & 90.67\tiny{{$\pm$}0.01} & 98.28\tiny{{$\pm$}0.01} & 93.42\tiny{{$\pm$}0.01} & - & - & - \\
\multicolumn{2}{c}{} & F & {\color{blue}96.24\tiny{{$\pm$}0.01}} & 95.90\tiny{{$\pm$}0.01} & 95.14\tiny{{$\pm$}0.01} & 93.78\tiny{{$\pm$}0.01} & 93.29\tiny{{$\pm$}0.01} & 93.47\tiny{{$\pm$}0.01} & 93.85\tiny{{$\pm$}0.01} & 90.24\tiny{{$\pm$}0.02} & 90.89\tiny{{$\pm$}0.02} & 87.12\tiny{{$\pm$}0.01} & - & - & - \\
\midrule
\multicolumn{2}{c}{\multirow{4}{*}{\begin{tabular}[c]{@{}c@{}}\textbf{Weighted} \\\textbf{Accuracy}\end{tabular}}} & E & 90.31\tiny{{$\pm$}0.03} & {\color{blue}90.92\tiny{{$\pm$}0.02}} & 77.42\tiny{{$\pm$}0.19} & 85.18\tiny{{$\pm$}0.02} & 88.55\tiny{{$\pm$}0.02} & 82.26\tiny{{$\pm$}0.13} & 83.75\tiny{{$\pm$}0.12} & 86.43\tiny{{$\pm$}0.02} & 75.84\tiny{{$\pm$}0.07} & 76.73\tiny{{$\pm$}0.03} & - & - & - \\
\multicolumn{2}{c}{} & M & {\color{blue}98.26\tiny{{$\pm$}0.01}} & 97.75\tiny{{$\pm$}0.01} & 98.01\tiny{{$\pm$}0.01} & 96.17\tiny{{$\pm$}0.01} & 95.34\tiny{{$\pm$}0.01} & 96.65\tiny{{$\pm$}0.01} & 97.67\tiny{{$\pm$}0.00} & 93.91\tiny{{$\pm$}0.00} & 95.98\tiny{{$\pm$}0.02} & 92.32\tiny{{$\pm$}0.02} & - & - & - \\
\multicolumn{2}{c}{} & H & {\color{blue}99.53\tiny{{$\pm$}0.00}} & 99.04\tiny{{$\pm$}0.00} & 98.67\tiny{{$\pm$}0.01} & 98.16\tiny{{$\pm$}0.01} & 95.79\tiny{{$\pm$}0.01} & 97.87\tiny{{$\pm$}0.01} & 99.09\tiny{{$\pm$}0.00} & 90.66\tiny{{$\pm$}0.01} & 98.26\tiny{{$\pm$}0.02} & 93.39\tiny{{$\pm$}0.00} & - & - & - \\
\multicolumn{2}{c}{} & F & {\color{blue}96.09\tiny{{$\pm$}0.01}} & 95.93\tiny{{$\pm$}0.01} & 94.14\tiny{{$\pm$}0.01} & 93.76\tiny{{$\pm$}0.01} & 93.19\tiny{{$\pm$}0.00} & 92.64\tiny{{$\pm$}0.01} & 92.49\tiny{{$\pm$}0.01} & 90.22\tiny{{$\pm$}0.01} & 88.46\tiny{{$\pm$}0.03} & 86.99\tiny{{$\pm$}0.01} & - & - & - \\
\bottomrule
\end{tabular}}
\label{tab:leaderboard_reddit_top10}
\end{table}

\begin{table}[!htp]
\vspace{-0.3cm}
\caption{\textit{grb-aminer} leaderboard (Top 5 ATK.\ vs.\ Top 10 DEF.) in \textit{graph injection} scenario.}
\centering
\scalebox{0.46}{
\begin{tabular}{cccccccccccccccc} 
\toprule
\multicolumn{3}{c}{\multirow{2}{*}{\diagbox{\textbf{Attacks}}{\textbf{Defenses}}}} & \textbf{1} & \textbf{2} & \textbf{3} & \textbf{4} & \textbf{5} & \textbf{6} & \textbf{7} & \textbf{8} & \textbf{9} & \textbf{10} & \multirow{2}{*}{\begin{tabular}[c]{@{}c@{}}\textbf{Avg.}\\\textbf{Accuracy~}\end{tabular}} & \multirow{2}{*}{\begin{tabular}[c]{@{}c@{}}\textbf{Avg. 3-Max}\\\textbf{Accuracy}\end{tabular}} & \multirow{2}{*}{\begin{tabular}[c]{@{}c@{}}\textbf{Weighted}\\\textbf{Accuracy}\end{tabular}} \\
\multicolumn{3}{c}{} & \textbf{GAT\tiny{+AT}} & \textbf{R-GCN\tiny{+AT}} & \textbf{SGCN\tiny{+LN}} & \textbf{R-GCN} & \textbf{GCN\tiny{+LN}} & \textbf{GAT\tiny{LN}} & \textbf{GIN\tiny{+LN}} & \textbf{TAGCN\tiny{+LN}} & \textbf{TAGCN\tiny{+AT}} & \textbf{GAT} &  &  &  \\
\midrule
\multirow{4}{*}{\textbf{1}} & \multirow{4}{*}{\textbf{TDGIA}} & E & 59.54\tiny{{$\pm$}0.05} & 56.83\tiny{{$\pm$}0.06} & 56.73\tiny{{$\pm$}0.06} & 56.12\tiny{{$\pm$}0.07} & 53.51\tiny{{$\pm$}0.21} & 43.93\tiny{{$\pm$}0.41} & 51.10\tiny{{$\pm$}0.12} & 54.63\tiny{{$\pm$}0.20} & 49.59\tiny{{$\pm$}0.50} & 42.40\tiny{{$\pm$}0.52} & {\color{red}52.44\tiny{{$\pm$}0.17}} & {\color{red}57.70\tiny{{$\pm$}1.31}} & {\color{red}58.08\tiny{{$\pm$}0.04}} \\
 &  & M & 68.39\tiny{{$\pm$}0.02} & 65.61\tiny{{$\pm$}0.02} & 66.11\tiny{{$\pm$}0.02} & 65.23\tiny{{$\pm$}0.03} & 66.78\tiny{{$\pm$}0.05} & 61.84\tiny{{$\pm$}1.20} & 64.49\tiny{{$\pm$}0.10} & 64.62\tiny{{$\pm$}0.02} & 67.27\tiny{{$\pm$}0.04} & 62.47\tiny{{$\pm$}1.01} & {\color{red}65.28\tiny{{$\pm$}0.23}} & {\color{red}67.48\tiny{{$\pm$}0.68}} & {\color{red}67.69\tiny{{$\pm$}0.02}} \\
 &  & H & 75.83\tiny{{$\pm$}0.02} & 72.35\tiny{{$\pm$}0.02} & 72.10\tiny{{$\pm$}0.00} & 71.94\tiny{{$\pm$}0.02} & 73.39\tiny{{$\pm$}0.02} & 75.22\tiny{{$\pm$}0.04} & 72.92\tiny{{$\pm$}0.02} & 68.94\tiny{{$\pm$}0.03} & 73.98\tiny{{$\pm$}0.01} & 75.03\tiny{{$\pm$}0.03} & 73.17\tiny{{$\pm$}0.01} & 75.36\tiny{{$\pm$}0.34} & {\color{red}75.33\tiny{{$\pm$}0.01}} \\
 &  & F & 67.69\tiny{{$\pm$}0.03} & 63.62\tiny{{$\pm$}0.32} & 62.20\tiny{{$\pm$}0.15} & 61.99\tiny{{$\pm$}0.22} & 60.38\tiny{{$\pm$}1.46} & 59.69\tiny{{$\pm$}1.57} & 59.59\tiny{{$\pm$}0.42} & 59.06\tiny{{$\pm$}1.75} & 57.24\tiny{{$\pm$}5.04} & 56.63\tiny{{$\pm$}6.75} & 60.81\tiny{{$\pm$}1.71} & {\color{red}64.52\tiny{{$\pm$}2.32}} & {\color{red}65.74\tiny{{$\pm$}0.21}} \\
 \midrule
\multirow{4}{*}{\textbf{2}} & \multirow{4}{*}{\textbf{SPEIT}} & E & 59.54\tiny{{$\pm$}0.07} & 56.80\tiny{{$\pm$}0.05} & 56.94\tiny{{$\pm$}0.10} & 55.64\tiny{{$\pm$}0.10} & 56.15\tiny{{$\pm$}0.06} & 56.13\tiny{{$\pm$}0.07} & 54.24\tiny{{$\pm$}0.09} & 56.61\tiny{{$\pm$}0.06} & 56.59\tiny{{$\pm$}0.08} & 57.36\tiny{{$\pm$}0.09} & 56.60\tiny{{$\pm$}0.04} & 57.95\tiny{{$\pm$}1.14} & 58.62\tiny{{$\pm$}0.05} \\
 &  & M & 68.37\tiny{{$\pm$}0.03} & 65.46\tiny{{$\pm$}0.03} & 66.20\tiny{{$\pm$}0.02} & 65.25\tiny{{$\pm$}0.05} & 66.75\tiny{{$\pm$}0.03} & 67.49\tiny{{$\pm$}0.06} & 65.05\tiny{{$\pm$}0.06} & 64.47\tiny{{$\pm$}0.04} & 66.95\tiny{{$\pm$}0.05} & 66.81\tiny{{$\pm$}0.04} & 66.28\tiny{{$\pm$}0.02} & 67.60\tiny{{$\pm$}0.59} & 67.86\tiny{{$\pm$}0.03} \\
 &  & H & 75.94\tiny{{$\pm$}0.04} & 72.27\tiny{{$\pm$}0.03} & 72.36\tiny{{$\pm$}0.03} & 71.86\tiny{{$\pm$}0.03} & 73.41\tiny{{$\pm$}0.01} & 75.34\tiny{{$\pm$}0.03} & 72.87\tiny{{$\pm$}0.03} & 68.88\tiny{{$\pm$}0.05} & 73.98\tiny{{$\pm$}0.02} & 73.83\tiny{{$\pm$}0.04} & {\color{red}73.07\tiny{{$\pm$}0.01}} & {\color{red}75.08\tiny{{$\pm$}0.82}} & 75.33\tiny{{$\pm$}0.02} \\
 &  & F & 68.04\tiny{{$\pm$}0.03} & 64.05\tiny{{$\pm$}0.04} & 64.84\tiny{{$\pm$}0.04} & 64.06\tiny{{$\pm$}0.04} & 65.51\tiny{{$\pm$}0.02} & 64.02\tiny{{$\pm$}0.04} & 63.11\tiny{{$\pm$}0.02} & 62.59\tiny{{$\pm$}0.04} & 63.77\tiny{{$\pm$}0.06} & 63.58\tiny{{$\pm$}0.06} & 64.36\tiny{{$\pm$}0.02} & 66.13\tiny{{$\pm$}1.38} & 66.89\tiny{{$\pm$}0.02} \\
 \midrule
\multirow{4}{*}{\textbf{3}} & \multirow{4}{*}{\textbf{RND}} & E & 59.56\tiny{{$\pm$}0.06} & 57.53\tiny{{$\pm$}0.06} & 57.41\tiny{{$\pm$}0.06} & 56.38\tiny{{$\pm$}0.11} & 57.76\tiny{{$\pm$}0.05} & 58.83\tiny{{$\pm$}0.10} & 54.41\tiny{{$\pm$}0.13} & 58.07\tiny{{$\pm$}0.12} & 58.14\tiny{{$\pm$}0.04} & 57.46\tiny{{$\pm$}0.10} & 57.55\tiny{{$\pm$}0.03} & 58.85\tiny{{$\pm$}0.57} & 59.09\tiny{{$\pm$}0.05} \\
 &  & M & 68.22\tiny{{$\pm$}0.04} & 65.86\tiny{{$\pm$}0.03} & 66.29\tiny{{$\pm$}0.03} & 65.34\tiny{{$\pm$}0.06} & 67.03\tiny{{$\pm$}0.03} & 68.62\tiny{{$\pm$}0.05} & 65.54\tiny{{$\pm$}0.06} & 64.98\tiny{{$\pm$}0.08} & 67.34\tiny{{$\pm$}0.04} & 67.71\tiny{{$\pm$}0.06} & 66.69\tiny{{$\pm$}0.02} & 68.18\tiny{{$\pm$}0.38} & 68.24\tiny{{$\pm$}0.03} \\
 &  & H & 75.75\tiny{{$\pm$}0.02} & 72.66\tiny{{$\pm$}0.02} & 72.42\tiny{{$\pm$}0.03} & 72.00\tiny{{$\pm$}0.03} & 73.52\tiny{{$\pm$}0.02} & 75.63\tiny{{$\pm$}0.03} & 73.36\tiny{{$\pm$}0.03} & 69.30\tiny{{$\pm$}0.06} & 74.04\tiny{{$\pm$}0.02} & 75.36\tiny{{$\pm$}0.03} & 73.40\tiny{{$\pm$}0.01} & 75.58\tiny{{$\pm$}0.17} & 75.39\tiny{{$\pm$}0.01} \\
 &  & F & 67.72\tiny{{$\pm$}0.04} & 64.98\tiny{{$\pm$}0.02} & 65.31\tiny{{$\pm$}0.04} & 64.45\tiny{{$\pm$}0.04} & 66.17\tiny{{$\pm$}0.02} & 67.54\tiny{{$\pm$}0.04} & 64.36\tiny{{$\pm$}0.06} & 64.33\tiny{{$\pm$}0.03} & 66.42\tiny{{$\pm$}0.03} & 66.23\tiny{{$\pm$}0.04} & 65.75\tiny{{$\pm$}0.02} & 67.23\tiny{{$\pm$}0.58} & 67.34\tiny{{$\pm$}0.03} \\
 \midrule
\multirow{4}{*}{\textbf{4}} & \multirow{4}{*}{\textbf{PGD}} & E & 59.70\tiny{{$\pm$}0.06} & 57.71\tiny{{$\pm$}0.05} & 57.73\tiny{{$\pm$}0.09} & 57.19\tiny{{$\pm$}0.07} & 57.60\tiny{{$\pm$}0.08} & 57.05\tiny{{$\pm$}0.17} & 54.69\tiny{{$\pm$}0.09} & 58.18\tiny{{$\pm$}0.07} & 58.27\tiny{{$\pm$}0.09} & 58.46\tiny{{$\pm$}0.11} & 57.66\tiny{{$\pm$}0.05} & 58.81\tiny{{$\pm$}0.64} & 59.14\tiny{{$\pm$}0.05} \\
 &  & M & 68.40\tiny{{$\pm$}0.05} & 66.12\tiny{{$\pm$}0.02} & 66.39\tiny{{$\pm$}0.04} & 65.67\tiny{{$\pm$}0.04} & 67.04\tiny{{$\pm$}0.03} & 68.24\tiny{{$\pm$}0.04} & 65.64\tiny{{$\pm$}0.08} & 65.17\tiny{{$\pm$}0.05} & 67.32\tiny{{$\pm$}0.03} & 67.85\tiny{{$\pm$}0.05} & 66.78\tiny{{$\pm$}0.02} & 68.16\tiny{{$\pm$}0.23} & 68.12\tiny{{$\pm$}0.03} \\
 &  & H & 75.83\tiny{{$\pm$}0.03} & 72.91\tiny{{$\pm$}0.02} & 72.47\tiny{{$\pm$}0.04} & 72.18\tiny{{$\pm$}0.05} & 73.52\tiny{{$\pm$}0.02} & 75.55\tiny{{$\pm$}0.05} & 73.58\tiny{{$\pm$}0.04} & 69.64\tiny{{$\pm$}0.05} & 73.89\tiny{{$\pm$}0.02} & 74.34\tiny{{$\pm$}0.04} & 73.39\tiny{{$\pm$}0.01} & 75.24\tiny{{$\pm$}0.65} & 75.36\tiny{{$\pm$}0.02} \\
 &  & F & 68.01\tiny{{$\pm$}0.02} & 65.41\tiny{{$\pm$}0.01} & 65.54\tiny{{$\pm$}0.03} & 65.05\tiny{{$\pm$}0.03} & 66.22\tiny{{$\pm$}0.02} & 66.49\tiny{{$\pm$}0.04} & 64.63\tiny{{$\pm$}0.04} & 64.82\tiny{{$\pm$}0.04} & 66.32\tiny{{$\pm$}0.02} & 66.14\tiny{{$\pm$}0.04} & 65.86\tiny{{$\pm$}0.01} & 66.94\tiny{{$\pm$}0.76} & 67.37\tiny{{$\pm$}0.02} \\
 \midrule
\multirow{4}{*}{\textbf{5}} & \multirow{4}{*}{\textbf{FGSM}} & E & 59.71\tiny{{$\pm$}0.05} & 57.69\tiny{{$\pm$}0.08} & 57.62\tiny{{$\pm$}0.06} & 57.16\tiny{{$\pm$}0.08} & 57.60\tiny{{$\pm$}0.06} & 56.97\tiny{{$\pm$}0.09} & 54.67\tiny{{$\pm$}0.08} & 58.20\tiny{{$\pm$}0.10} & 58.23\tiny{{$\pm$}0.06} & 58.46\tiny{{$\pm$}0.07} & 57.63\tiny{{$\pm$}0.05} & 58.81\tiny{{$\pm$}0.65} & 59.15\tiny{{$\pm$}0.04} \\
 &  & M & 68.37\tiny{{$\pm$}0.02} & 66.10\tiny{{$\pm$}0.03} & 66.38\tiny{{$\pm$}0.04} & 65.70\tiny{{$\pm$}0.05} & 67.03\tiny{{$\pm$}0.04} & 68.27\tiny{{$\pm$}0.04} & 65.61\tiny{{$\pm$}0.08} & 65.16\tiny{{$\pm$}0.05} & 67.30\tiny{{$\pm$}0.02} & 67.84\tiny{{$\pm$}0.07} & 66.78\tiny{{$\pm$}0.02} & 68.16\tiny{{$\pm$}0.23} & 68.11\tiny{{$\pm$}0.02} \\
 &  & H & 75.82\tiny{{$\pm$}0.02} & 72.92\tiny{{$\pm$}0.04} & 72.48\tiny{{$\pm$}0.03} & 72.18\tiny{{$\pm$}0.05} & 73.52\tiny{{$\pm$}0.02} & 75.55\tiny{{$\pm$}0.05} & 73.60\tiny{{$\pm$}0.04} & 69.64\tiny{{$\pm$}0.04} & 73.90\tiny{{$\pm$}0.01} & 74.34\tiny{{$\pm$}0.04} & 73.39\tiny{{$\pm$}0.01} & 75.23\tiny{{$\pm$}0.65} & 75.35\tiny{{$\pm$}0.02} \\
 &  & F & 68.00\tiny{{$\pm$}0.02} & 65.41\tiny{{$\pm$}0.02} & 65.54\tiny{{$\pm$}0.04} & 65.05\tiny{{$\pm$}0.04} & 66.22\tiny{{$\pm$}0.02} & 66.50\tiny{{$\pm$}0.06} & 64.65\tiny{{$\pm$}0.04} & 64.82\tiny{{$\pm$}0.03} & 66.34\tiny{{$\pm$}0.03} & 66.15\tiny{{$\pm$}0.06} & 65.87\tiny{{$\pm$}0.01} & 66.95\tiny{{$\pm$}0.75} & 67.37\tiny{{$\pm$}0.01} \\
 \midrule
\multirow{4}{*}{\textbf{6}} & \multirow{4}{*}{\textbf{W/O Attack}} & E & 59.67\tiny{{$\pm$}0.00} & 58.08\tiny{{$\pm$}0.00} & 60.22\tiny{{$\pm$}0.00} & 58.53\tiny{{$\pm$}0.00} & 58.14\tiny{{$\pm$}0.00} & 60.78\tiny{{$\pm$}0.00} & 56.83\tiny{{$\pm$}0.00} & 59.47\tiny{{$\pm$}0.00} & 59.62\tiny{{$\pm$}0.00} & 59.88\tiny{{$\pm$}0.00} & 59.12\tiny{{$\pm$}0.00} & 60.29\tiny{{$\pm$}0.37} & 60.42\tiny{{$\pm$}0.00} \\
 &  & M & 68.28\tiny{{$\pm$}0.00} & 66.14\tiny{{$\pm$}0.00} & 67.11\tiny{{$\pm$}0.00} & 66.35\tiny{{$\pm$}0.00} & 67.00\tiny{{$\pm$}0.00} & 68.98\tiny{{$\pm$}0.00} & 66.26\tiny{{$\pm$}0.00} & 65.41\tiny{{$\pm$}0.00} & 67.53\tiny{{$\pm$}0.00} & 68.41\tiny{{$\pm$}0.00} & 67.15\tiny{{$\pm$}0.00} & 68.56\tiny{{$\pm$}0.30} & 68.59\tiny{{$\pm$}0.00} \\
 &  & H & 75.85\tiny{{$\pm$}0.00} & 73.05\tiny{{$\pm$}0.00} & 72.69\tiny{{$\pm$}0.00} & 72.66\tiny{{$\pm$}0.00} & 73.46\tiny{{$\pm$}0.00} & 75.64\tiny{{$\pm$}0.00} & 73.69\tiny{{$\pm$}0.00} & 69.84\tiny{{$\pm$}0.00} & 74.10\tiny{{$\pm$}0.00} & 75.76\tiny{{$\pm$}0.00} & 73.67\tiny{{$\pm$}0.00} & 75.75\tiny{{$\pm$}0.09} & 75.52\tiny{{$\pm$}0.00} \\
 &  & F & 67.93\tiny{{$\pm$}0.00} & 65.76\tiny{{$\pm$}0.00} & 66.68\tiny{{$\pm$}0.00} & 65.85\tiny{{$\pm$}0.00} & 66.20\tiny{{$\pm$}0.00} & 68.47\tiny{{$\pm$}0.00} & 65.59\tiny{{$\pm$}0.00} & 64.91\tiny{{$\pm$}0.00} & 67.08\tiny{{$\pm$}0.00} & 68.02\tiny{{$\pm$}0.00} & 66.65\tiny{{$\pm$}0.00} & 68.14\tiny{{$\pm$}0.24} & 68.11\tiny{{$\pm$}0.00} \\
\midrule
\multicolumn{2}{c}{\multirow{4}{*}{\begin{tabular}[c]{@{}c@{}}\textbf{Avg.}\\\textbf{Accuracy}\end{tabular}}} & E & {\color{blue}59.62\tiny{{$\pm$}0.02}} & 57.44\tiny{{$\pm$}0.03} & 57.77\tiny{{$\pm$}0.03} & 56.84\tiny{{$\pm$}0.04} & 56.79\tiny{{$\pm$}0.04} & 55.62\tiny{{$\pm$}0.06} & 54.33\tiny{{$\pm$}0.04} & 57.53\tiny{{$\pm$}0.05} & 56.74\tiny{{$\pm$}0.09} & 55.67\tiny{{$\pm$}0.10} & - & - & - \\
\multicolumn{2}{c}{} & M & {\color{blue}68.34\tiny{{$\pm$}0.01}} & 65.88\tiny{{$\pm$}0.01} & 66.41\tiny{{$\pm$}0.01} & 65.59\tiny{{$\pm$}0.02} & 66.94\tiny{{$\pm$}0.02} & 67.24\tiny{{$\pm$}0.19} & 65.43\tiny{{$\pm$}0.03} & 64.97\tiny{{$\pm$}0.02} & 67.28\tiny{{$\pm$}0.01} & 66.85\tiny{{$\pm$}0.18} & - & - & - \\
\multicolumn{2}{c}{} & H & {\color{blue}75.84\tiny{{$\pm$}0.01}} & 72.69\tiny{{$\pm$}0.01} & 72.42\tiny{{$\pm$}0.01} & 72.14\tiny{{$\pm$}0.02} & 73.47\tiny{{$\pm$}0.01} & 75.49\tiny{{$\pm$}0.01} & 73.33\tiny{{$\pm$}0.02} & 69.38\tiny{{$\pm$}0.02} & 73.98\tiny{{$\pm$}0.00} & 74.78\tiny{{$\pm$}0.02} & - & - & - \\
\multicolumn{2}{c}{} & F & {\color{blue}67.90\tiny{{$\pm$}0.01}} & 64.87\tiny{{$\pm$}0.05} & 65.02\tiny{{$\pm$}0.03} & 64.41\tiny{{$\pm$}0.04} & 65.12\tiny{{$\pm$}0.25} & 65.45\tiny{{$\pm$}0.26} & 63.65\tiny{{$\pm$}0.07} & 63.42\tiny{{$\pm$}0.29} & 64.53\tiny{{$\pm$}0.84} & 64.46\tiny{{$\pm$}1.13} & - & - & - \\
\midrule
\multicolumn{2}{c}{\multirow{4}{*}{\begin{tabular}[c]{@{}c@{}}\textbf{Avg. 3-Min} \\\textbf{Accuracy}\end{tabular}}} & E & {\color{blue}59.55\tiny{{$\pm$}0.03}} & 57.05\tiny{{$\pm$}0.04} & 57.02\tiny{{$\pm$}0.03} & 56.05\tiny{{$\pm$}0.07} & 55.73\tiny{{$\pm$}0.07} & 52.33\tiny{{$\pm$}0.12} & 53.25\tiny{{$\pm$}0.07} & 56.43\tiny{{$\pm$}0.07} & 54.77\tiny{{$\pm$}0.16} & 52.41\tiny{{$\pm$}0.17} & - & - & - \\
\multicolumn{2}{c}{} & M & {\color{blue}68.28\tiny{{$\pm$}0.01}} & 65.64\tiny{{$\pm$}0.02} & 66.20\tiny{{$\pm$}0.01} & 65.28\tiny{{$\pm$}0.03} & 66.84\tiny{{$\pm$}0.02} & 65.85\tiny{{$\pm$}0.40} & 65.02\tiny{{$\pm$}0.04} & 64.69\tiny{{$\pm$}0.03} & 67.17\tiny{{$\pm$}0.02} & 65.66\tiny{{$\pm$}0.34} & - & - & - \\
\multicolumn{2}{c}{} & H & {\color{blue}75.80\tiny{{$\pm$}0.02}} & 72.42\tiny{{$\pm$}0.02} & 72.29\tiny{{$\pm$}0.01} & 71.93\tiny{{$\pm$}0.02} & 73.42\tiny{{$\pm$}0.01} & 75.36\tiny{{$\pm$}0.02} & 73.05\tiny{{$\pm$}0.02} & 69.04\tiny{{$\pm$}0.03} & 73.92\tiny{{$\pm$}0.01} & 74.17\tiny{{$\pm$}0.03} & - & - & - \\
\multicolumn{2}{c}{} & F & {\color{blue}67.78\tiny{{$\pm$}0.02}} & 64.22\tiny{{$\pm$}0.11} & 64.12\tiny{{$\pm$}0.06} & 63.50\tiny{{$\pm$}0.08} & 64.02\tiny{{$\pm$}0.49} & 63.39\tiny{{$\pm$}0.53} & 62.35\tiny{{$\pm$}0.14} & 61.99\tiny{{$\pm$}0.58} & 62.44\tiny{{$\pm$}1.69} & 62.11\tiny{{$\pm$}2.26} & - & - & - \\
\midrule
\multicolumn{2}{c}{\multirow{4}{*}{\begin{tabular}[c]{@{}c@{}}\textbf{Weighted} \\\textbf{Accuracy}\end{tabular}}} & E & {\color{blue}59.53\tiny{{$\pm$}0.04}} & 56.93\tiny{{$\pm$}0.04} & 56.94\tiny{{$\pm$}0.04} & 55.93\tiny{{$\pm$}0.08} & 54.63\tiny{{$\pm$}0.14} & 48.21\tiny{{$\pm$}0.27} & 52.23\tiny{{$\pm$}0.08} & 55.55\tiny{{$\pm$}0.14} & 52.18\tiny{{$\pm$}0.33} & 47.45\tiny{{$\pm$}0.35} & - & - & - \\
\multicolumn{2}{c}{} & M & {\color{blue}68.25\tiny{{$\pm$}0.02}} & 65.57\tiny{{$\pm$}0.02} & 66.17\tiny{{$\pm$}0.02} & 65.28\tiny{{$\pm$}0.02} & 66.79\tiny{{$\pm$}0.02} & 63.85\tiny{{$\pm$}0.80} & 64.77\tiny{{$\pm$}0.07} & 64.60\tiny{{$\pm$}0.03} & 67.06\tiny{{$\pm$}0.03} & 64.07\tiny{{$\pm$}0.68} & - & - & - \\
\multicolumn{2}{c}{} & H & {\color{blue}75.78\tiny{{$\pm$}0.02}} & 72.37\tiny{{$\pm$}0.02} & 72.20\tiny{{$\pm$}0.01} & 71.92\tiny{{$\pm$}0.03} & 73.41\tiny{{$\pm$}0.01} & 75.30\tiny{{$\pm$}0.02} & 72.98\tiny{{$\pm$}0.02} & 68.99\tiny{{$\pm$}0.04} & 73.91\tiny{{$\pm$}0.01} & 74.08\tiny{{$\pm$}0.03} & - & - & - \\
\multicolumn{2}{c}{} & F & {\color{blue}67.73\tiny{{$\pm$}0.03}} & 63.96\tiny{{$\pm$}0.21} & 63.19\tiny{{$\pm$}0.10} & 62.80\tiny{{$\pm$}0.15} & 62.18\tiny{{$\pm$}0.98} & 61.58\tiny{{$\pm$}1.05} & 61.00\tiny{{$\pm$}0.28} & 60.54\tiny{{$\pm$}1.18} & 59.82\tiny{{$\pm$}3.38} & 59.37\tiny{{$\pm$}4.53} & - & - & - \\
\bottomrule
\end{tabular}}
\label{tab:leaderboard_aminer_top10}
\end{table}

\end{document}